\documentclass[twoside]{article}

\usepackage[accepted]{aistats2022}
\usepackage[round]{natbib}

\usepackage{import}

\usepackage{microtype}
\usepackage{graphicx}

\usepackage{xspace} 
\usepackage{amsmath,amssymb,amsfonts}
\usepackage{etoolbox}

\usepackage[thmmarks, amsmath, thref]{ntheorem}
\usepackage{bbm}
\usepackage{bm}
\usepackage{dsfont}

\usepackage[normalem]{ulem}

\usepackage{epsfig}
\usepackage{graphicx}

\usepackage{xcolor}

\usepackage{subcaption}

\usepackage{diagbox}

\usepackage{lipsum}  

\usepackage{threeparttable}
\usepackage[singlelinecheck=true,justification=centering]{caption}
\usepackage{makecell}
\usepackage{multirow}
\usepackage{color,soul}
\usepackage{xcolor}

\definecolor{dimgray}{rgb}{0.35, 0.35, 0.35}

\usepackage[linesnumbered,algoruled,noend,noline]{algorithm2e}

\subimport{./}{macros}
\theoremstyle{definition}  

\newtheorem{Example}{Example}

\theoremstyle{definition}

\usepackage{booktabs}

\usepackage{float}
\usepackage[backref, colorlinks,citecolor=blue]{hyperref}
\hypersetup{
  colorlinks   = true, 
  urlcolor     = blue, 
  linkcolor    = blue, 
  citecolor   = blue, 
}

\usepackage{ellipsis} 

 \hypersetup{final}  
                    
\newcommand{\footnotestandalone}[1]{\let\thefootnote\relax\footnotetext{#1}}
\begin{document}

%
\runningtitle{Sampling in Dirichlet Process Mixture Models for Clustering Streaming Data}

%
\runningauthor{Or Dinari and Oren Freifeld}

\twocolumn[

\aistatstitle{Sampling in Dirichlet Process Mixture Models \\ for Clustering Streaming Data}

\aistatsauthor{Or Dinari \And Oren Freifeld}

\aistatsaddress{ Ben-Gurion University \And  Ben-Gurion University} ]
\begin{abstract}
    Practical tools for clustering streaming data must be fast enough to handle the arrival rate of the observations. Typically, they also must adapt on the fly to possible lack of stationarity; \ie, the data statistics may be time-dependent due to various forms of drifts, changes in the number of clusters, \etc. The Dirichlet Process Mixture Model (DPMM), whose Bayesian nonparametric nature allows it to adapt its complexity to the data, seems a natural choice for the streaming-data case. In its classical formulation, however, the DPMM cannot capture common types of drifts in the data statistics. Moreover, and regardless of that limitation, existing methods for online DPMM inference are too slow to handle rapid data streams. In this work we propose adapting both the DPMM and a known DPMM sampling-based non-streaming inference method for streaming-data clustering. We demonstrate the utility of the proposed method on several challenging settings, where it obtains state-of-the-art results while being on par with other methods in terms of speed.
  
\end{abstract}

\section{INTRODUCTION}
In today's Age of Data, the need for data-analysis methods that are both fast and effective is more important than ever. In this context, this work focuses on the challenging problem of streaming-data clustering;
namely, the unsupervised-learning task at hand is to cluster observations where the latter arrive constantly at some rate, forming an ever-growing, possibly infinite, data stream.
In this prevalent setting, traditional clustering methods are mostly  inapplicable due to the following reasons.
1) Storing the entire stream in memory is often impractical  
or even impossible. 
2) Even when storing the entire stream is possible, users usually cannot wait until the stream ends; rather, they often need, at any given moment, a reliable estimate of the model based on the data seen so far, and keep updating that estimate as more data arrives.  
3) The statistical properties of the data might be non-stationary~\citep{RamrezGallego:NC:2017:ASO} as, \eg, clusters may appear or disappear,
the distribution underlying a specific cluster may be time-dependent due to various forms of drift, \etc. 
Thus, as new observations arrive, we must be able to update the model and adapt it to the newer data without having to revisit previously-seen observations. 

Here we propose a new streaming-data clustering method based on the Dirichlet Process Mixture Model (DPMM) and, in particular, a specific non-streaming DPMM sampler~\citep{Chang:NIPS:2013:ParallelSamplerDP}.  
The DPMM~\citep{Ferguson:AoS:1973:Bayesian,Antoniak:AoS:1974:DPMM} is a  classical Bayesian Nonparametric (BNP) model~\citep{Hjort:Book:2010:BNP} which is often used in clustering problems where $K$, the number of clusters, is unknown~\citep{Muller:Book:2015:BNP}. 
The DPMM's flexibility and adaptiveness to the data complexity support the inference of $K$ 
(which is also the number of \emph{instantiated} mixture components) together with the rest of the model.
\footnotestandalone{\small {\textbf{Acknowledgements.}
This work was supported by the Lynn and William
Frankel Center at BGU CS, by
the Israeli Council for Higher Education via the BGU Data Science Research Center, and by Israel Science Foundation Personal Grant \#360/21. O.D.~was also funded 
by the Jabotinsky Scholarship from Israel's Ministry of Technology and Science, and by BGU's Hi-Tech Scholarship.}
}
When clustering streaming data,
the ability to modify $K$ is an important property, sought after by most methods in this field. 
As the DPMM possesses that property, to some extent it can be used as is for streaming-data clustering. 
However, one problem that arises is that the DPMM cannot ``forget'' previously-seen observations, even those from the distant past. 
This is a limitation even with DPMM methods
that do no need to keep all the data in memory (such as the Stochastic Online Variational Bayes (SoVB)~\citep{Hoffman:JMLR:2013:sovb} or the Memoized Variational Bayes~\citep{Hughes:NIPS:2013:memoizedDP}); \eg, when the number of observations grows larger and larger, new points have almost no effect on the estimated model.
Another problem caused by the lack of forgetting is the inability to handle incremental concept drifts~\citep{RamrezGallego:NC:2017:ASO}:
when the components change over time, a DPMM method will often opt to instantiate a new component, instead of modifying existing ones.
Lastly, as DPMM inference (at least in most formulations) requires multiple passes over the data, it cannot scale
(in terms of speed and memory) to long data streams and is inapplicable in the case of infinitely-long streams.

To adapt the DPMM for streaming data, we utilize an idea widely used in streaming-data clustering, the Damped Window~\citep{Zubarouglu:AIR:2021:review} (\ie, using finitely-many time-decaying weights). 
When applying this idea in the calculation of the DPMM's posterior distribution, we are able to not only better adapt to new data (thereby accommodating various concept drifts)
but also obviate the need to reiterate over previously-processed observations.
Concretely, we incorporate the Damped Window within a fast and recent implementation~\citep{Dinari:CCGRID:2019:distributed} of Chang and Fisher's
DPMM sampler~\citep{Chang:NIPS:2013:ParallelSamplerDP} and eliminate the need to revisit, during inference,  previously-seen data batches. Consequently, the sampler's speed improves to the point that it becomes on par with other streaming-data clustering methods. 
Additionally, we also introduce a deterministic subroutine (mostly based on predictive posterior distributions)
that, when added to the sampler, significantly improves the performance in the streaming-data clustering task.  
A thorough experimental study (\autoref{Sec:Results}), shows that 
compared with several key methods, our method almost uniformly dominates in several common metrics. 

To summarize, our two main contributions are: 
1) a novel and fast streaming-data clustering method, called \emph{ScStream}, that achieves state-of-the-art (SOTA) results and that was obtained by adapting a DPMM method to the streaming-data setting; 
2) we created and/or adapted several datasets for evaluating clustering methods for streaming data. 
Finally, our Julia code and is available at \url{github.com/BGU-CS-VIL/DPMMSubClustersStreaming.jl}
while its (optional) Python wrapper is available at \url{github.com/BGU-CS-VIL/dpmmpythonStreaming}.

\setlength{\textfloatsep}{8pt}
\setlength{\abovedisplayskip}{6pt}
\setlength{\belowdisplayskip}{5pt}
\section{RELATED WORK}\label{Sec:RelatedWork}
\textbf{Streaming-data clustering:} Most methods for clustering streaming data use a two-phase process. The first phase is online during which new data is obtained, processed, and summarized. 
The second is offline,  
is typically called only upon request,
and generates the clusters. 
An early example of such methods 
is BIRCH~\citep{Zhang:ACM:1996:birch} which introduced \emph{micro-clusters} and \emph{macro-clusters}. 
By maintaining a clustering features (CF) tree structure, whose nodes are called a micro-clusters, 
new points are assigned to micro-clusters based on feature similarity. A new leaf is 
created whenever the best similarity is
insufficient. 
In the offline step the micro-clusters are clustered into so-called macro-clusters whose number is predefined. 
The macro-clusters 
then serve as a predicting model for labeling new data.  
CluStream~\citep{Aggarwal:VLDB:2003:CluStream} improves BIRCH by allowing the clustering over different 
time horizons,
storing not just the point summary but also time-dependent snapshots of the micro clusters.
DenStream~\citep{Cao:SIAM:2006:DenStream} extends BIRCH via a 
time-decaying CF (reducing the weight of older micro-clusters). BIRCH,
together with CluStream and DenStream, inspired other works 
(\eg: A-BIRCH~\citep{Lorbeer:INNS:2017:A-BIRCH}; ScaleKM~\citep{Bradley:KDD:1998:scaleKM}; ACSC~\citep{Fahy:Cybernetics:2018:ACSC}; HCluStream~\citep{Yang:ICDWM:2006:HCluStream}; 
SDStream~\citep{Ren:ICFKD:2009:SDStream}; C-DenStream~\citep{Ruiz:DS:2009:C-DenStream}; 
HDenStream~\citep{Lin:ISECS:2009:HDenStream}; HCDD~\citep{Zgraja:Book:2018:HCDD};
LeaDen-Stream~\citep{Amini:JCC:2013:LeaDen-Stream}) that utilize its key idea.  
\\\\
Another approach is to store cluster medoids (instead of the CF). 
An example is StreamKM++~\citep{Ackermann:JEA:2012:streamkm++}, where a weighted subset (a coreset),
of the data is stored in a tree. BICO~\citep{Fichtenberger:ESA:2013:BICO} combines the StreamKM++ coreset approach with BIRCH's CF 
by storing the coresets in a tree structure, where each node is a CF. As an alternative for storing the medoids, 
some methods, such as STREAM~\citep{OCallaghan:ICDE:2002:STREAM}, only store the clusters' centers. 
In competitive-learning stream algorithms, the centroids of the clusters evolve over time. Examples for such methods are SOStream~\citep{IsakssonMLDM:2012:SOStream}, 
DBSTREAM~\citep{Hahsler:JOSS:2017introduction}, evoStream~\citep{Matthias:BDR:2018:evoStream} and G-Stream~\citep{Ghesmoune:ICONIP:2014:G-Stream}.
\\\\
While
the algorithms described so far represent different approaches, what they all have in common
is that they are density-based. In another approach, based on partitioning the space using a grid, 
 the macro-clusters are usually found by grouping together adjacent grid cells.
In D-Stream~\citep{Chen:ACM-SIGKDD:2007:dstream}, a fixed grid is used. 
ExCC~\citep{Bhatnagar:DEXA:2007:ExCC} lets the user set the grid boundaries and number of cells.
Stats-Grid~\citep{Park:SIGMOD:2004:Stats-Grid} recursively splits the grid until the cells are sufficiently small.
Some methods (\eg,~\cite{Amini:TSWJ:2014:HDCStream}) combine the density- and grid-based approaches.
For a thorough study of the methods above and more, see a recent review by~\cite{Carnein:BISE:2019:review}.
\\\\
While most of the algorithms use the two-phase approach, there are several ones which are fully online. A prime example is Mini-Batch K-Means~\citep{Sculley:ICWWW:2010:batch-kmeans}, 
a very fast algorithm
which, having adapted the classical K-Means~\citep{Lloyd:PCM:1982:kmeans,Macqueen:1967:kmeans}
to a batched version, updates centroids using a step in a gradient-based direction 
(as opposed to a full recalculation as in K-Means). 
Adaptive Streaming K-Means~\citep{Puschmann:ITJ:2016:adaptive} is another online approach that aims at handling concept drifts.
Another online algorithm is pcStream~\citep{Mirsky:PACKDD:2015:pcstream} which uses Principal Component Analysis (PCA) to capture contexts in the data.
\\ \\
\textbf{Online DPMM:} While DPMM seems a natural candidate model for streaming-data clustering due to its innate ability
of adapting its complexity to the data, efficient and scalable DPMM inference remains a great challenge. Moreover, its applicability to streaming data is not trivial. 
There have been a few works which adapted the DPMM to an online setting. 
To the best of our knowledge, all of them rely on variational inference. This is a key difference from our sampling-based method. 
SoVB~\citep{Hoffman:JMLR:2013:sovb} provides a framework which can be applied to BNP models, the DPMM included. 
In that version the data is processed in batches, and the model is updated according to a chosen learning rate. While the process in SoVB does not need to revisit previous batches, that method makes the strong assumption that the statistics are similar across the batches; our method does not suffer from this limitation.
MemoizedVB~\citep{Hughes:NIPS:2013:memoizedDP} is an online variational inference method intended 
for large datasets where the entire data does not fit in the memory. 
It processes the data is processed in batches 
and store the sufficient statistics of each cluster
in each batch separately. MemoizedVB, however, must revisit each batch multiple times, 
so it is less relevant for streaming data. MVCL~\citep{yang:IEEE:2019:memorized} 
is a similar method which extends MemoizedVB to continual learning with multiple datasets. That method too, however, is inapplicable in a stream setting as it requires revisiting batches.
\cite{Lin:NIPS:2013:online} proposed a learning algorithm for DPMM that requires only a single pass on the data and thus, theoretically, can operate in a streaming setting, 
at least on stationary data; however, it cannot adapt to concept drifts. This is also a limitation
of the distributed method from~\citep{Campbell15:NIPS:2015:Streaming}.
\\\\
\cite{Campbell:NIPS:2013:DynamicMeans} propose a method that uses the Dependent Dirichlet Process~\citep{Maceachern:BSS:1999:dependent} 
(an extension of the Dirichlet Process that supports evolving mixture models) for handling batch sequential data of an unknown number of evolving clusters.
D-Means~\citep{Campbell:TPAMI:2018:DMeans} is a related BNP clustering algorithm for evolving linearly-separable spherical clusters based on small-variance asymptotic analysis. 
Unlike those two works above which are restricted to spherical Gaussian components,
our implementation supports full-covariance Gaussians as well as multinomial components.
More generally, our method supports any exponential family.  
Moreover, we are unaware of Dependent DPMM methods (let alone implementations) that scale to streaming data. 
\\ \\
\textbf{Scalable DPMM Samplers:} 
While all the online DPMM methods above are variational,
we leverage a fast sampling-based DPMM method and adapt it to streaming data.  
Concretely,~\cite{Chang:NIPS:2013:ParallelSamplerDP} proposed a DPMM sampler (for non-streaming data), which,
by using an augmented space and parallel sampling, allows for fast inference with convergence guarantees and that can perform large moves, escaping many (though not all) poor local maxima. 
A summary of their sampler is presented in~\autoref{sec:backgroud}. 
More recently, \cite{Dinari:CCGRID:2019:distributed} proposed a more efficient
and even faster implementation of Chang and Fisher's sampler.  
Our work is largely based on those two works, which we have adapted into a streaming setting. Particularly, we are unaware of implementations of either variational or sampling-based DPMM inference that are (in the non-streaming case) as fast as~\cite{Dinari:CCGRID:2019:distributed}, let alone ones that support full-covariance Gaussians and/or multinomials. Thus, that is the implementation we chose to modify. 
Importantly, while~\cite{Dinari:CCGRID:2019:distributed} is still not fast enough for streaming data, our proposed algorithmic and implementation changes eliminate this problem.

 \section{BACKGROUND}\label{sec:backgroud} 
For simplicity, 
our presentation below assumes that all the random vectors involved
have either a probability density function (pdf) or a probability mass function (pmf). 
One known DPMM construction is as follows: 
\begin{align}
& \bpi|\alpha \sim \mathrm{GEM}(\alpha), \label{Eqn:DPLatentVar:pi}\\
& \theta_k |H  \overset{\iid}{\sim} f_\theta(H), \qquad \forall k\in \set{1,2,\ldots}, \\
& z_i|\bpi \overset{\iid}{\sim} \mathrm{Cat}(\bpi), \qquad \forall i\in \set{1,2,\ldots,N}, \\
& \bx_i|z_i,\theta_{z_i} \sim f_\bx(\bx_i;\theta_{z_i}), \qquad\forall i\in \set{1,2,\ldots,N}\,. \label{Eqn:DPLatentVar:xi}
\end{align}
Here $\iid$ stands for independent and identically distributed, $H$ is the \emph{base measure}, $f_\theta$ is the pdf or pmf associated with $H$, 
the infinite-length vector $\bpi=(\pi_k)_{k=1}^\infty$ is drawn from the Griffiths-Engen-McCloskey stick-breaking process (GEM)~\citep{Pitman:Book:2002:Combinatorial} with a concentration parameter $\alpha>0$ (particularly, $\pi_k>0$ for every $k$ and $\sum_{k=1}^{\infty}\pi_k=1$) 
while $\theta_k$ is drawn from $f_\theta$. 
Each of the $N$ \iid  observations $(\bx_i)_{i=1}^N$ is generated 
by first drawing a label, $z_i\in \Zplus$, from $\bpi$ (\ie, $ \mathrm{Cat}$ is the categorical distribution), and then  $\bx_i$ is drawn from (a pdf or a pmf) $f_\bx$ parameterized by $\theta_{z_i}$. 
Loosely speaking, the DPMM entertains the notion of a mixture model of infinitely many components: 
\begin{align}
 \bx_i \overset{\iid}{\sim} \sum\nolimits_{k=1}^{\infty} \pi_k f_\bx(\bx_i;\theta_k) \, .
\end{align}
Each $f_\bx(\cdot;\theta_k)$ is called a \emph{component} and we make no distinction between a component, $f_\bx(\cdot,\theta_k)$, and its parameter, $\theta_k$. 
The so-called labels $(z_i)_{i=1}^N$ encode
the observation-to-component assignments. 
A cluster is a collection of points sharing a label; \ie, $\bx_i$ is in cluster $k$, denoted by $C_k$,
if and only if $z_i=k$.
Let (the random variable) $K$
be the number of unique labels: $K=|\set{k:z_i=k \text{ for some } i\in \set{1,\ldots,N}}|$; \ie, $K$ is also the number of clusters and is bounded above by $N$. 
Typically, and as assumed in this paper, $H$ is chosen such that $f_\theta$
will be a conjugate prior~\citep{Gelman:Book:2013:Bayesian} to $f_\bx$.
The latent variables here are $K$, $(\theta_k)_{k=1}^\infty$, $\bpi$, and $(z_i)_{k=1}^N$.
For more details (and other constructions), see~\cite{Sudderth:PhD:2006:GraphicalModels}.

We now briefly review a DPMM sampler proposed by~\cite{Chang:NIPS:2013:ParallelSamplerDP}.
It consists of a restricted Gibbs sampler~\citep{Robert:Book:2013:Monte}
and a split/merge framework which together form an ergodic Markov chain.
The operations in each step of that sampler are highly parallelizable. 
Of note, the splits and merges let the sampler make \emph{large moves}
along the (posterior) probability surface as in such operations multiple labels change
their label together to the same different label; this is in contrast to what happens, \eg, in methods that change each label separately from the others. We now describe the essential details. 

\textbf{The augmented space.}
The latent variables, 
$(\theta_k)_{k=1}^\infty$, $\bpi$, and $(z_i)_{k=1}^N$, 
are augmented with auxiliary variables. 
For each component $\theta_k$ two subcomponents are added, $\bar{\theta}_{k,1},\bar{\theta}_{k,2}$, with subcomponent weights $\bar{\bpi}_k=(\bar{\pi}_{k,1},\bar{\pi}_{k,2})$. 
Implicitly, this means that every cluster $C_k$ is  augmented with two subclusters, $\bar{C}_{k,1}$ and $\bar{C}_{k,2}$. 
For each cluster label $z_i$, an additional \emph{subcluster label}, $\bar{z}_i\in\set{1,2}$, is added; \ie, subcluster $\bar{C}_{k,1}\subset C_{k}$ consists of all the points in $C_k$ whose subcluster label is $1$ ($\bar{C}_{k,2}$, is defined similarly). 
The goal of this auxiliary two-component mixture
is to facilitate useful cluster splitting proposals 
(see below).  

\textbf{The restricted Gibbs sampler.}
This restricted sampler is not allowed to change (the current estimate of) $K$; 
rather, it can change only 
the parameters of the existing clusters and subclusters, and when sampling the labels,
it can assign an observation only to an existing cluster.
Note that for each instantiated component $k$, changing $\theta_k$, $\bar{\theta}_{k,1}$,
and  $\bar{\theta}_{k,2}$ is done using 
$p(\theta_k|C_k;H)$, $p(\bar{\theta}_{k,1}|\bar{C}_{k,1};H)$, and $p(\bar{\theta}_{k,2}|\bar{C}_{k,2};H)$, respectively, where the latter three are the conditional distributions of the cluster or subcluster parameters given the cluster or subclusters. 
For more details about the restricted Gibbs sampler, see our \textbf{appendix}.

\textbf{The split/merge framework.}
Splits and merges allow the sampler to change $K$ 
using the Metropolis-Hastings framework~\citep{Hastings:1970:MC}. 
Particularly, the auxiliary variables are used to propose splitting an existing 
cluster or merging two exiting ones. 
When a split is accepted, each of the newly-born
clusters is augmented with two new subclusters.
The Hastings ratio of a split 
is~\citep{Chang:NIPS:2013:ParallelSamplerDP}
\begin{align}
 \mathrm{H}_{\text{split}}=\frac{\alpha 
 \Gamma(N_{k,1})f_\bx(\bar{C}_{k,1};H)
 \Gamma(N_{k,2})f_\bx(\bar{C}_{{k},2};H)}{\Gamma(N_{k})f_\bx(C_k;H)}
 \label{Eqn:HastingsRatioSplit}
\end{align}
where $\Gamma$ is the Gamma function,
$N_k$, $N_{k,1}$ and $N_{k,2}$ are the number of points in
$C_k$, $\bar{C}_{k,1}$ and $\bar{C}_{k,2}$, respectively,
while $f_\bx(C_k;H)$, $f_\bx(\bar{C}_{k,1};H)$,
and $f_\bx(\bar{C}_{k,2};H)$ represent the \emph{marginal}
likelihood of $C_k$, $\bar{C}_{k,1}$ and $\bar{C}_{k,2}$ respectively. Concrete expressions for the marginal likelihood, in the case of Gaussian or Multinomial components
(the component types considered in our experiments) appear in our~\textbf{appendix}. 
Finally, a \emph{merge} proposal is based on taking two existing clusters
and proposing merging them into one. The corresponding Hastings ratio is
 $\mathrm{H}_{\text{merge}} =1 / \mathrm{H}_{\text{split}}$ where $\bar{C}_{k,1}$ and $\bar{C}_{k,2}$
  are replaced with the two clusters, and $C_k$ is replaced with the result of the merge.
For a derivation of these ratios, see~\cite{Chang:NIPS:2013:ParallelSamplerDP}.

 \begin{table*}[t]
    \caption{Comparing our method (ScStream) with BIRCH~\cite{Zhang:ACM:1996:birch}, CluStream~\cite{Aggarwal:VLDB:2003:CluStream}, D-Stream~\cite{Chen:ACM-SIGKDD:2007:dstream}, DBSTREAM~\citep{Hahsler:TKDE:2016:DBSTREAM}, StreamKM++~\citep{Ackermann:JEA:2012:streamkm++}, Mini Batch K-Means~\citep{Sculley:ICWWW:2010:batch-kmeans}, pcStream~\citep{Mirsky:PACKDD:2015:pcstream}, SoVB~\citep{Hoffman:JMLR:2013:sovb}. Also included is DPMM sampler~\citep{Dinari:CCGRID:2019:distributed}.  N/A indicates that a method did not scale enough  or lacks support for the data type.}
    \resizebox{\columnwidth*2}{!}{
    \setlength\tabcolsep{3pt}
    
    \begin{tabular}{@{}lllllllllll||l@{}}
        
        \toprule
                                                                            &                                                                                        & BIRCH                                                   & CluStream$^\dagger$                                                            & D-Stream                                                                   & DBSTREAM                                                           & StreamKM++$^\dagger$                                                        &\begin{tabular}[c]{@{}l@{}}Mini Batch\\ K-Means$^\dagger$\end{tabular}         & pcStream                                                       &SoVB                                                         & \begin{tabular}[c]{@{}l@{}}ScStream\\ (Ours)\end{tabular}                          & \begin{tabular}[c]{@{}l@{}}DPMM\\ Sampler\end{tabular}  \\ \midrule
        2D Gaussians                                                      & \begin{tabular}[c]{@{}l@{}}ARI:\\ NMI:\\ Purity:\\ F-Measure:\\ Full-NMI:\end{tabular} &\begin{tabular}[c]{@{}l@{}}$.81\pm.12$\\$.89\pm.04$\\$.83\pm.06$\\$.84\pm.10$\\N/A\end{tabular}&\begin{tabular}[c]{@{}l@{}}$.86\pm.11$\\$.94\pm.03$\\$\mathbf{.94\pm.03}$\\$.88\pm.09$\\N/A\end{tabular}&\begin{tabular}[c]{@{}l@{}}$.88\pm.16$\\$.94\pm.05$\\$.91\pm.09$\\$.90\pm.13$\\N/A\end{tabular}&\begin{tabular}[c]{@{}l@{}}$.90\pm.11$\\$.94\pm.04$\\$.91\pm.06$\\$.91\pm.09$\\N/A\end{tabular}&\begin{tabular}[c]{@{}l@{}}$.53\pm.11$\\$.71\pm.05$\\$.57\pm.05$\\$.61\pm.08$\\N/A\end{tabular}&\begin{tabular}[c]{@{}l@{}}$.82\pm.09$\\$.89\pm.03$\\$.83\pm.05$\\$.85\pm.08$\\$.48\pm.00$\end{tabular}&\begin{tabular}[c]{@{}l@{}}$.60\pm.12$\\$.76\pm.07$\\$.70\pm.08$\\$.66\pm.10$\\$.37$\end{tabular}&\begin{tabular}[c]{@{}l@{}}$.58\pm.11$\\$.75\pm.05$\\$.68\pm.06$\\$.65\pm.09$\\$.52$\end{tabular}&\begin{tabular}[c]{@{}l@{}}$\mathbf{.93\pm.08}$\\$\mathbf{.95\pm.03}$\\$.92\pm.05$\\$\mathbf{.94\pm.07}$\\$\mathbf{.68\pm.01}$\end{tabular}&\begin{tabular}[c]{@{}l@{}}$.92\pm.14$\\$.94\pm.10$\\$.91\pm.10$\\$.93\pm.11$\\N/A\end{tabular}  \\ \midrule
        CoverType                                                        & \begin{tabular}[c]{@{}l@{}}ARI:\\ NMI:\\ Purity:\\ F-Measure:\\ Full-NMI:\end{tabular} &\begin{tabular}[c]{@{}l@{}}$.07\pm.08$\\$.14\pm.09$\\$.66\pm.10$\\$.44\pm.10$\\N/A\end{tabular}&\begin{tabular}[c]{@{}l@{}}$.10\pm.07$\\$.19\pm.09$\\$.71\pm.11$\\$.33\pm.05$\\N/A\end{tabular}&\begin{tabular}[c]{@{}l@{}}$.07\pm.11$\\$.19\pm.11$\\$.70\pm.10$\\$.58\pm.14$\\N/A\end{tabular}&\begin{tabular}[c]{@{}l@{}}$.10\pm.13$\\$.18\pm.15$\\$.68\pm.11$\\$\mathbf{.60\pm.13}$\\N/A\end{tabular}&\begin{tabular}[c]{@{}l@{}}$.09\pm.09$\\$.15\pm.08$\\$.68\pm.11$\\$.42\pm.10$\\N/A\end{tabular}&\begin{tabular}[c]{@{}l@{}}$.07\pm.06$\\$.13\pm.06$\\$.66\pm.12$\\$.37\pm.06$\\$.06\pm.01$\end{tabular}&\begin{tabular}[c]{@{}l@{}}$.03\pm.02$\\$.20\pm.07$\\$\mathbf{.79\pm.08}$\\$.11\pm.05$\\$.08$\end{tabular}&\begin{tabular}[c]{@{}l@{}}$.10\pm.09$\\$.13\pm.10$\\$.66\pm.13$\\$.48\pm.08$\\$.01$\end{tabular}&\begin{tabular}[c]{@{}l@{}}$\mathbf{.15\pm.11}$\\$\mathbf{.21\pm.14}$\\$.71\pm.11$\\$.47\pm.08$\\$\mathbf{.13\pm.01}$\end{tabular}&\begin{tabular}[c]{@{}l@{}}$.10\pm.11$\\$.16\pm.12$\\$.67\pm.12$\\$.48\pm.09$\\N/A\end{tabular} \\ \midrule
        ImageNet100                                                      & \begin{tabular}[c]{@{}l@{}}ARI:\\ NMI:\\ Purity:\\ F-Measure:\\ Full-NMI:\end{tabular} &\begin{tabular}[c]{@{}l@{}}$.21\pm.11$\\$.35\pm.11$\\$.64\pm.12$\\$.39\pm.08$\\N/A\end{tabular}&\begin{tabular}[c]{@{}l@{}}$.30\pm.13$\\$.45\pm.09$\\$.75\pm.12$\\$.44\pm.10$\\N/A\end{tabular}&\begin{tabular}[c]{@{}l@{}}N/A\\N/A\\N/A\\N/A\\N/A\end{tabular}&\begin{tabular}[c]{@{}l@{}}$.13\pm.15$\\$.22\pm.17$\\$.43\pm.13$\\$.43\pm.09$\\N/A\end{tabular}&\begin{tabular}[c]{@{}l@{}}$.55\pm.15$\\$.62\pm.09$\\$\mathbf{.91\pm.06}$\\$.62\pm.14$\\N/A\end{tabular}&\begin{tabular}[c]{@{}l@{}}$.49\pm.17$\\$.58\pm.12$\\$.87\pm.09$\\$.57\pm.15$\\$\mathbf{.57\pm.02}$\end{tabular}&\begin{tabular}[c]{@{}l@{}}$.20\pm.09$\\$.33\pm.08$\\$.66\pm.10$\\$.33\pm.09$\\$.26$\end{tabular}&\begin{tabular}[c]{@{}l@{}}$.31\pm.18$\\$.45\pm.20$\\$.49\pm.13$\\$.55\pm.11$\\$.23$\end{tabular}&\begin{tabular}[c]{@{}l@{}}$\mathbf{.63\pm.19}$\\$\mathbf{.69\pm.15}$\\$.78\pm.14$\\$\mathbf{.73\pm.12}$\\$.48\pm.01$\end{tabular}&\begin{tabular}[c]{@{}l@{}}$.64\pm.28$\\$.72\pm.24$\\$.74\pm.22$\\$.76\pm.17$\\N/A\end{tabular}    \\ \midrule
        ImageNet1K                                                       & \begin{tabular}[c]{@{}l@{}}ARI:\\ NMI:\\ Purity:\\ F-Measure:\\ Full-NMI:\end{tabular} &\begin{tabular}[c]{@{}l@{}}N/A\\N/A\\N/A\\N/A\\N/A\end{tabular}&\begin{tabular}[c]{@{}l@{}}$.30\pm.14$\\$.45\pm.10$\\$.74\pm.14$\\$.44\pm.09$\\N/A\end{tabular}&\begin{tabular}[c]{@{}l@{}}N/A\\N/A\\N/A\\N/A\\N/A\end{tabular}&\begin{tabular}[c]{@{}l@{}}$.30\pm.16$\\$.40\pm.14$\\$.62\pm.13$\\$.48\pm.11$\\N/A\end{tabular}&\begin{tabular}[c]{@{}l@{}}N/A\\N/A\\N/A\\N/A\\N/A\end{tabular}&\begin{tabular}[c]{@{}l@{}}$.45\pm.12$\\$.59\pm.07$\\$\mathbf{.97\pm.03}$\\$.51\pm.12$\\$\mathbf{.63\pm.01}$\end{tabular}&\begin{tabular}[c]{@{}l@{}}$.19\pm.07$\\$.38\pm.06$\\$.76\pm.08$\\$.28\pm.09$\\$.30$\end{tabular}&\begin{tabular}[c]{@{}l@{}}$.00\pm.02$\\$.00\pm.02$\\$.25\pm.04$\\$.38\pm.04$\\$.00$\end{tabular}&\begin{tabular}[c]{@{}l@{}}$\mathbf{.62\pm.17}$\\$\mathbf{.68\pm.13}$\\$.78\pm.13$\\$\mathbf{.72\pm.12}$\\$.41\pm.02$\end{tabular}&\begin{tabular}[c]{@{}l@{}}N/A\\N/A\\N/A\\N/A\\N/A\end{tabular} \\ \midrule
        100D Multinomials                                                 & \begin{tabular}[c]{@{}l@{}}ARI:\\ NMI:\\ Purity:\\ F-Measure:\\ Full-NMI:\end{tabular} &\begin{tabular}[c]{@{}l@{}}N/A\\N/A\\N/A\\N/A\\N/A\end{tabular}&\begin{tabular}[c]{@{}l@{}}$.00\pm.01$\\$.11\pm.05$\\$.09\pm.03$\\$.04\pm.01$\\N/A\end{tabular}&\begin{tabular}[c]{@{}l@{}}N/A\\N/A\\N/A\\N/A\\N/A\end{tabular}&\begin{tabular}[c]{@{}l@{}}$.00\pm.00$\\$.00\pm.00$\\$.03\pm.00$\\$.04\pm.01$\\N/A\end{tabular}&\begin{tabular}[c]{@{}l@{}}$.34\pm.24$\\$.65\pm.16$\\$.53\pm.25$\\$.35\pm.24$\\N/A\end{tabular}&\begin{tabular}[c]{@{}l@{}}$.41\pm.24$\\$.69\pm.16$\\$.61\pm.25$\\$.42\pm.24$\\$.54\pm.01$\end{tabular}&\begin{tabular}[c]{@{}l@{}}N/A\\N/A\\N/A\\N/A\\N/A\end{tabular}&\begin{tabular}[c]{@{}l@{}}$.21\pm.14$\\$.52\pm.14$\\$.31\pm.15$\\$.23\pm.13$\\$.27$\end{tabular}&\begin{tabular}[c]{@{}l@{}}$\mathbf{.78\pm.24}$\\$\mathbf{.89\pm.12}$\\$\mathbf{.84\pm.20}$\\$\mathbf{.78\pm.24}$\\$\mathbf{.72\pm.01}$\end{tabular}&\begin{tabular}[c]{@{}l@{}}$.45\pm.22$\\$.62\pm.30$\\$.53\pm.25$\\$.46\pm.22$\\N/A\end{tabular}    \\ \midrule
        20NewsGroup                                                      & \begin{tabular}[c]{@{}l@{}}ARI:\\ NMI:\\ Purity:\\ F-Measure:\\ Full-NMI:\end{tabular} &\begin{tabular}[c]{@{}l@{}}N/A\\N/A\\N/A\\N/A\\N/A\end{tabular}&\begin{tabular}[c]{@{}l@{}}$.00\pm.00$\\$.12\pm.02$\\$.13\pm.02$\\$.10\pm.00$\\N/A\end{tabular}&\begin{tabular}[c]{@{}l@{}}N/A\\N/A\\N/A\\N/A\\N/A\end{tabular}&\begin{tabular}[c]{@{}l@{}}N/A\\N/A\\N/A\\N/A\\N/A\end{tabular}&\begin{tabular}[c]{@{}l@{}}$.01\pm.00$\\$.07\pm.01$\\$.11\pm.01$\\$.10\pm.00$\\N/A\end{tabular}&\begin{tabular}[c]{@{}l@{}}$.01\pm.00$\\$.09\pm.01$\\$.12\pm.01$\\$.09\pm.00$\\$.05\pm.01$\end{tabular}&\begin{tabular}[c]{@{}l@{}}N/A\\N/A\\N/A\\N/A\\N/A\end{tabular}&\begin{tabular}[c]{@{}l@{}}$.06\pm.01$\\$.20\pm.02$\\$.13\pm.01$\\$.14\pm.01$\\$.17$\end{tabular}&\begin{tabular}[c]{@{}l@{}}$\mathbf{.13\pm.01}$\\$\mathbf{.36\pm.03}$\\$\mathbf{.28\pm.02}$\\$\mathbf{.20\pm.01}$\\$\mathbf{.32\pm0.03}$\end{tabular}&\begin{tabular}[c]{@{}l@{}}$.12\pm.01$\\$.33\pm.02$\\$.24\pm.02$\\$.19\pm.01$\\N/A\end{tabular}   \\ \bottomrule
        \multicolumn{4}{l}{\small $^\dagger$ Parametric methods given the true $K$.} \\ 
     \end{tabular}
    }
    \label{tab:results}
    \vspace*{-0.2cm}
    \end{table*}

  \section{METHOD}
Henceforth we will refer to the sampler from~\autoref{sec:backgroud}
as the DPMM sampler. 
We base our method on that sampler but introduce important modifications of the latter in order to: 1) make it compatible with streaming data that arrives (possibly rapidly and/or indefinitely) batch by batch; 2) improve the clustering  and the label consistency across batches. 

\subsection{Batches and Time-based Weighting}\label{Sec:Method:Batches}
For each instantiated component $k$ in the DPMM sampler,  
finding $p(\theta_k|C_k;H)$ requires computing
sufficient statistics based on \emph{all} of the points in $C_k$
(similar logic applies to $(p(\bar{\theta}_{k,j}|\bar{C}_{k,j};H))_{j\in\set{1,2}}$). 
As the sufficient statistics are based on summation, it is tempting to try to use online updates by incrementally increasing the sums whenever a new batch arrives. 
However, the sufficient statistics often change with each iteration 
of the DPMM sampler (due to label changes). Thus, the online updates cannot be done without having all the batches that arrived so far stored in memory -- but that becomes infeasible quickly as the stream grows. More generally,
once a batch is processed it cannot be revisited again. 
Moreover, even if somehow, via an expensive and tedious bookkeeping, one could 
keep track of the ever-changing sufficient statistics, this would usually be a bad idea: 
once enough batches arrive new ones will hardly influence the sufficient statistics despite the fact that they, in the common case of non-stationarity, are more relevant than much older batches.

We address the no-revisits constraint and the non-stationarity as follows. 
Let $X_B=(\bx_i)_{i=1}^{n_B}$ be a data batch at time $B$ 
where $n_B$ is the number of points in it.   
For $b\in\set{1,\ldots,B}$, 
let $n^b_k$ be the number of points in batch $b$ assigned
to cluster $k$, and let $s^b_k$ be the  sufficient statistics computed, for cluster $k$, based solely on those $n^b_k$ points. 
Let $h^{1:B}_k$ denote the history record of sufficient statistics and counts for cluster $k$:  
$
    h^{1:B}_k = ((s^b_k,n^b_k))_{b=1}^B\ 
    $.
The subcluster history is similarly defined; \ie, 
$\bar{h}^{1:B}_{k,j}=((\bar{s}^b_{k,j},\bar{n}^b_{k,j}))_{b=1}^B$
where $j\in\set{1,2}$ and $\ \bar{ }\ $ indicates subcluster-related quantities.
Note that while per-batch sufficient statistics were also used in~\cite{Hughes:NIPS:2013:memoizedDP}
their method is inapplicable for streaming data as they must revisit batches. 
We now define the time-weighted sufficient statistics and count for cluster $k$: 
\begin{align}
\hspace{-.1cm}
    &S^B_k = \sum\nolimits_{b={q}}^{B} \Kcal(B,b) s^{b}_k
\,,\,
    N^B_k = \sum\nolimits_{b={q}}^{B} \Kcal(B,b) n^{b}_k \, 
    \label{eqn:suff_finite}
\end{align}
where $\Kcal(\cdot,\cdot):\RR\times \RR\to \Rnonneg$ is a weighting function, ${q}=\min$ $\set{b:b\in\set{1,\ldots,B}, \Kcal(B,b)>\epsilon}$,
and $\epsilon>0$ is a user-defined threshold (we used $\epsilon=1e-08$ in all our experiments).  
The analogous subcluster quantities, (\ie, $\bar{S}^b_{k,1},\bar{S}^b_{k,2}$, $\bar{N}^b_{k,1}$, and $\bar{N}^b_{k,2}$)
are defined similarly. 
Note that, in the summations above,  usually $q>1$.  
This limits the space/memory/time complexity, and implies we need to maintain the history records
only up to a fixed maximal length; \ie, for each batch $B$ we update these records such that we keep the information from a previous batch $b$ only if $\Kcal(B,b)>\epsilon$  and discard it otherwise.
As in many damped-window methods, we use $\Kcal(B,b) =
2^{-\lambda (B-b)}$ where $\lambda>0$ is user-defined (that said, other kernels 
may also be used).
\begin{Example}\label{Example:GaussiansSuff}
Consider $D$-dimensional Gaussian components where the \textit{Normal Inverse Wishart} (NIW) distribution
serves as the base measure. Let $(\kappa,\bm,\nu,\bPsi)$ denote the hyperparameters of the NIW prior. 
Let $X_b=(\bx_1,\dots,\bx_{n_b})$ denote the data points (in $\RD$) in batch $b$. 
Using a classical result~\citep{Gelman:Book:2013:Bayesian}, the sufficient statistics here are 
    \begin{align}
        s^b_k=\left(\sum\nolimits_{\bx_i \in X_b}\bx_i \indicator_{z_i=k},\sum\nolimits_{\bx_i \in X_b}\bx_i \bx_i^T\indicator_{z_i=k}\right)\ ,
    \end{align}
    where the indicator function $\indicator_{z_i=k}$ is 1 if $z_i=k$ and 0 otherwise. 
    Using conjugacy~\citep{Gelman:Book:2013:Bayesian} 
    as well as the replacement 
    of the standard sufficient statistics and counts
    with their weighted versions, 
    the hyperparameters of the NIW posterior for cluster $k$ are:  
    \begin{align}
        &\kappa^{*}_k = \kappa + N^B_k\ , \quad \nu^{*}_k = \nu + N^B_k\ , 
        \nonumber \\
        &\bm^{*}_k = \frac{1}{\kappa^{*}_k}\left(\kappa\bm  + \sum\nolimits_{b={q}}^B 
        \left[\Kcal(B,b) \sum\nolimits_{\bx_i \in X_b}\bx_i\indicator_{z_i=k}\right]\right) \ , 
        \nonumber
        \\
        &\bPsi^{*}_k = \frac{1}{\nu^{*}_k}
        \left(\nu\bPsi + \sum\nolimits_{b={q}}^B \left[ \Kcal(B,b)\sum\nolimits_{\bx_i \in X_b}\bx_i\bx_i^T\indicator_{z_i=k}\right]
        \right)\,  \label{Eqn:GaussianUpdate}
    \end{align}
    (note that together, the two nested sums in these equations constitute $S^B_k$). 
    This fully defines the posterior distribution over the parameters of Gaussian $k$.
\end{Example}
See the \textbf{appendix} for an analogous example for the case of multinomial components. 
More generally, out method applies to components from any exponential family 
when used with its conjugate prior. 
As we will show in~\autoref{Sec:Method:Subsec:Alg},
to determine $z_i$, our method uses the predictive posterior distribution, 
$p(\bx_i|H,S^B_k,N^B_k,z_i=k)$ (where we replaced the standard sufficient statistics and number of points
with their weighted versions) as the latter induces, via proportionality, 
the predicted probability of  observation $\bx_i$
    to belong to cluster $k$: 
    \begin{align}\label{Eq:AssignmentByPredictive}
    \hspace{-.20cm}
        p(z_i\hspace{-.09cm} =\hspace{-.09cm} k|\bx_i,H,S^B_k,N^B_k)
    \hspace{-.06cm}
    \propto \hspace{-.06cm}
    p(\bx_i|H,S^B_k,N^B_k,z_i\hspace{-.08cm} =\hspace{-.08cm} k)\, .\hspace{-.02cm} 
\end{align}
\begin{Example}
    Continuing~\autoref{Example:GaussiansSuff},
    in the Gaussian case with an NIW prior, 
    \EQN\eqref{Eq:AssignmentByPredictive} becomes~\citep{Chang:Thesis:2014:sampling} 
   \begin{align}
    &\hspace{-.1cm}p(z_i\hspace{-.08cm} =\hspace{-.08cm}k|\bx_i,H,S^B_k,N^B_k)
    \hspace{-.06cm}
    \propto \hspace{-.06cm}
      \pi_k
        \overbrace{p(\bx_i|\kappa^{*}_k,\bm^{*}_k,\nu^{*}_k,\bPsi^{*}_k,z_i\hspace{-.08cm} =\hspace{-.08cm}k)}^{p(\bx_i|H,S^B_k,N^B_k,
        z_i=k)}
\nonumber  \\ 
         &= t_{\nu^{*}_k-D+1}\left(\bx_i;\bm^{*}_k,\frac{\kappa^{*}_k+1}{\kappa^{*}_k(\nu^{*}_k-D+1)}\nu^{*}_k\bPsi^{*}_k\right)\ 
        \label{eqn:posterior}
    \end{align}
    where $t(\cdot)$ is Student's t-distribution (see the \textbf{appendix} for the multinomial case).
\end{Example}
\subsection{The Proposed Algorithm}\label{Sec:Method:Subsec:Alg}
Before diving into the proposed algorithm,  
which is a novel extension of the DPMM sampler, 
let us consider the underlying model behind it. 
In the stationary case where all the data points are drawn from the same DPMM, the algorithm 
almost coincides with the DPMM sampler (especially if $\lambda$ is high). 
A more interesting insight is the following. 
\cite{Hu:CEMNLP:2015:tsdpmm} showed how to incorporate prior knowledge into a DPMM. 
When our method processes batch $B$, 
it can be viewed as DPMM inference with such an incorporated prior knowledge. 
Specifically, the prior knowledge consists of the $K$ instantiated clusters where the prior knowledge of 
each $\theta_k$ ($k\in\set{1,\ldots,K}$) is captured via the posterior distribution over $\theta_k$ 
implied by $H$ and $h^{1:B-1}_k$. 

With this in mind, we proceed to describe the proposed algorithm, summarized in~\autoref{alg:streaming_dpmm}
(which uses~\autoref{alg:restricted_iter} as its main subroutine). 
 Let $\bX=(X_B)_{B\in\set{1,2,\ldots}}$ be a possibly-infinite data stream, where each $X_B$ is a batch of $n_B$ data points.
 One of our proposed modifications is that throughout the entire run of the algorithm,
 instead of using the standard sufficient statistics and point number, we use \EQN\ref{eqn:suff_finite} which in turn is based on the history records from~\autoref{Sec:Method:Batches}. 

 Upon the arrival of $X_B$, we run the restricted Gibbs sampler on it for $T$ iterations, 
 where each iteration is followed by allowing splits/merges as in the DPMM sampler. If $B=1$,
 we set $T=\infty$, meaning we run it till convergence (which is~\emph{very} fast
 since batches are small (\eg, $n_B=10^3$). If $B>1$ then we use $T=1$ (\ie, a single iteration).
 Either way, after those $T$ iterations,
 we perform an additional iteration (again followed by proposing and accepting/rejecting
 splits and merges stochastically), but this time 
 replace the restricted Gibbs sampler  with a deterministic routine 
 (lines 6--11 in~\autoref{alg:restricted_iter}) 
 based on 1) modes (namely, the argmax of the relevant distributions) instead of sampling, and, more importantly, 2) 
 the predictive posterior distributions
 of the labels and subcluster labels. 

  {    \SetKwComment{Comment}{}{}
    \IncMargin{1.0em}
    \begin{algorithm}[t]
    \KwIn{$H$, $\alpha$, $\Kcal$, $\epsilon$,$T$\\}
    \KwData{Stream $\bX$}
    \DontPrintSemicolon
    $X_1\leftarrow \bX.next$ \\
    $C_1\leftarrow X_1$ \\
    $K\leftarrow 1$ \\
    Randomly partition $C_1$ into subclusters $C_{1,1}$ and $C_{1,2}$\\ 
    ${q}\leftarrow 1$  \\
    Extract $h^{1:1}_1=(s^1_1,n^1_1)$,
    $\bar{h}^{1:1}_{1,1}=(\bar{s}^1_{1,1},\bar{n}^1_{1,1})$ and
    $\bar{h}^{1:1}_{1,2}=(\bar{s}^1_{1,2},\bar{n}^1_{1,2})$
  from $(C_1,C_{1,1},C_{1,2})$ \\
    $\Mcal\leftarrow 
    (h^{1:1}_1,\bar{h}^{1:1}_{1,1},\bar{h}^{1:1}_{1,2})$
    \\
    \While{Not Converged}
    {    
    $K,\Mcal\leftarrow $\autoref{alg:restricted_iter}$(X_1;H,\alpha,K,\Kcal,\mathit{\infty},\_,
    {q},B,
    \Mcal)$\\
    }
    \While{$X_B \leftarrow \bX.next$} { 
      $(h^{{q}:(B-1)}_k,\bar{h}^{{q}:(B-1)}_{k,1},\bar{h}^{{q}:(B-1)}_{k,2})_{k=1}^{K}
      \leftarrow \Mcal$ \\
               ${q}\leftarrow \min\set{b:b\in\set{1,\ldots,B}, \Kcal(B,b)>\epsilon}$ \\ 
      $\Mcal \leftarrow (h^{{q}:B-1}_k,\bar{h}^{{q}:B-1}_{k,1},\bar{h}^{{q}:B-1}_{k,2})_{k=1}^{K}$ \\
      \For{$t=1:T+1$}
      {
        $\hspace*{-0.1cm}K,\Mcal\leftarrow $\autoref{alg:restricted_iter}$(X_B;H,\alpha,\ldots ,t,
        {q},B,\Mcal)$ \\
      }
      Yield $\Mcal$
    }
    \caption{ScStream}\label{alg:streaming_dpmm}
    \end{algorithm}
    }

  \IncMargin{0.1em}
{    \SetKwComment{Comment}{}{}
    \begin{algorithm}[t]
    \KwIn{$H$, $\alpha$, $K$, $\Kcal$,$T$,$t$,
            ${q}$,$B$, $\Mcal=(h^{{q}:B}_k,(\bar{h}^{{q}:B}_{k,j})_{j\in\set{1,2}})_{k=1}^K$}
    \KwOut{$K'$,$\Mcal'$}
    \KwData{$X_B$}
    \DontPrintSemicolon
    \If{$t < T+1$ }
    {
      $(h^{{q}:B}_1,\bar{h}^{{q}:B}_{1,1},\bar{h}^{{q}:B}_{1,2})\leftarrow \Mcal$ \\
      Compute $(S^B_k)_{k=1}^K$ and $(N^B_k)_{k=1}^K$ using \EQN\eqref{eqn:suff_finite} \\ 
        1 iteration of the restricted sampler
      from~\autoref{sec:backgroud} using 
      $(S^B_k)_{k=1}^K$ and $(N^B_k)_{k=1}^K$ (see \textbf{appendix} for details)
      \\
    }
    \Else{
                    $\bpi \leftarrow \left(\tfrac{N^{B}_1}{\sum\limits_{k=1}^K N^{B}_{k} +\alpha}
                    ,\ldots, \tfrac{N^{B}_K}{\sum\limits_{k=1}^K N^{B}_{k} +\alpha},
                    \tfrac{\alpha}{\sum\limits_{k=1}^K N^{B}_{k} +\alpha}\right)$
          \\
      \For{$k\in\set{1,\ldots,K}$}
      {
        $\bar{\bpi}_k \leftarrow \left(\frac{\frac{\alpha}{2}+\bar{N}^B_{k,1}}{\alpha + \sum_{s=\set{1,2}} \bar{N}^B_{k,s}},\frac{\frac{\alpha}{2}+\bar{N}^B_{k,2}}{\alpha + \sum_{s=\{1,2\}} \bar{N}^B_{k,s}}\right)$\\
      }
       \For {$x_i\in X_B$}
       {$z_i\leftarrow \underset{k\in\{1,\ldots,K\}}{\text{arg max}} \pi_k p(z_i=k|\bx_i,H,S^B_k,N^B_k)$ 
       \\
      $\bar{z}_i\leftarrow \underset{j\in\{1,2\}}{\text{arg max}} \bar{\pi}_{z_i} p(\bar{z}_i=j|\bx_i,H,\bar{S}^B_{z_i,j},\bar{N}^B_{z_i,j})$}
     }
    \For{$k\in\{1,\ldots,K\}$}
    {
          Extract $(s^B_k,n^B_k)$,
    $(\bar{s}^B_{k,1},\bar{n}^B_{k,1})$ and
    $(\bar{s}^B_{k,2},\bar{n}^B_{k,2})$
  (from $C_k$, $\bar{C}_{k,1}$ and $\bar{C}_{k,2}$, respectively)
  and update 
      $(h^{{q}:B}_k,\bar{h}^{{q}:B}_{k,1},\bar{h}^{{q}:B}_{k,2})$
      accordingly  
    }
    \For{$k\in\set{1,\ldots,K}$}
    {
      Propose splitting $C_k$ to its subsclusters and accept the split with probability $\min(1,H_{\text{split}})$ (\EQN\eqref{Eqn:HastingsRatioSplit})
    }
    \For{$k,k'\in\set{1,\ldots,K}$}
    {
      Propose merging $C_k$ and $C_{k'}$ and accept the merge with probability $\min(1,H_{\text{merge}})$ 
    }
    $\Mcal'\leftarrow  (h^{{q}:B}_k,(\bar{h}^{{q}:B}_{k,j})_{j\in\set{1,2}})_{k=1}^{K'}$ where $K'$ is the new number of clusters\\
    \caption{Iteration of the Modified DPMM Sampler}\label{alg:restricted_iter}
    \end{algorithm}
    }

 Concretely, Chang and Fisher III's restricted sampler (see~\autoref{sec:backgroud})
 determines the labels as follows:
it draws $\bpi$ and $(\bar{\bpi}_k)_{k=1}^K$ from their respective conditional Dirichlet distributions,
and, for each (instantiated) component $k$, 
draws 
 $\theta_k\sim p(\theta_k|C_k;H)$,
 $\bar{\theta}_{k,1}\sim p(\bar{\theta}_{k,1}|\bar{C}_{k,1};H)$,
 and 
  $\bar{\theta}_{k,2}\sim p(\bar{\theta}_{k,2}|\bar{C}_{k,2};H)$.
  Next, it uses these drawn parameters and weights
  to construct, for each $\bx_i$, a likelihood-based pmf and then draws $z_i$ from it:
  \begin{align}
   z_i \appropto \pi_k f_\bx(\bx_i;\theta_k,z_i=k) \,    
  \end{align}
  (where $\appropto$ denotes sampling proportional to the right-hand side of the equation). 
In contrast, our deterministic subroutine: 
1) updates $\bpi$ and $(\bar{\bpi}_k)_{k=1}^K$ using the mode of each of their respective conditional Dirichlet distributions; 
2) avoids sampling $(\theta_k,\bar{\theta}_{k,1},\bar{\theta}_{k,2})$ and, instead of computing a likelihood-based pmf, 
computes a pmf based on the predictive posterior distribution
 (\EQN~\eqref{Eq:AssignmentByPredictive}) 
 and uses its argmax to determine $z_i$:
 \begin{align}
  z_i = \argmax{k\in \set{1,\ldots,K}}\pi_k p(\bx_i|H,S^B_k,N^B_k,z_i=k) \ .
 \end{align}
 
While the transition from sampling to argmax only slightly improves the stability of the results, 
the use of the predictive posterior leads to usually-drastic improvements in the quality and the inter-batch consistency
of the predicted labels (see~\autoref{Sec:Results}).
The high label consistency of our method is in sharp contrast to many streaming-data clustering methods
(especially those with an offline reclustering step) which usually suffer from significant label switching.

Unsurprisingly,
the running time grows linearly with $T$. Thus, the choice of $T$ seemingly suggests a trade-off between performance and speed.
However, we empirically found (see the~\textbf{appendix}) that \emph{the improvement in performance is only sublinear}
and that, in practice, a \emph{very} low value of $T$ usually suffices. Thus, we suggest using $T=1$. 
That said, if the stream is slow enough, one may benefit from using a larger $T$. 

\textbf{Remark.}   It is, in fact, possible to use $T=1$ even on $X_1$ since even in this case, as the stream progresses the method will eventually reach good performance. However, this will hurt the results in (only) the first few batches of the stream. 

\textbf{Summarizing the Main Differences from the DPMM Sampler.}
 First and foremost, our use of (a finite-length history of) weighted batched sufficient statistics is key here. 
 Apart from the fact that it allows us to handle concept drifts gracefully (unlike the original sampler,
 which assumes that the data statistics are stationary) it also allows us to visit each batch only once.
 The second main difference is our deterministic subroutine:
 although it plays a less important role than the first change
 (as we show in~\autoref{Sec:Results}, even without that subroutine our method already achieves SOTA
 results) it allows our method to have consistent labels across batches, 
 a feat most methods are incapable of.

\begin{table*}[t]
    \captionsetup{justification=centering, singlelinecheck=false}
    \caption{Running time (in seconds)}
    \resizebox{\columnwidth*2}{!}{
    \setlength\tabcolsep{3pt}
    
    \begin{tabular}{@{}llllllllll||l@{}}
        
        \toprule
                                                                       & BIRCH                                                   & CluStream                                                            & D-Stream                                                                   & DBSTREAM                                                           & StreamKM++                                                        &\begin{tabular}[c]{@{}l@{}}Mini Batch\\ K-Means\end{tabular}         & pcStream                                                       &SoVB                                                         & \begin{tabular}[c]{@{}l@{}}ScStream\\ (Ours)\end{tabular}                          & \begin{tabular}[c]{@{}l@{}}DPMM\\ Sampler\end{tabular} \\ \midrule
        2D Gaussians                                                      & 112.5                      & 31.3                               &24.7                                &17.0                              & 15.4                                & 1.4                                   & 1020.7                            & 53.3                         & 22.9 & 589.5                  \\
        CoverType                                                        & 95.8  & 45.9 & 1723.8 & 12.1 & 25.5 & 0.8 & 1610.3& 115.6 & 6.1 & 254.8 \\
        ImageNet100                                                      & 57.9 & 66.7& N/A & 65.2 & 242.7 & 12.0 & 15.7 & 100.5 & 23.1 & 1039.9 \\
        ImageNet1K                                                       & N/A & 1454 & N/A & 814 & N/A & 148 & 195 & 9219 & 1005 & N/A \\
        100D Multinomials                                                 & N/A & 44.7 & N/A & 12.9 & 25.5 & 0.8 & N/A & 115.6 & 23.5 & 254.8 \\
        20NewsGroup                                                      & N/A & 71.9 & N/A & N/A & 61.1&0.2 & N/A & 3.1 & 12.7 & 122.6 \\ \bottomrule
        \end{tabular}
    }
    \label{tab:timing}
    \end{table*}

\section{RESULTS}\label{Sec:Results}
For evaluation, we have chosen several datasets of varying difficulty levels 
and compared with multiple methods. Some of these datasets are known while the ones we created/modified are accessible
from our code repository. 
We chose the component type in our method according to the data type:
Gaussian components for $\RD$-valued data and multinomial components for discrete count data.
As in~\cite{Dinari:CCGRID:2019:distributed}, our code can run on either a single multithreaded process,
or be distributed across processes and/or machines. Here, in our experiments, due to the small batch size we have chosen the former configuration.
 
 \textbf{Methods.} Due to the abundance of existing algorithms, 
 we focused on several popular methods for  clustering streaming data:
 BIRCH~\citep{Zhang:ACM:1996:birch}, CluStream~\citep{Aggarwal:VLDB:2003:CluStream}, D-Stream~\citep{Chen:ACM-SIGKDD:2007:dstream},
 DBSTREAM~\citep{Hahsler:TKDE:2016:DBSTREAM}, StreamKM++~\citep{Ackermann:JEA:2012:streamkm++}, Mini-Batch K-Means~\citep{Sculley:ICWWW:2010:batch-kmeans}
 and pcStream~\citep{Mirsky:PACKDD:2015:pcstream}.
 That choice was based on their problem settings, popularity, 
 and available software (taken from~\cite{Hahsler:JOSS:2017introduction,Bifet:JCMKDD:2011:moa,Pedregosa:JMLR:2011:scikit-learn}).
 In addition, we have compared with SoVB~\citep{Hoffman:JMLR:2013:sovb}, as implemented by~\cite{Hughes:NIPS:2014:bnpy}. 
 We also compare our method with the DPMM sampler~\citep{Chang:NIPS:2013:ParallelSamplerDP}, using its faster reimplementation~\citep{Dinari:CCGRID:2019:distributed}.
For fairness of comparison, we have tuned each of the methods on each of the datasets using available black-box optimizers.
The parameters of each method were tuned on the first 10 batches of each stream, 
setting the mean adjusted rand index (ARI) as the objective function. The R-based methods were tuned using
the \textit{irace} R package~\citep{Lopez:ORP:2016:irace}. 
The tuning for pcStream~\citep{Mirsky:PACKDD:2015:pcstream} was done using the \textit{black-box}~\citep{Knysh:arxiv:2016:blackbox} Python package.
For the DPMM-related methods we tuned the hyperparameters using the \textit{BlackBoxOptim} Julia package.
Note that some of the methods (BIRCH; CluStream; Mini-Batch Kmeans; StreamKM) are parametric,
\emph{thus they were always provided with the true number of clusters}, 
while others (the DPMM sampler; DBSTREAM; D-Stream; pcStream; ours) infer the number of clusters; 
this gives an unfair advantage to the former over the latter.

\textbf{Datasets with points in $\RD$.}
We created a synthetic dataset 
with $10^7$ points in $\Rtwo$ divided into 20 clusters, where the points of each cluster were drawn from a different Gaussian. Next, we have inserted an incremental concept drift~\citep{RamrezGallego:NC:2017:ASO} to that dataset, meaning that the clusters moved, independently of each other, as the stream progressed.
 In addition, we have used two real datasets: CoverType~\citep{Blackard:AGRI:1999:forest}
 (which is often used for evaluating streaming-data clustering methods), 
 in which each observation provides a point in $\RR^{10}$ (describing $30$ squared meters of forest).
 The second real dataset we used is ImageNet's~\citep{Deng:CVPR:2009:imagenet} train set, 
 where we initially extracted features using SWAV~\citep{Caron:NIPS:2020:unsupervised},
 and then used them in two different settings. In the first one we have used a subset of 100 classes (out of 1000) and used PCA to 
 reduce the dimension to 64. 
 In the second setting we have used the full dataset, and used PCA to reduce the dimension to 128. 
In both of the ImageNet settings we have added a recurring concept drift~\citep{RamrezGallego:NC:2017:ASO}.
In all of the three real dataset settings we have normalized the data by subtracting the mean of each feature and dividing by its standard deviation.

\textbf{Datasets with count data.}
We have created a synthetic dataset of $10^7$ 100-dimensional points divided
into 100 clusters, where the points in each cluster were drawn from a multinomial distribution. 
Next, we have inserted gradual concept drift~\citep{RamrezGallego:NC:2017:ASO} to that dataset.
For a real dataset we have used the 20newsgroup~\citep{Lang95},
where each sample is a news article in one out of twenty possible subjects. We have used only the 1000 most commons words for classification in each sample.

Note that not all the methods were evaluated on all of the datasets. The reason is that some methods do not support count data and/or do not scale (\eg did not finish in comparable time or ran out of memory) to high dimensions.
\begin{figure*}[t!]
    \centering
    \newcommand{\MySpace}{\vspace{-.15cm}}
    \newcommand{\MyHeight}{1.5cm}
    \newcommand{\MyWidth}{0.15\linewidth}
    \newcommand{\MySmallWidth}{0.008\linewidth}
    \vspace{-.2cm}
    \begin{subfigure}{\MySmallWidth}\rotatebox{90}{\scriptsize Original}\end{subfigure}
    \begin{subfigure}{\MyWidth}\includegraphics[height=\MyHeight]{../figures/video_seg/original/00001.png}\end{subfigure}
    \begin{subfigure}{\MyWidth}\includegraphics[height=\MyHeight]{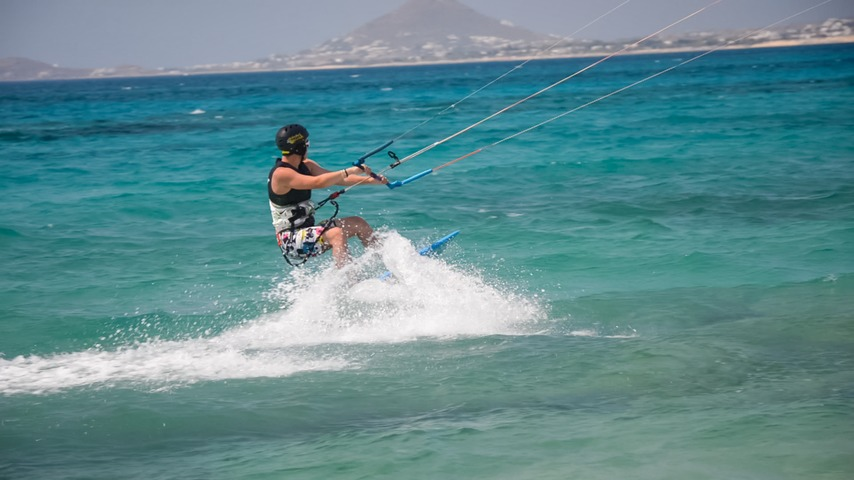}\end{subfigure}
    \begin{subfigure}{\MyWidth}\includegraphics[height=\MyHeight]{../figures/video_seg/original/00026.png}\end{subfigure}
    \begin{subfigure}{\MyWidth}\includegraphics[height=\MyHeight]{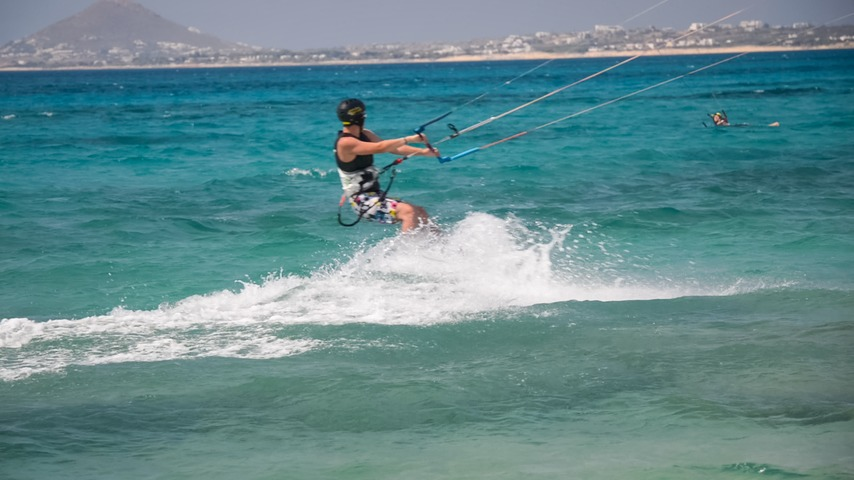}\end{subfigure}
    \begin{subfigure}{\MyWidth}\includegraphics[height=\MyHeight]{../figures/video_seg/original/00039.png}\end{subfigure}
    \begin{subfigure}{\MyWidth}\includegraphics[height=\MyHeight]{../figures/video_seg/original/00049.png}\end{subfigure}\\
    \begin{subfigure}{\MySmallWidth}\rotatebox{90}{\scriptsize ScStream}\end{subfigure}
    \begin{subfigure}{\MyWidth}\includegraphics[height=\MyHeight]{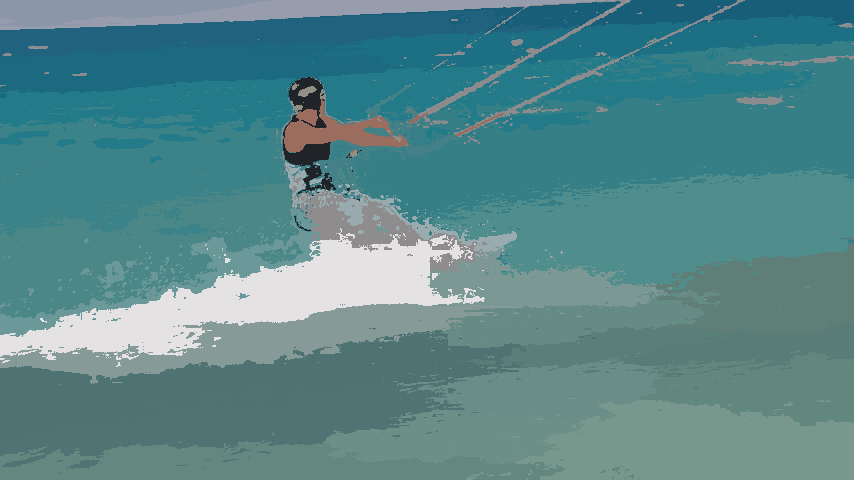}\end{subfigure}
    \begin{subfigure}{\MyWidth}\includegraphics[height=\MyHeight]{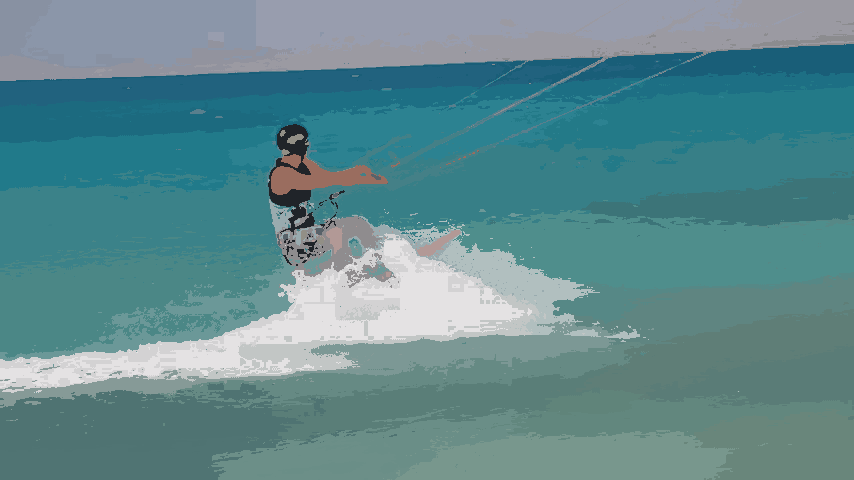}\end{subfigure}
    \begin{subfigure}{\MyWidth}\includegraphics[height=\MyHeight]{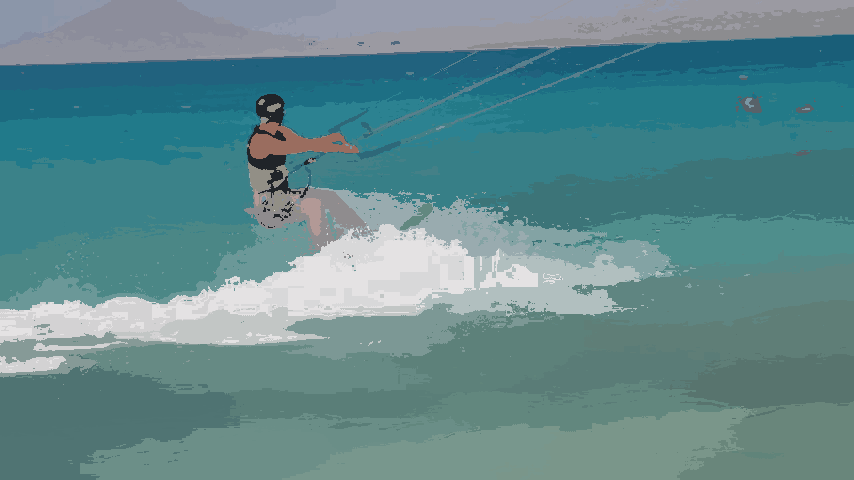}\end{subfigure}
    \begin{subfigure}{\MyWidth}\includegraphics[height=\MyHeight]{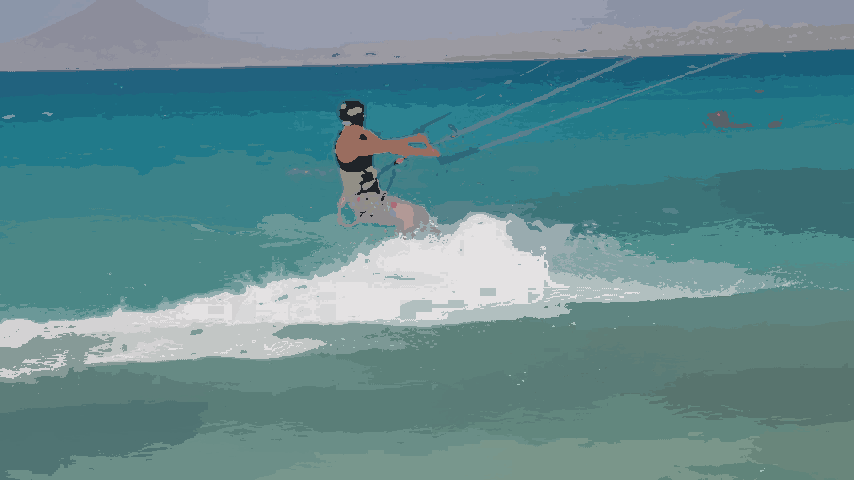}\end{subfigure}
    \begin{subfigure}{\MyWidth}\includegraphics[height=\MyHeight]{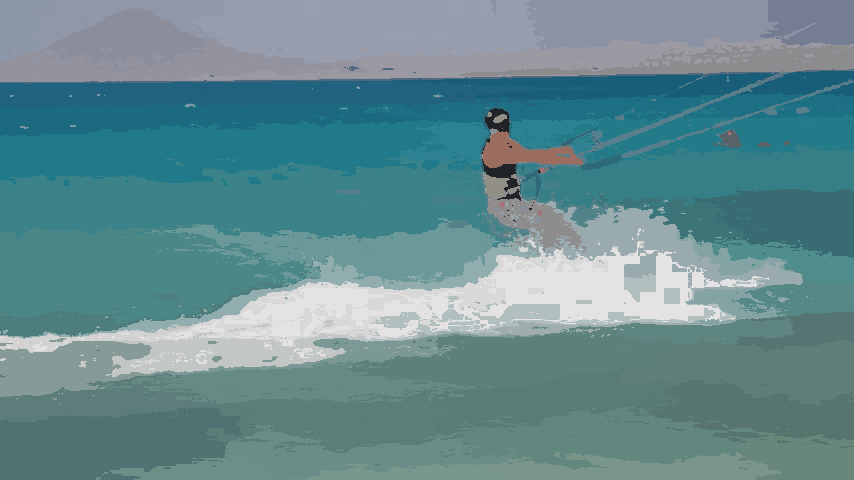}\end{subfigure}
    \begin{subfigure}{\MyWidth}\includegraphics[height=\MyHeight]{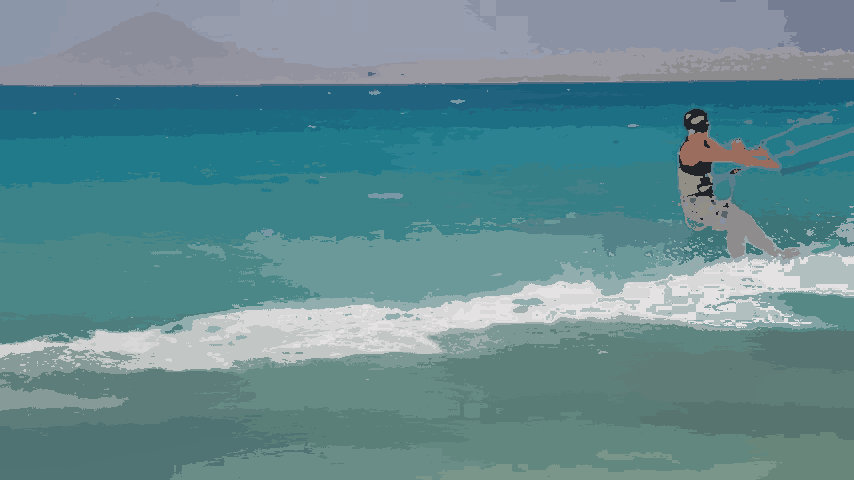}\end{subfigure}\\
    \captionsetup{justification=centering, singlelinecheck=false}
    \caption{Select frames from a video-segmentation task. Results shown for ScStream (which inferred 80 clusters).
    For comparison, see also additional results of a competing method (MiniBacth K-Means) in the \textbf{appendix}.}
     \label{Fig:video:OriginalAndScStream}
     \vspace*{-0.2cm}
\end{figure*}

\textbf{Evaluation.} 
In all the experiments we used a fixed batch size of $1000$ points. 
In order to evaluate the performance of each model we have used several popular metrics. 
Before processing each batch  we have used the current model to predict the labels of the batch,
and only then updated the model according to the new data. 
The only exception for this was the (unmodified) DPMM sampler, 
where we have clustered each batch separately and used the results as the labels.
The metrics we used on the predicted labels were ARI, Normalized Mutual Information (NMI), Purity and Pairwise F-Measure, where in all cases we report the mean result across all of the batches. In addition,
for the methods that support it, we have checked the consistency of the labels between batches,
comparing the predictions for the entire dataset and the true labels. For this we used Full NMI;
\ie, concatenating all the predictions (across the batches) and comparing the result, via NMI, to the true labels.  
\autoref{tab:results} summarizes the results. Our method almost uniformly outperforms the other streaming-data clustering methods. 
The exceptional metrics are Purity and Full NMI in the high-$K$ cases, as there the unfair advantage we gave the parametric methods is especially significant. 
See the \textbf{appendix} for the box plots of the different metrics,
revealing additional information (beyond the mean+std.~dev.). 
\autoref{tab:timing} shows that our running time  is on-par with most methods, 
except Mini-Batch K-means which is the fastest.  

\begin{table}[b]
    \centering
    \setlength\tabcolsep{3pt}
    \begin{tabular}{@{}lll@{}}
    \toprule
             &  Another sampling iter.   & Proposed subroutine  \\ \midrule
    ARI & $0.57\pm.18$ & $\mathbf{.63\pm.19}$          \\
    NMI & $0.65\pm.20$ & $\mathbf{.69\pm.15}$          \\
    Full NMI & $0.43\pm.02$ & $\mathbf{.48\pm.01}$          \\ \bottomrule
    \end{tabular}
    \caption{Using the deterministic subroutine
    as opposed to another sampling iteration. Data: ImageNet100 experiment.}
    \label{tab:pred_post}
\end{table}

\textbf{The deterministic subroutine.} 
We have repeated the ImageNet100 experiment, but this time without using that subroutine (and instead used an additional iteration of the unmodified restricted Gibbs sampler). 
The results, in~\autoref{tab:pred_post}, show clear benefits from using the proposed subroutine.  

\textbf{Video Temporal Segmentation.}
Image (non-semantic) segmentation is a popular computer-vision task, where an image is partitioned into 
several different segments. Here we consider a simple case where the segmentation is based on each
pixel's color and location. However, to demonstrate the utility of our streaming-data method,
we apply it to video segmentation (as opposed to a single-image segmentation). 
In this experiment, each batch is a video frame. The resolution of each RGB frame in the `kite-surf' video
(taken from the DAVIS dataset~\citep{Perazzi:CVPR2016:Davis})
was $480 \times 854$. This implies that each batch was of size $N=409920$ and $D=5$
(3 colors channels plus the 2D location of each pixel). The video contains $50$ frames, so in 
total we had $20.4M$ 5D samples. Each frame took ScStream 0.6 [sec] to process (including I/O). 
In this quantitative experiment, the pros of using our method stand out. 
For example. unlike many other methods (such as DBSTREAM, Birch, \etc), 
we do not have label switching, thus it is possible to have  temporally-consistent clusters.
In contrast, methods that suffer from label switching cannot guarantee inter-frame consistency. 
\autoref{Fig:video:OriginalAndScStream} shows example results of our method. Results of 
another method that does not suffer from label switching, 
  MiniBatch K-means~\citep{Sculley:ICWWW:2010:batch-kmeans}, appear in the 
  \textbf{appendix}. However, even if we ignore that fact
  that MiniBatch K-means is parametric (so it must be given the value of $K$), 
  it cannot handle concept drifts; 
 \eg, as the figure in the appendix shows, each cluster remains in a similar spatial location between frames, 
 and the surfer is barely distinguishable from the background.
  However, ScStream solves these problems as is evident
  by the fact that the surfer is clearly distinguishable from the background,
  and his label is consistent across the frames, despite
  the fact that his associated  clusters change their statistics over time.

 \section{CONCLUSION}
We have proposed a new BNP method, called ScStream, for clustering streaming data. ScStream, a streaming-data extension of a recent fast implementation of a (non-streaming) DPMM sampler, is fast and achieves SOTA results. While ScStream supports any exponential family for the component type, 
it cannot handle arbitrary-shaped clusters; this is arguably its key limitation. For simplicity, our presentation here assumed that the batches arrive at times $\set{1,2,3,\ldots}$; however, our method is general enough to support arrivals at any monotonically-increasing time sequence (\ie,  $t_1<t_2<\ldots$)
and our code already supports this generality. 
The only difference in such a case is that $\Kcal(t_B,t_b)$ is used instead of $\Kcal(B,b)$.  
Our publicly-available code is easy to use and offers the user an interface in either Julia or Python.

\clearpage
\small
\bibliographystyle{abbrvnat}
\bibliography{./refs}

\clearpage
\appendix
\onecolumn \makesupplementtitle
\thispagestyle{empty}
\begin{abstract}
  This documents contains the following:
\begin{enumerate}
\item an additional figure for the video segmentation experiment; 
\item additional Boxplots for our main experiment, omitted from the paper due to space limits.
\item an empirical verification that our runtime grows linearly with $T$ (\ie, the number of iterations
 of the restricted Gibbs sampler) and that a \emph{very low} $T$ suffices for good results;
 \item the details of a single iteration in the original restricted Gibbs sampler (Chang and Fisher III, 2013); 
\item the expressions for the marginal likelihood for Gaussian and multinomials components;
\item the posterior calculations in the multinomial case(the Gaussian case was already included in our paper);
\item the predictive posterior in the multinomal case  (the Gaussian case was already included in our paper);
\item the full details of our experiments hyper-params and machine specification.
 \end{enumerate}

\end{abstract}
\clearpage
\section{Additional Video Segmentation Results}
\begin{figure*}[h]
  \centering
  \newcommand{\MySpace}{\vspace{-.15cm}}
  \newcommand{\MyHeight}{1.5cm}
  \newcommand{\MyWidth}{0.15\linewidth}
  \newcommand{\MySmallWidth}{0.008\linewidth}
  \vspace{-.2cm}
  \begin{subfigure}{\MySmallWidth}\rotatebox{90}{\scriptsize Original}\end{subfigure}
  \begin{subfigure}{\MyWidth}\includegraphics[height=\MyHeight]{./figures/video_seg/original/00001.png}\end{subfigure}
  \begin{subfigure}{\MyWidth}\includegraphics[height=\MyHeight]{./figures/video_seg/original/00013.png}\end{subfigure}
  \begin{subfigure}{\MyWidth}\includegraphics[height=\MyHeight]{./figures/video_seg/original/00026.png}\end{subfigure}
  \begin{subfigure}{\MyWidth}\includegraphics[height=\MyHeight]{./figures/video_seg/original/00032.png}\end{subfigure}
  \begin{subfigure}{\MyWidth}\includegraphics[height=\MyHeight]{./figures/video_seg/original/00039.png}\end{subfigure}
  \begin{subfigure}{\MyWidth}\includegraphics[height=\MyHeight]{./figures/video_seg/original/00049.png}\end{subfigure}\\
  \begin{subfigure}{\MySmallWidth}\rotatebox{90}{\scriptsize MBK(K=30)}\end{subfigure}
  \begin{subfigure}{\MyWidth}\includegraphics[height=\MyHeight]{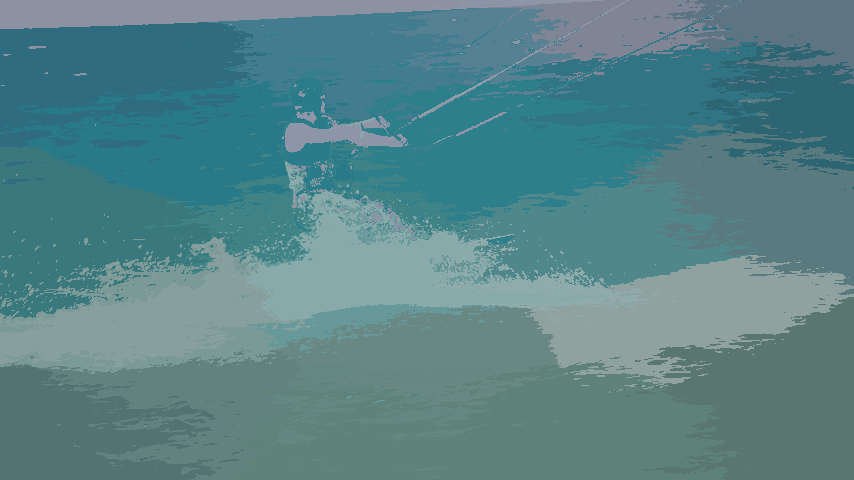}\end{subfigure}
  \begin{subfigure}{\MyWidth}\includegraphics[height=\MyHeight]{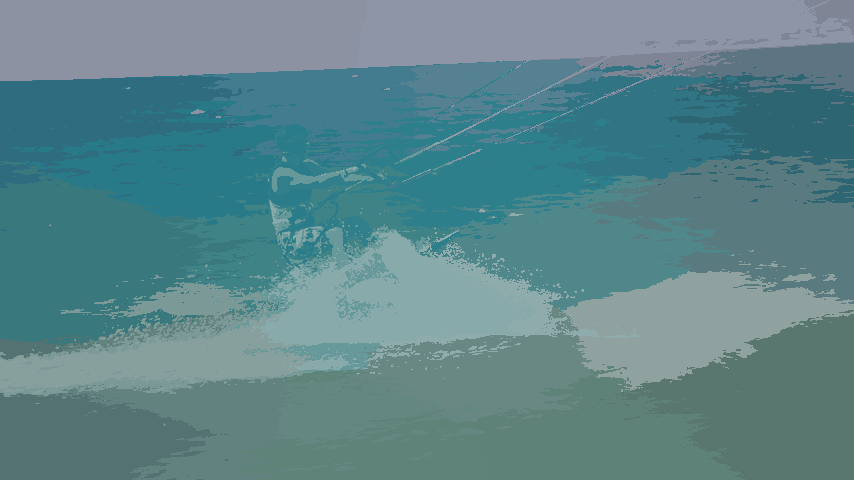}\end{subfigure}
  \begin{subfigure}{\MyWidth}\includegraphics[height=\MyHeight]{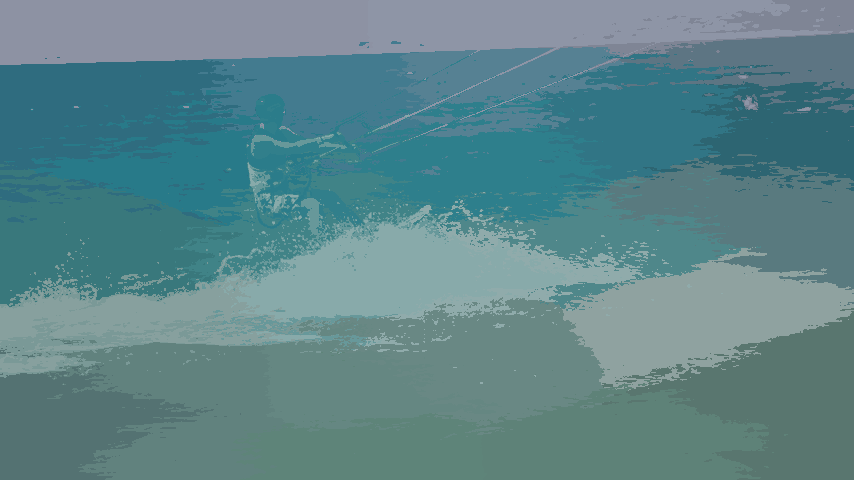}\end{subfigure}
  \begin{subfigure}{\MyWidth}\includegraphics[height=\MyHeight]{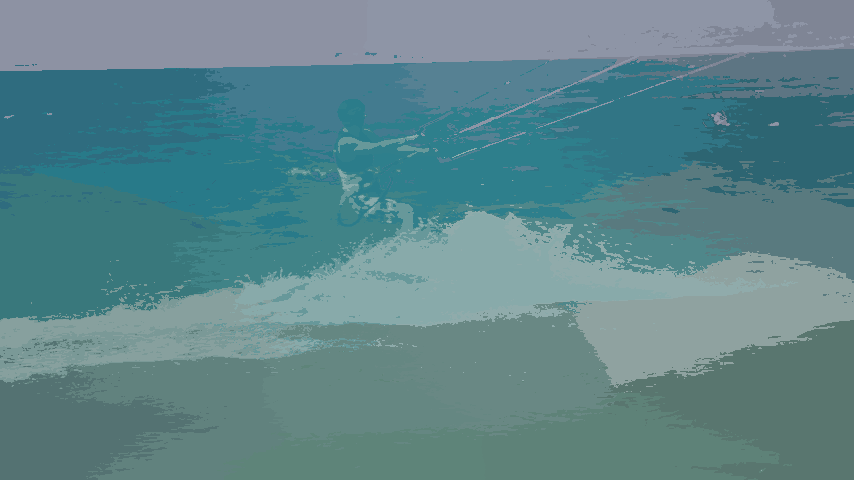}\end{subfigure}
  \begin{subfigure}{\MyWidth}\includegraphics[height=\MyHeight]{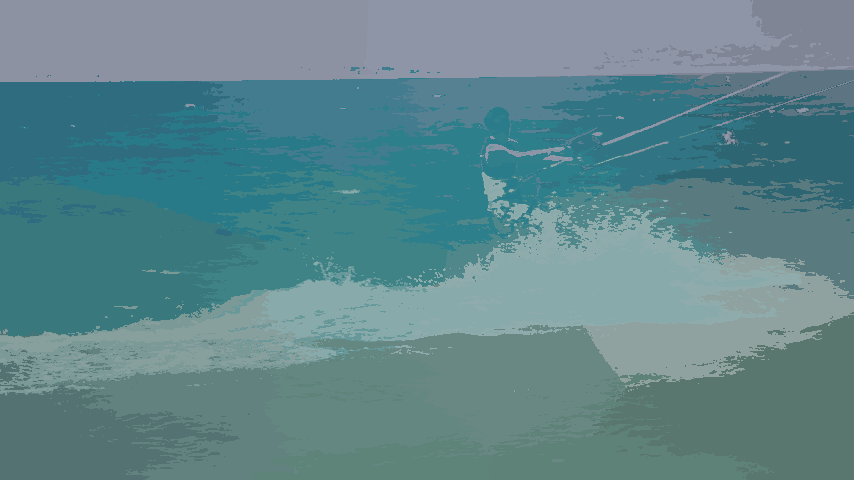}\end{subfigure}
  \begin{subfigure}{\MyWidth}\includegraphics[height=\MyHeight]{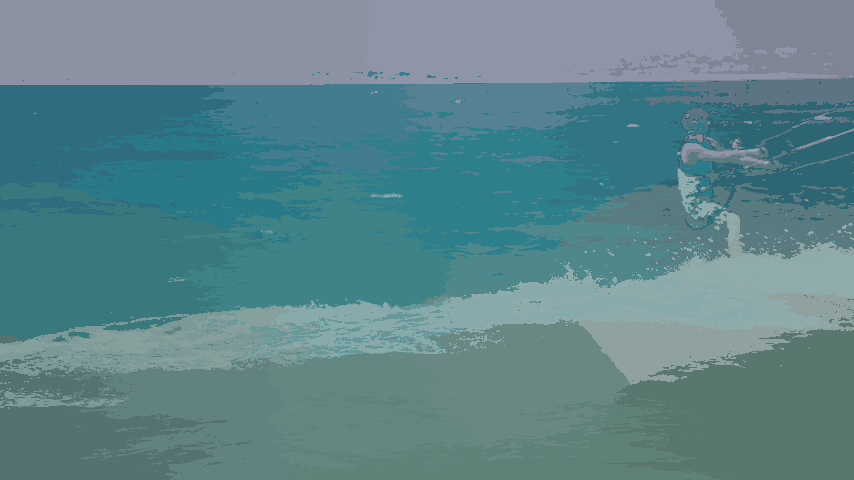}\end{subfigure}\\
  \begin{subfigure}{\MySmallWidth}\rotatebox{90}{\scriptsize MBK(K=80)}\end{subfigure}
  \begin{subfigure}{\MyWidth}\includegraphics[height=\MyHeight]{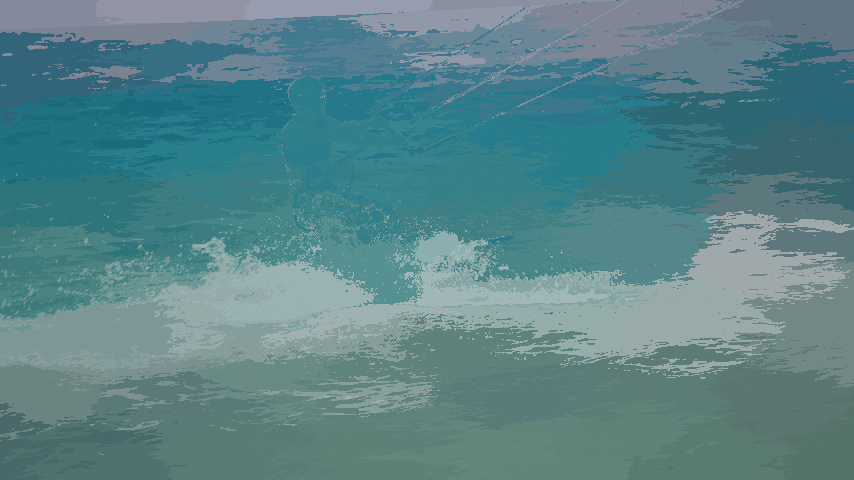}\end{subfigure}
  \begin{subfigure}{\MyWidth}\includegraphics[height=\MyHeight]{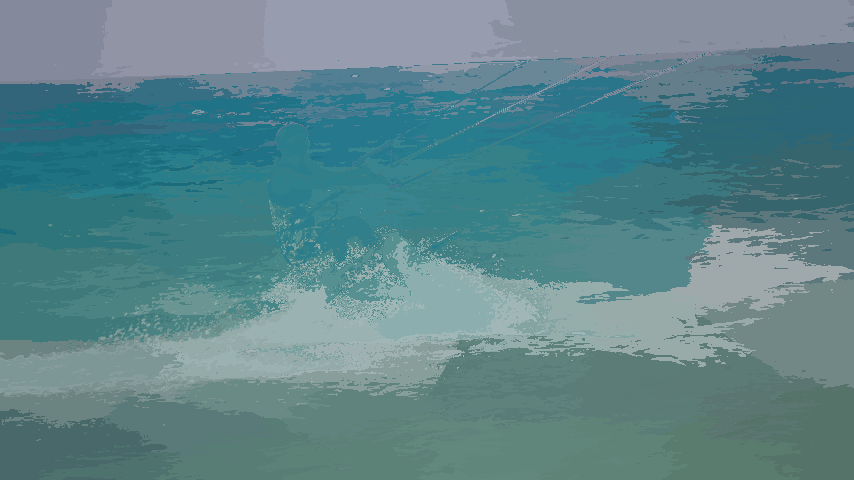}\end{subfigure}
  \begin{subfigure}{\MyWidth}\includegraphics[height=\MyHeight]{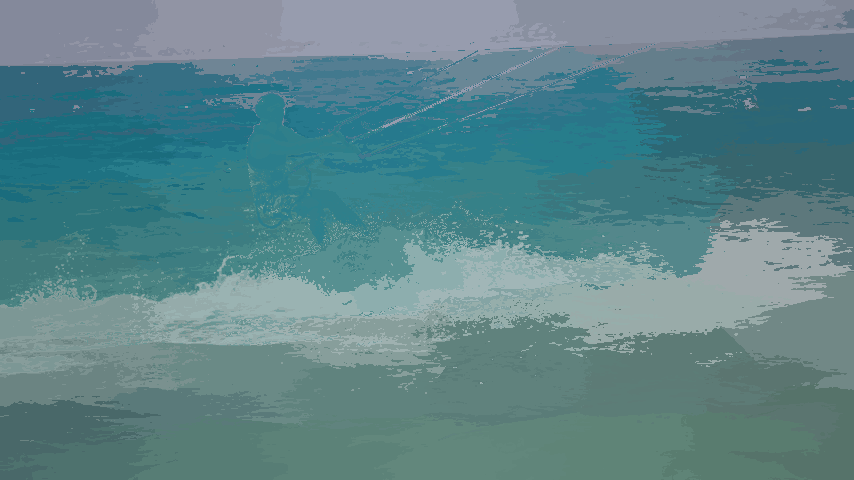}\end{subfigure}
  \begin{subfigure}{\MyWidth}\includegraphics[height=\MyHeight]{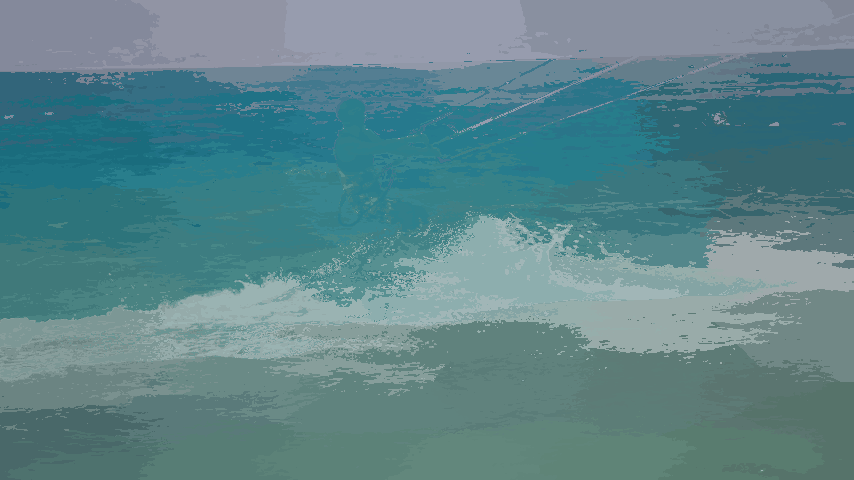}\end{subfigure}
  \begin{subfigure}{\MyWidth}\includegraphics[height=\MyHeight]{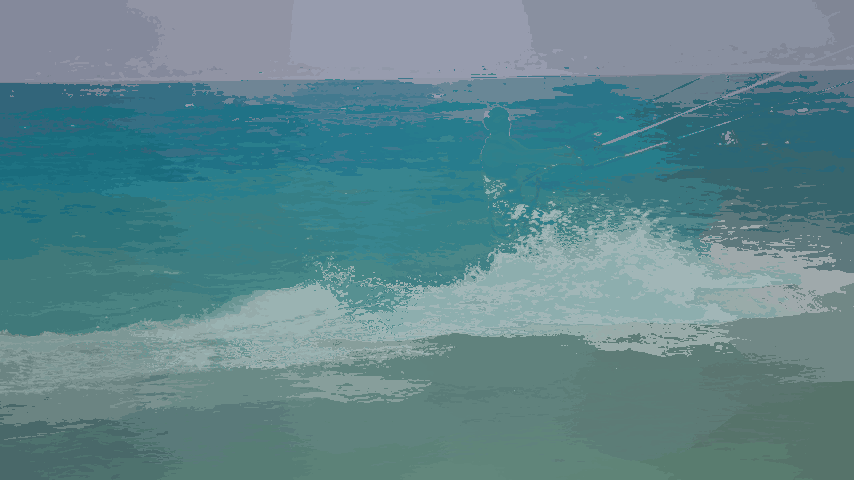}\end{subfigure}
  \begin{subfigure}{\MyWidth}\includegraphics[height=\MyHeight]{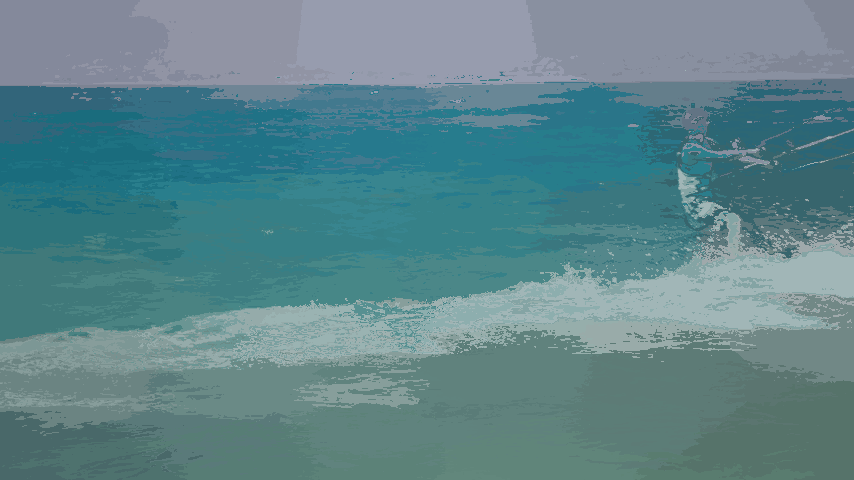}\end{subfigure}\\
  \begin{subfigure}{\MySmallWidth}\rotatebox{90}{\scriptsize MBK(K=500)}\end{subfigure}
  \begin{subfigure}{\MyWidth}\includegraphics[height=\MyHeight]{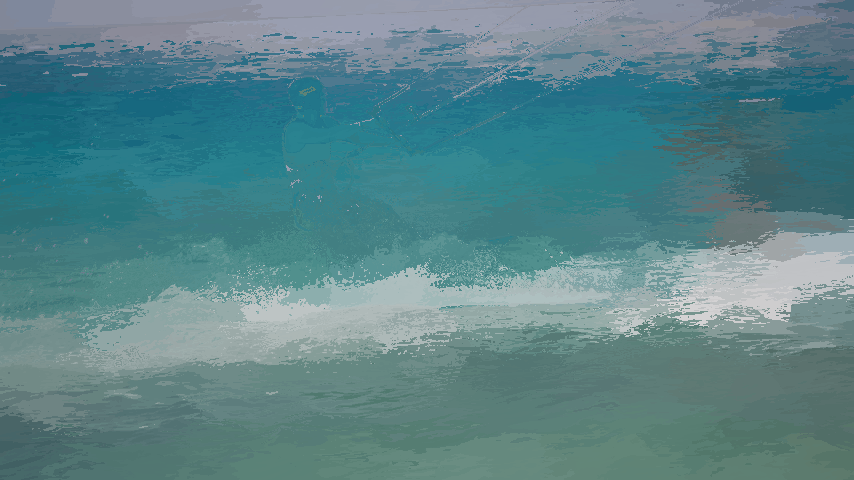}\end{subfigure}
  \begin{subfigure}{\MyWidth}\includegraphics[height=\MyHeight]{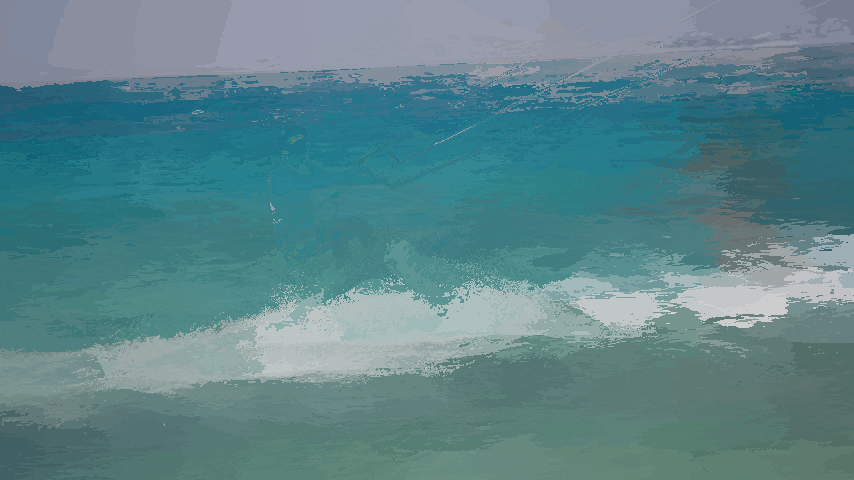}\end{subfigure}
  \begin{subfigure}{\MyWidth}\includegraphics[height=\MyHeight]{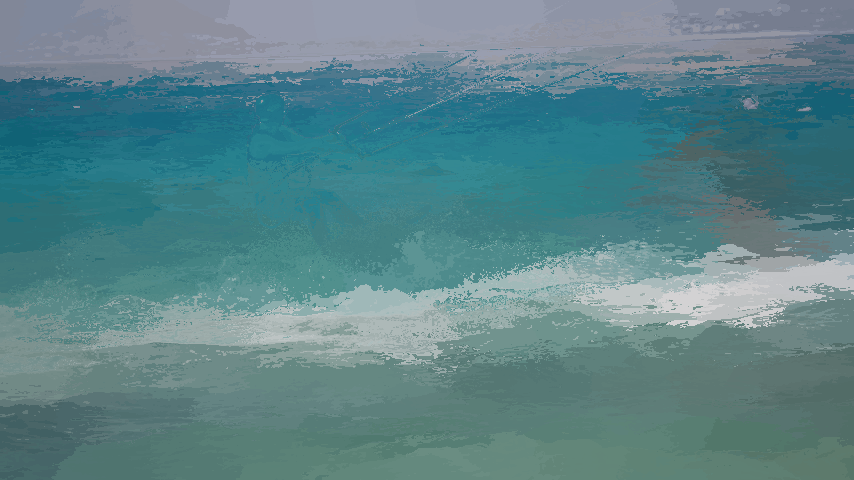}\end{subfigure}
  \begin{subfigure}{\MyWidth}\includegraphics[height=\MyHeight]{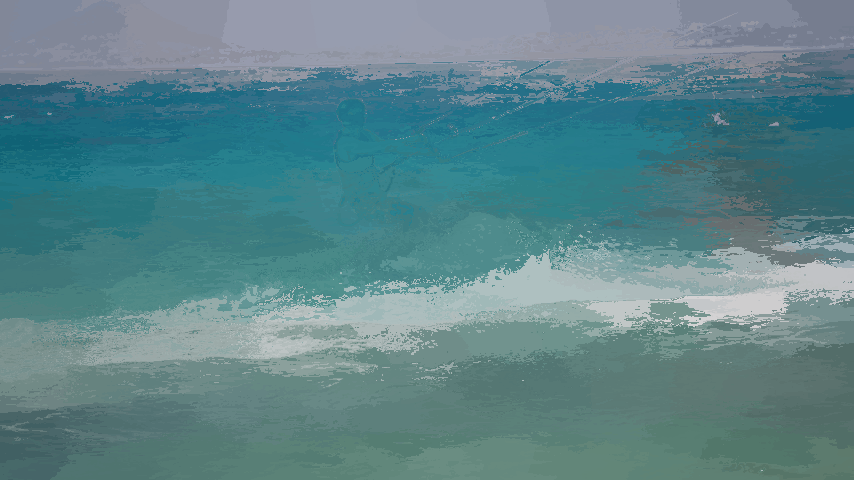}\end{subfigure}
  \begin{subfigure}{\MyWidth}\includegraphics[height=\MyHeight]{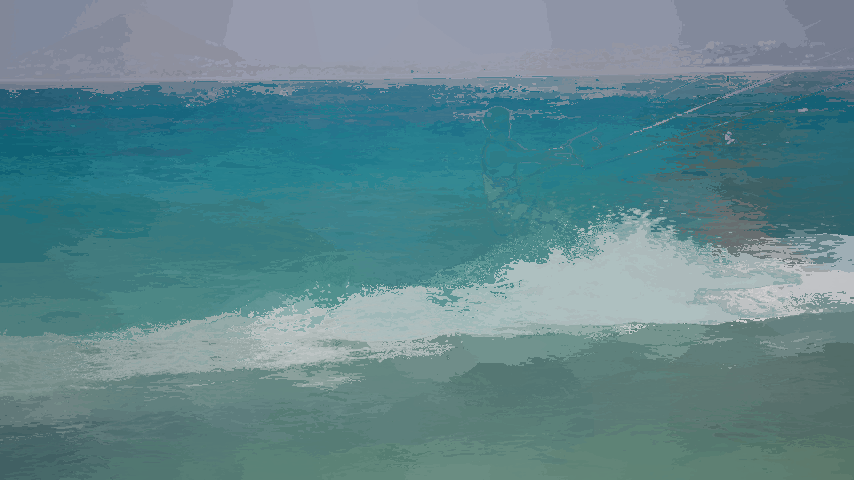}\end{subfigure}
  \begin{subfigure}{\MyWidth}\includegraphics[height=\MyHeight]{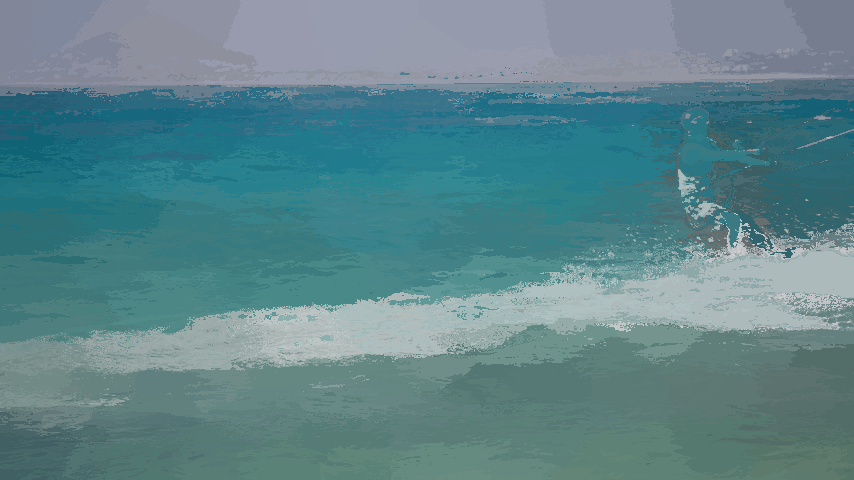}\end{subfigure}\\
  \begin{subfigure}{\MySmallWidth}\rotatebox{90}{\scriptsize ScStream}\end{subfigure}
  \begin{subfigure}{\MyWidth}\includegraphics[height=\MyHeight]{./figures/video_seg/scstream/00001.png}\end{subfigure}
  \begin{subfigure}{\MyWidth}\includegraphics[height=\MyHeight]{./figures/video_seg/scstream/00013.png}\end{subfigure}
  \begin{subfigure}{\MyWidth}\includegraphics[height=\MyHeight]{./figures/video_seg/scstream/00026.png}\end{subfigure}
  \begin{subfigure}{\MyWidth}\includegraphics[height=\MyHeight]{./figures/video_seg/scstream/00032.png}\end{subfigure}
  \begin{subfigure}{\MyWidth}\includegraphics[height=\MyHeight]{./figures/video_seg/scstream/00039.png}\end{subfigure}
  \begin{subfigure}{\MyWidth}\includegraphics[height=\MyHeight]{./figures/video_seg/scstream/00049.png}\end{subfigure}\\

  \captionsetup{justification=centering, singlelinecheck=false}
  \caption{Video segmentation (example frames). Results shown for MiniBatch-Kmeans (denoted as MBK)
  with several different $K$ values, as well as for ScStream (which inferred 80 clusters).
  }
   \label{Fig:video}
   \vspace*{-0.2cm}
\end{figure*}
\section{Boxplots}
Here we provide boxplots for the ARI, NMI, Purity and F-measure metrics (for the same experiments as in the paper). Recall that for some of these metrics, the parametric methods (namely, methods that need to know $K$) enjoy an unfair advantage as they were given the true value of $K$. The parametric methods are: 
BIRCH; CluStream; StreamKM++; MB-Kmeans.

\begin{figure}[h]
    \centering
    \newcommand{\MySpace}{\vspace{-.15cm}}
    \newcommand{\MyHeight}{3.0cm}
    \vspace{-.2cm}
    \subcaptionbox{Gaussian 2D\label{Fig:ARI:Gaussian2d}}[0.32\linewidth]{\includegraphics[height=\MyHeight,trim=0.0cm 0.4cm 0.0cm 0.2cm]{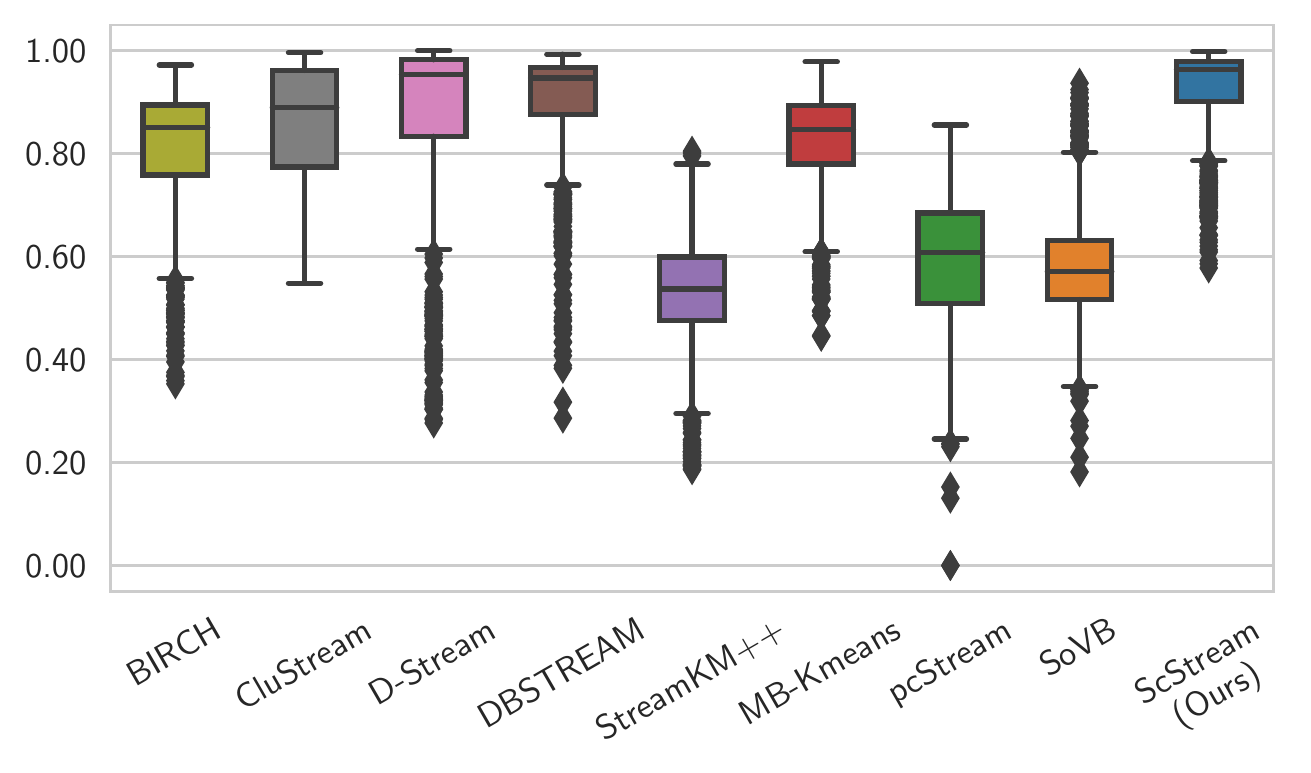}\MySpace}
    \subcaptionbox{CoverType\label{Fig:ARI:CoverType}}[0.32\linewidth]{\includegraphics[height=\MyHeight,trim=0.0cm 0.4cm 0.0cm 0.2cm]{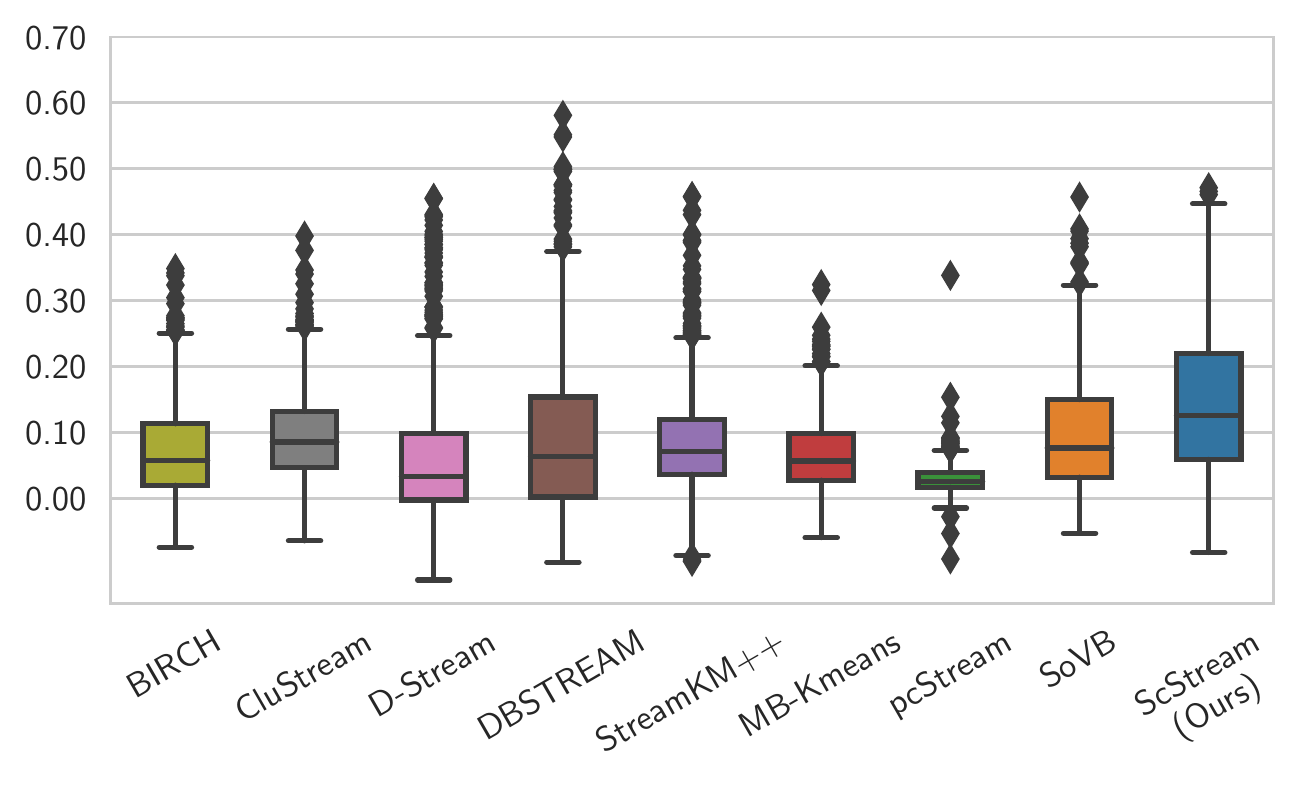}\MySpace}
    \subcaptionbox{ImageNet100\label{Fig:ARI:ImageNet100}}[0.32\linewidth]{\includegraphics[height=\MyHeight,trim=0.0cm 0.4cm 0.0cm 0.2cm]{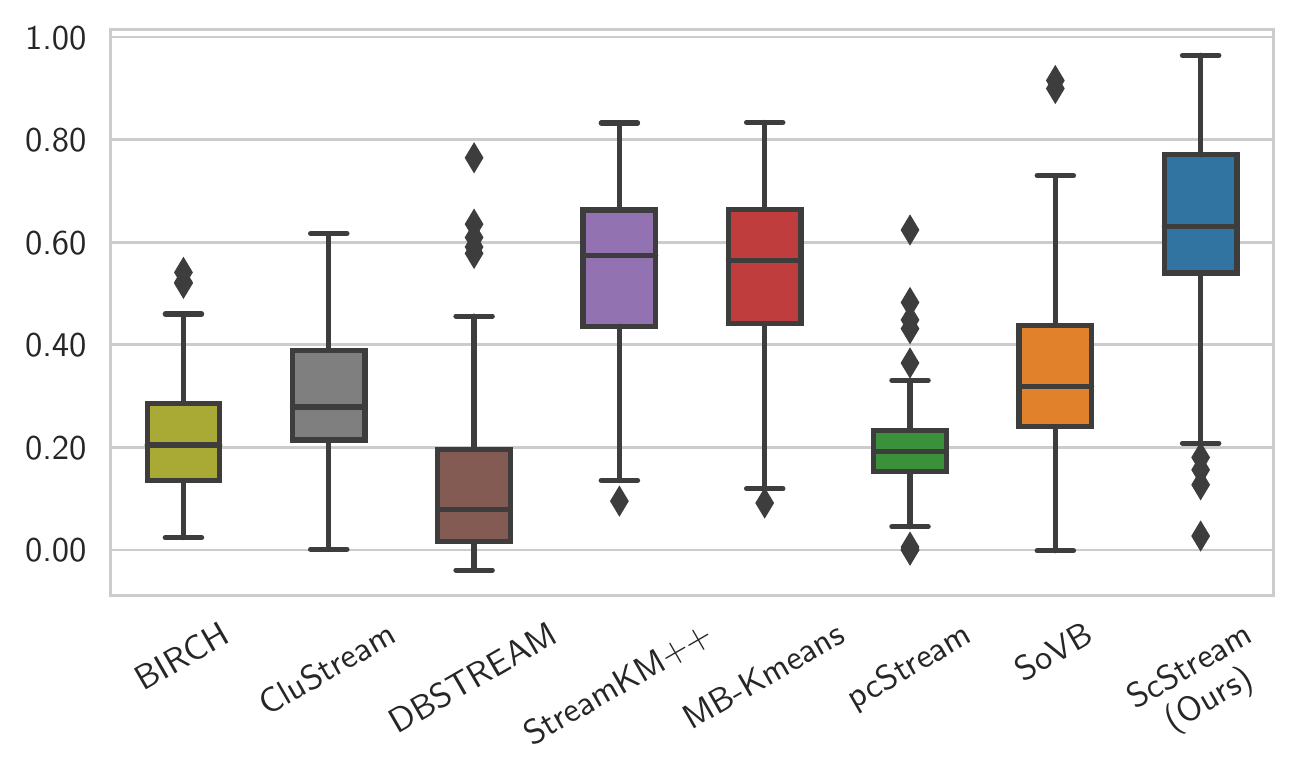}\MySpace}
    \subcaptionbox{ImageNet1K\label{Fig:ARI:ImageNet1K}}[0.32\linewidth]{\includegraphics[height=\MyHeight,trim=0.0cm 0.4cm 0.0cm 0.2cm]{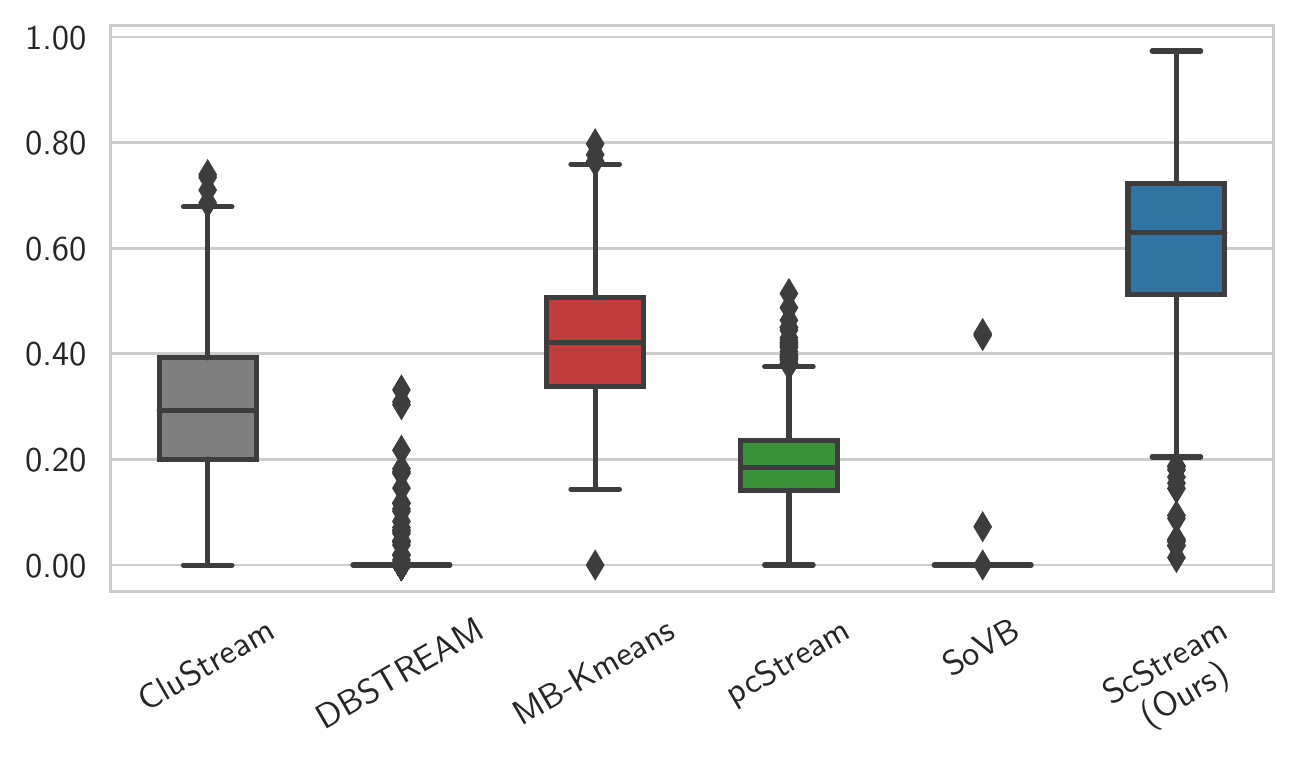}\MySpace}
    \subcaptionbox{Multinomial\label{Fig:ARI:Multinomial}}[0.32\linewidth]{\includegraphics[height=\MyHeight,trim=0.0cm 0.4cm 0.0cm 0.2cm]{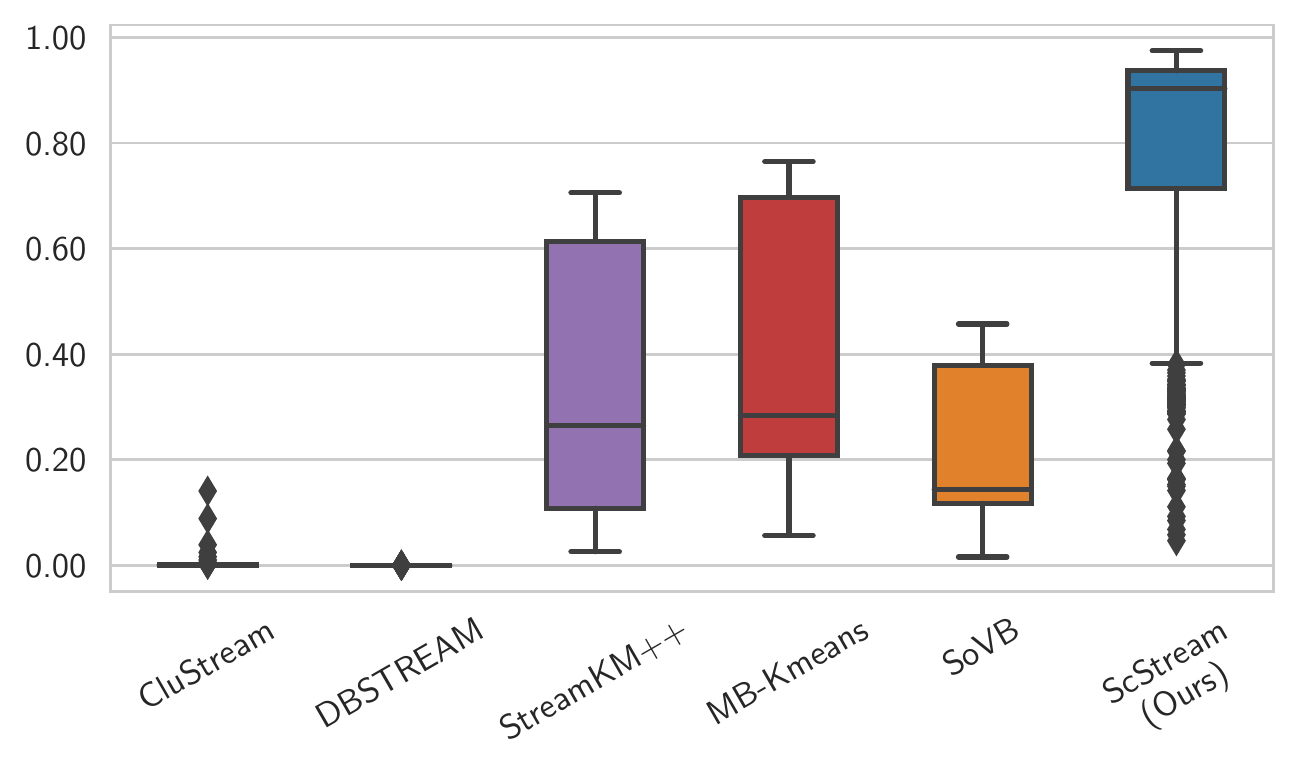}\MySpace}
    \subcaptionbox{20NewsGroups\label{Fig:ARI:20NewsGroups}}[0.32\linewidth]{\includegraphics[height=\MyHeight,trim=0.0cm 0.4cm 0.0cm 0.2cm]{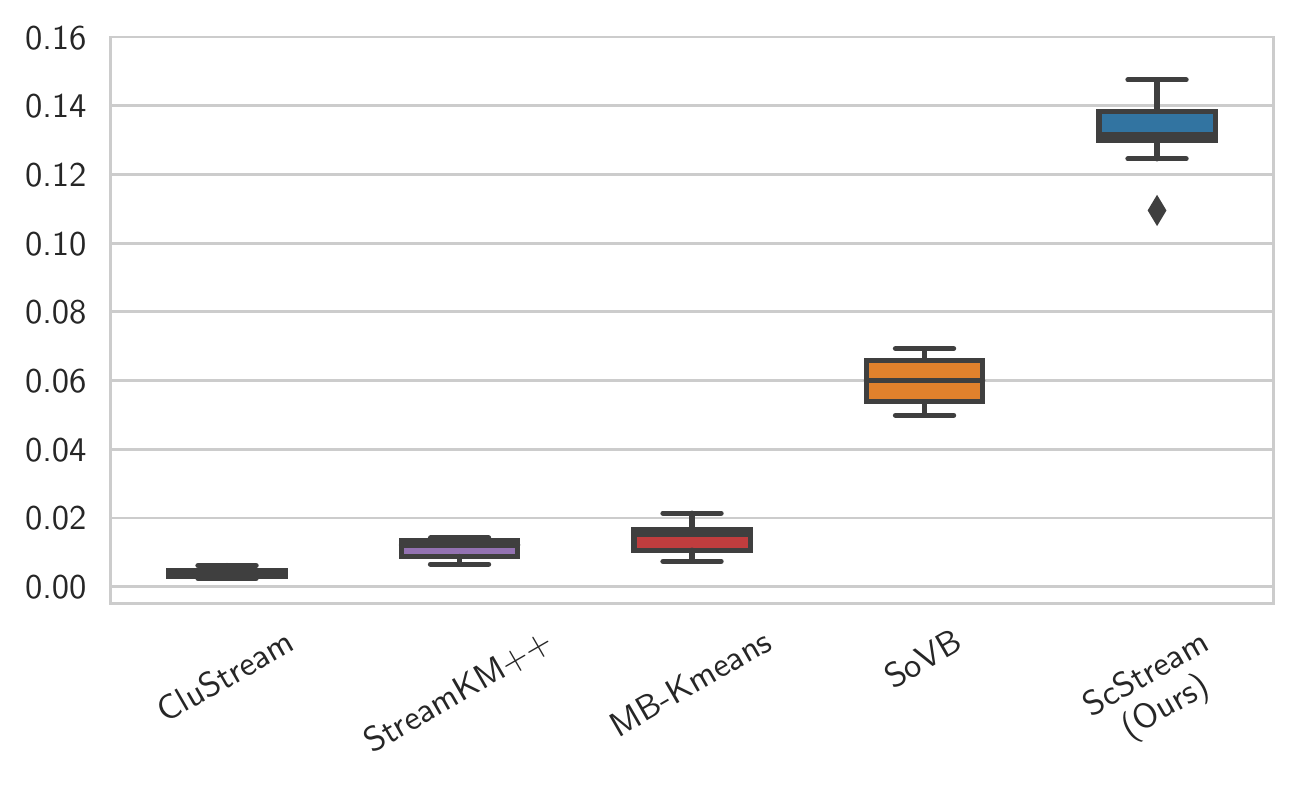}\MySpace}
    \captionsetup{justification=centering, singlelinecheck=false}
    \caption{Box plots of the ARI metric for each of the experiments.
    }
     \label{Fig:aris}
     \vspace*{-0.2cm}
\end{figure}

\begin{figure}[h]
    \centering
    \newcommand{\MySpace}{\vspace{-.15cm}}
    \newcommand{\MyHeight}{3.0cm}
    \subcaptionbox{Gaussian 2D\label{Fig:NMI:Gaussian2d}}[0.32\linewidth]{\includegraphics[height=\MyHeight,trim=0.0cm 0.0cm 0.0cm 0.0cm]{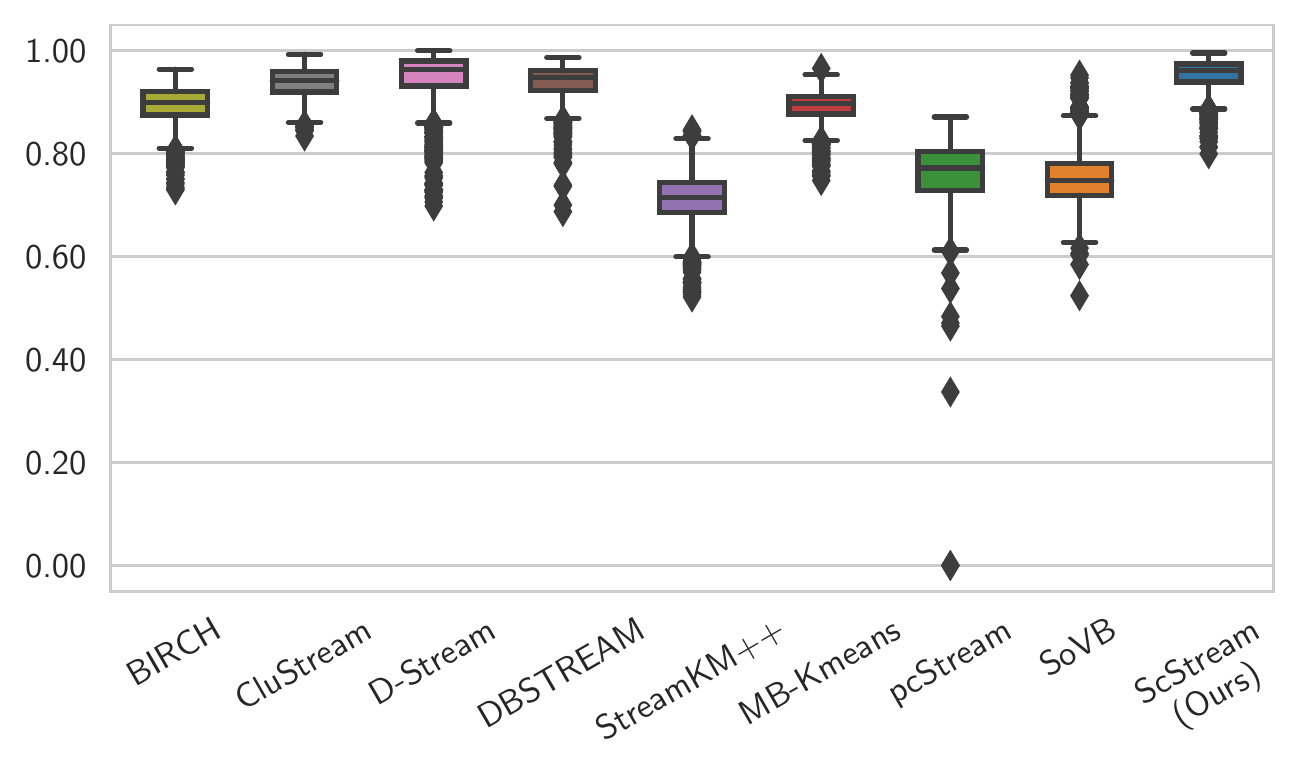}\MySpace}
    \subcaptionbox{CoverType\label{Fig:NMI:CoverType}}[0.32\linewidth]{\includegraphics[height=\MyHeight,trim=0.0cm 0.0cm 0.0cm 0.0cm]{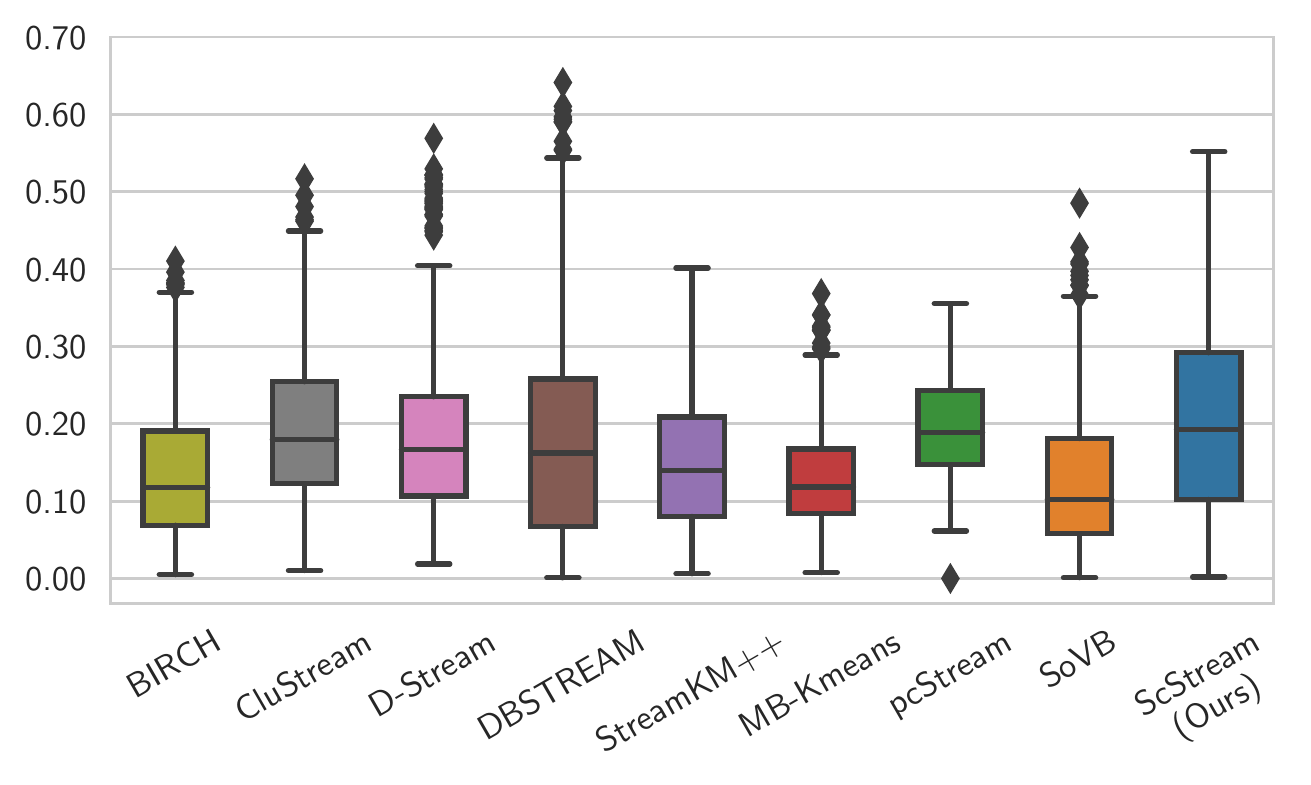}\MySpace}
    \subcaptionbox{ImageNet100\label{Fig:NMI:ImageNet100}}[0.32\linewidth]{\includegraphics[height=\MyHeight,trim=0.0cm 0.0cm 0.0cm 0.0cm]{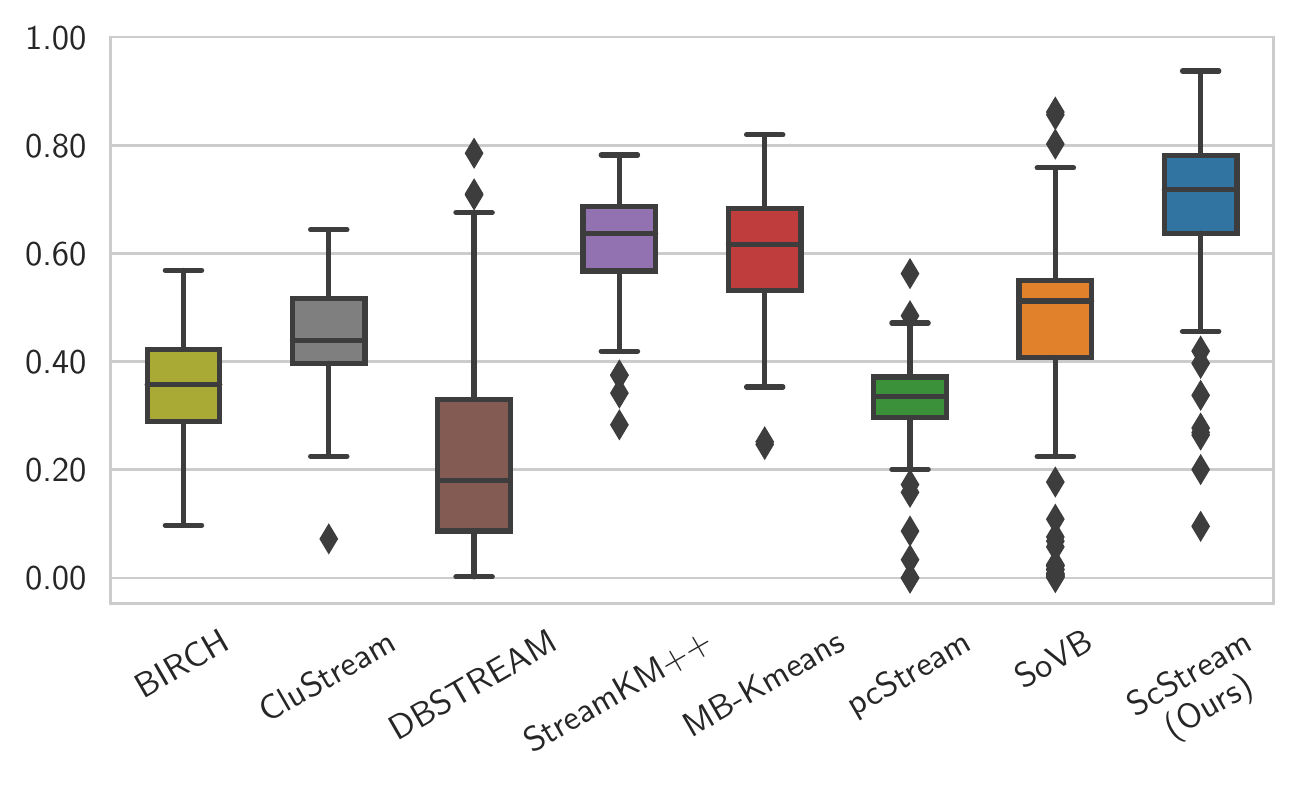}\MySpace}
    \subcaptionbox{ImageNet1K\label{Fig:NMI:ImageNet1K}}[0.32\linewidth]{\includegraphics[height=\MyHeight,trim=0.0cm 0.0cm 0.0cm 0.0cm]{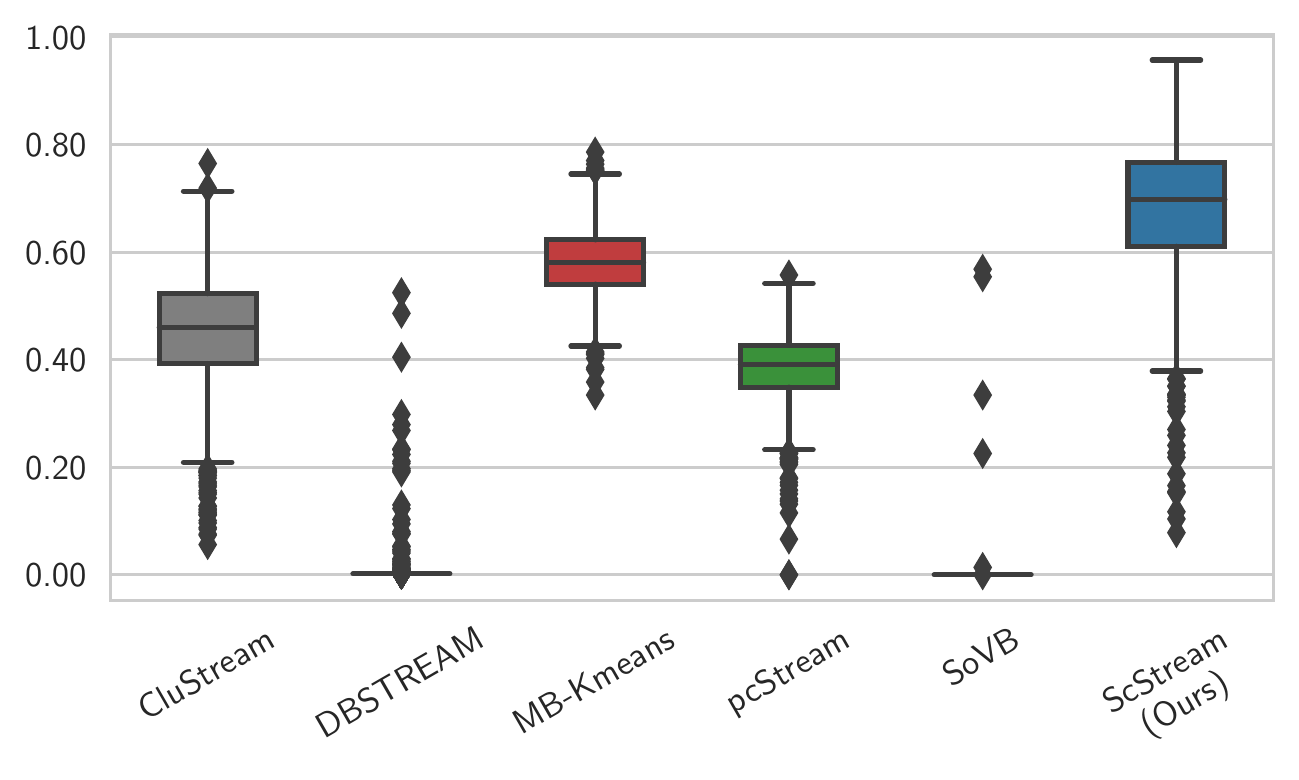}\MySpace}
    \subcaptionbox{Multinomial\label{Fig:NMI:Multinomial}}[0.32\linewidth]{\includegraphics[height=\MyHeight,trim=0.0cm 0.0cm 0.0cm 0.0cm]{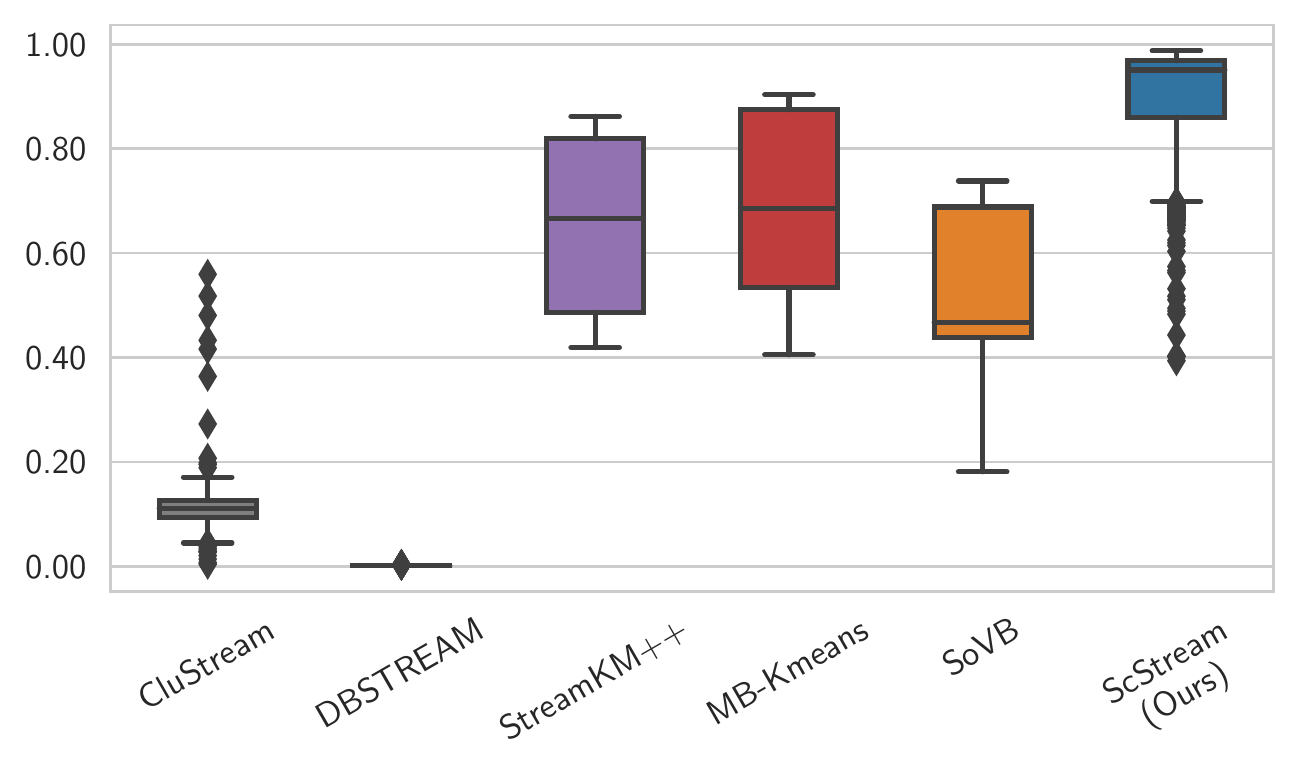}\MySpace}
    \subcaptionbox{20NewsGroups\label{Fig:NMI:20NewsGroups}}[0.32\linewidth]{\includegraphics[height=\MyHeight,trim=0.0cm 0.0cm 0.0cm 0.0cm]{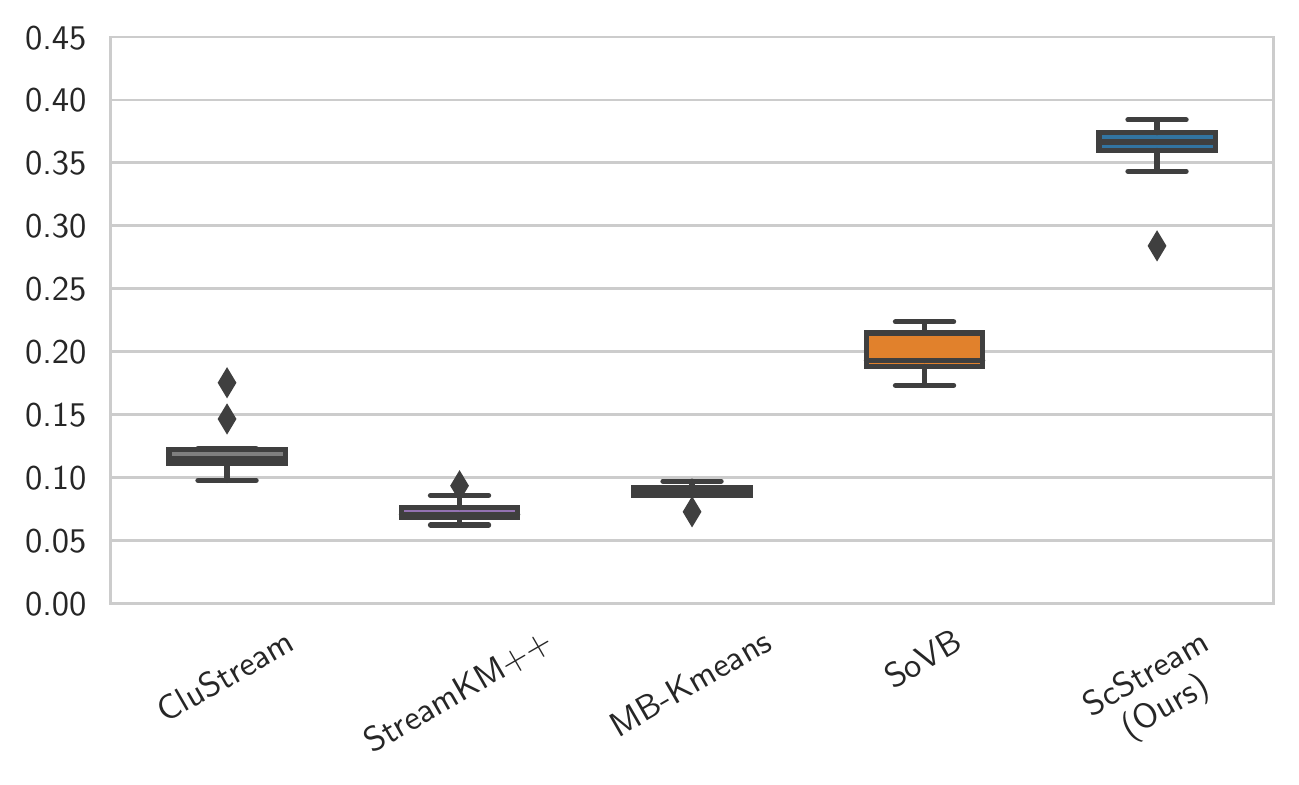}\MySpace}
\captionsetup{justification=centering, singlelinecheck=false}    
\caption{Box plots of the NMI metric for each of the experiments.
    }
     \label{Fig:nmis}
\end{figure}

\begin{figure*}[h]
    \centering
    \newcommand{\MySpace}{\vspace{-.15cm}}
    \newcommand{\MyHeight}{3.0cm}
    \subcaptionbox{Gaussian 2D\label{Fig:Purity:Gaussian2d}}[0.32\linewidth]{\includegraphics[height=\MyHeight,trim=0.0cm 0.0cm 0.0cm 0.0cm]{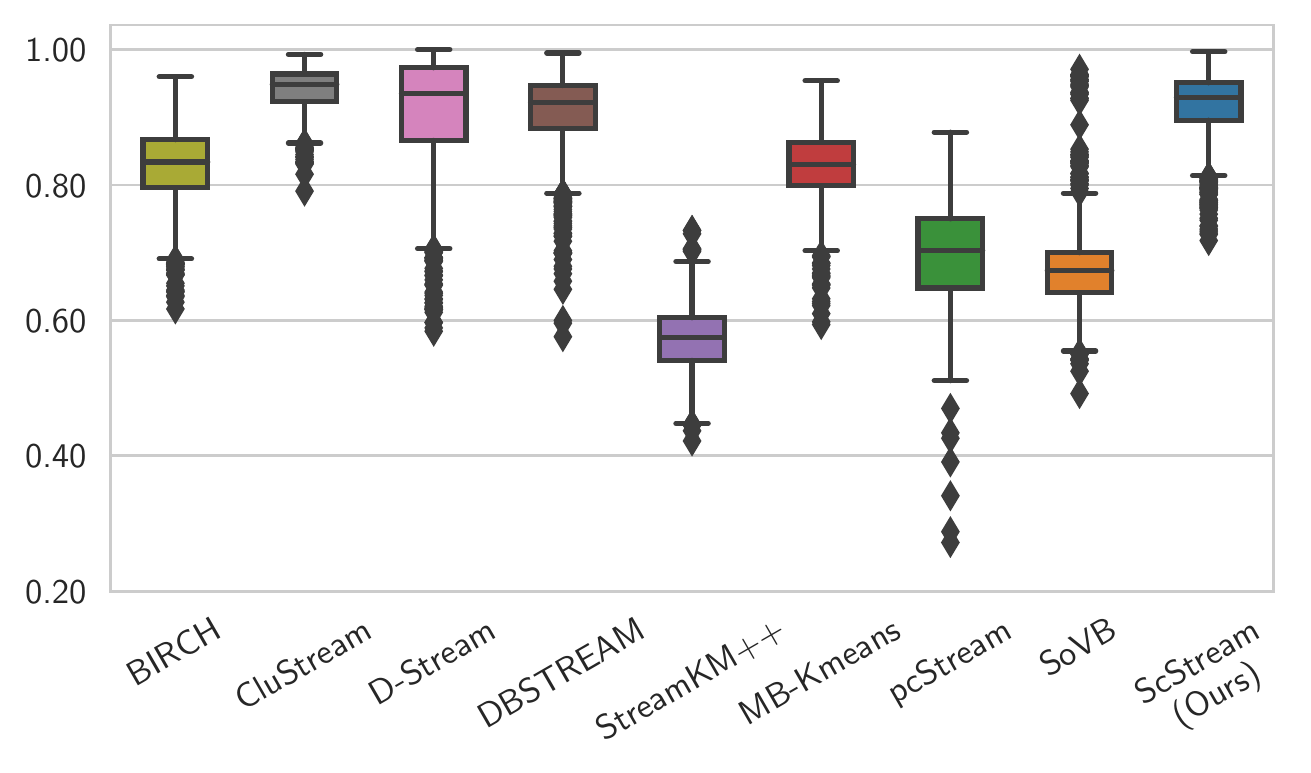}\MySpace}
    \subcaptionbox{CoverType\label{Fig:Purity:CoverType}}[0.32\linewidth]{\includegraphics[height=\MyHeight,trim=0.0cm 0.0cm 0.0cm 0.0cm]{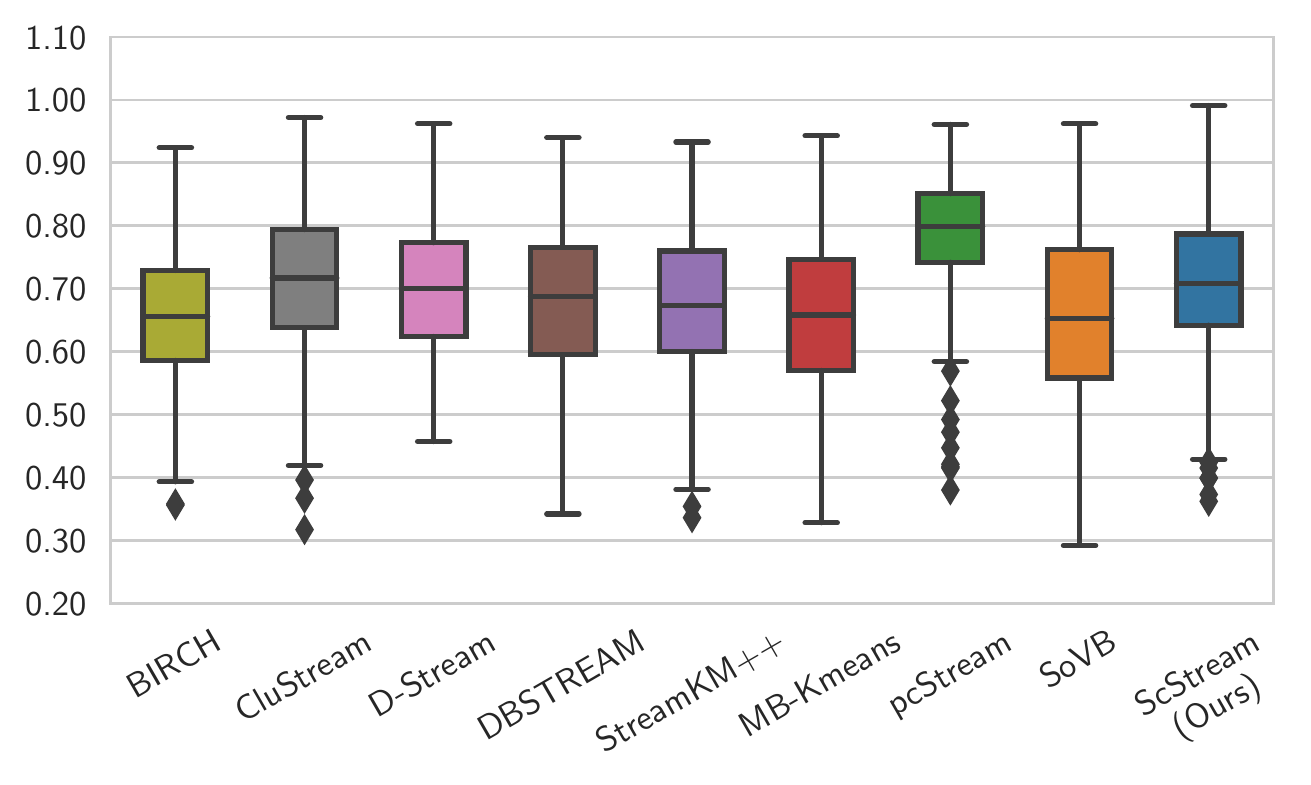}\MySpace}
    \subcaptionbox{ImageNet100\label{Fig:Purity:ImageNet100}}[0.32\linewidth]{\includegraphics[height=\MyHeight,trim=0.0cm 0.0cm 0.0cm 0.0cm]{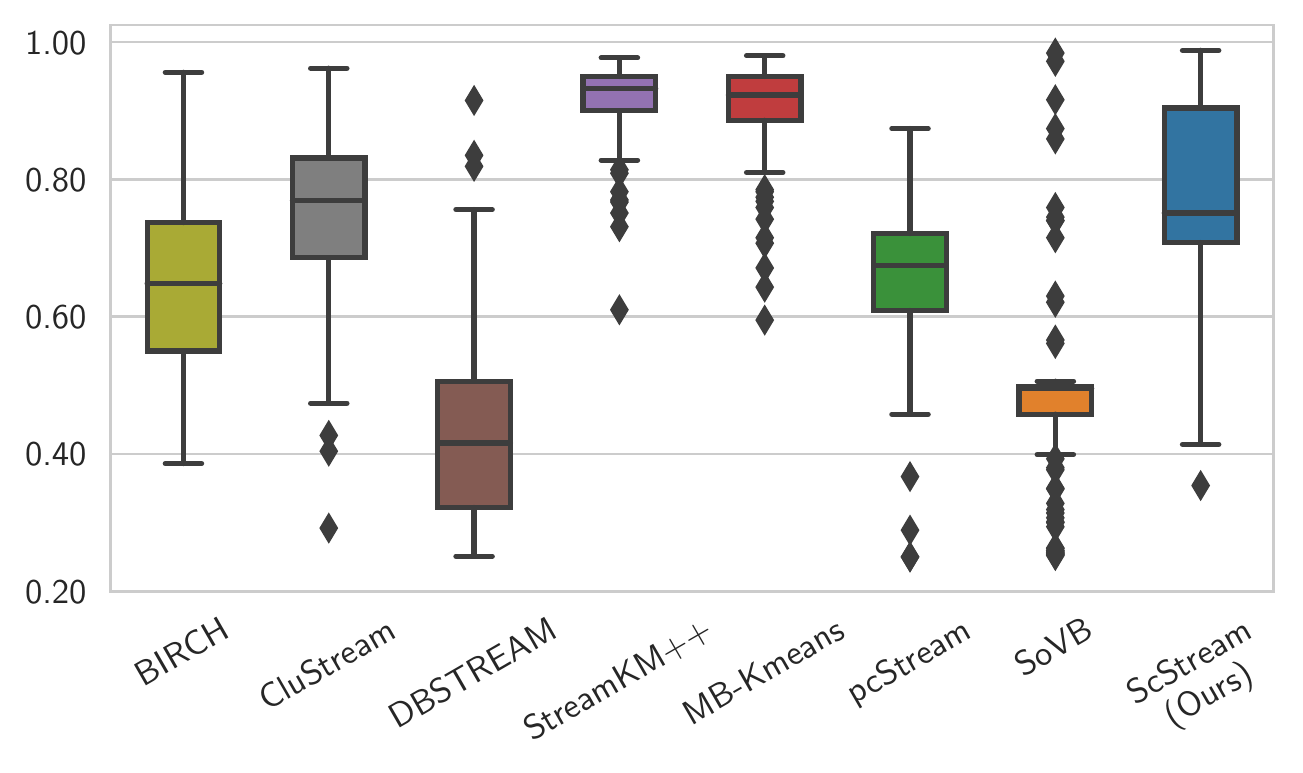}\MySpace}
    \subcaptionbox{ImageNet1K\label{Fig:Purity:ImageNet1K}}[0.32\linewidth]{\includegraphics[height=\MyHeight,trim=0.0cm 0.0cm 0.0cm 0.0cm]{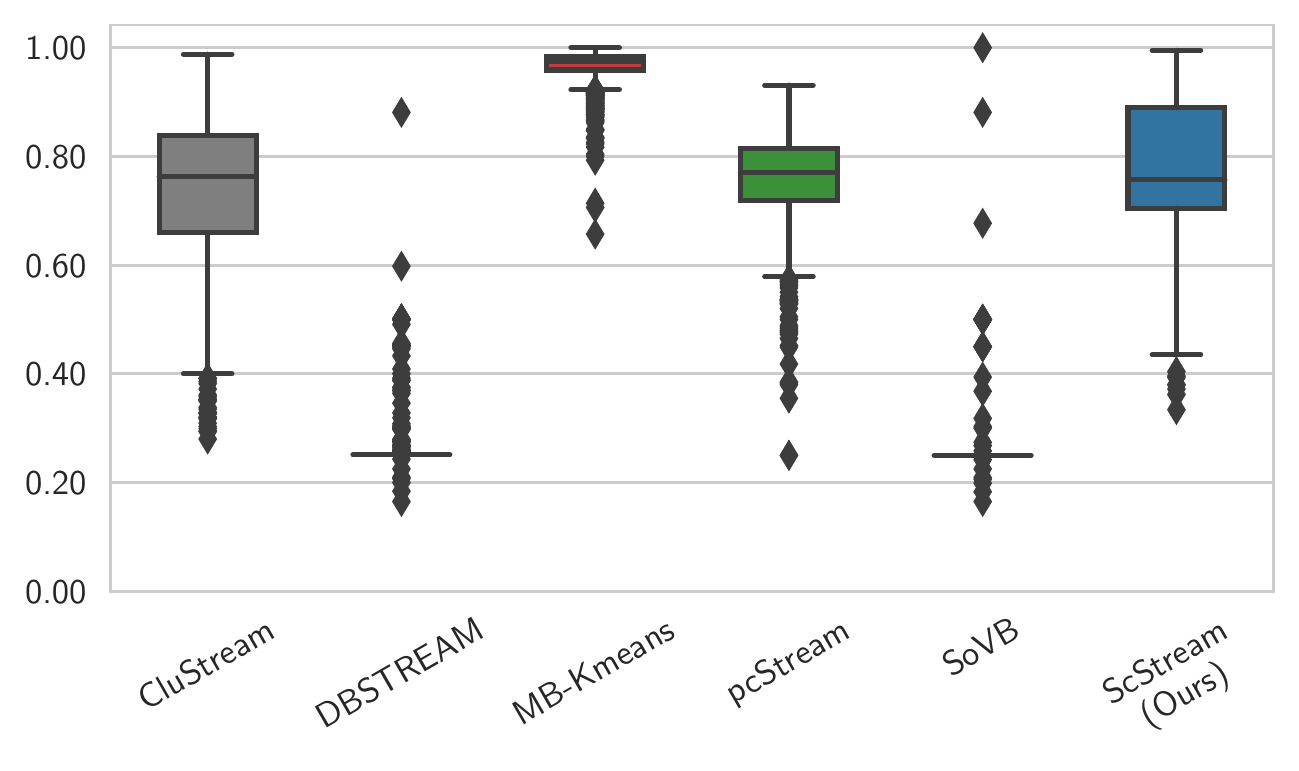}\MySpace}
    \subcaptionbox{Multinomial\label{Fig:Purity:Multinomial}}[0.32\linewidth]{\includegraphics[height=\MyHeight,trim=0.0cm 0.0cm 0.0cm 0.0cm]{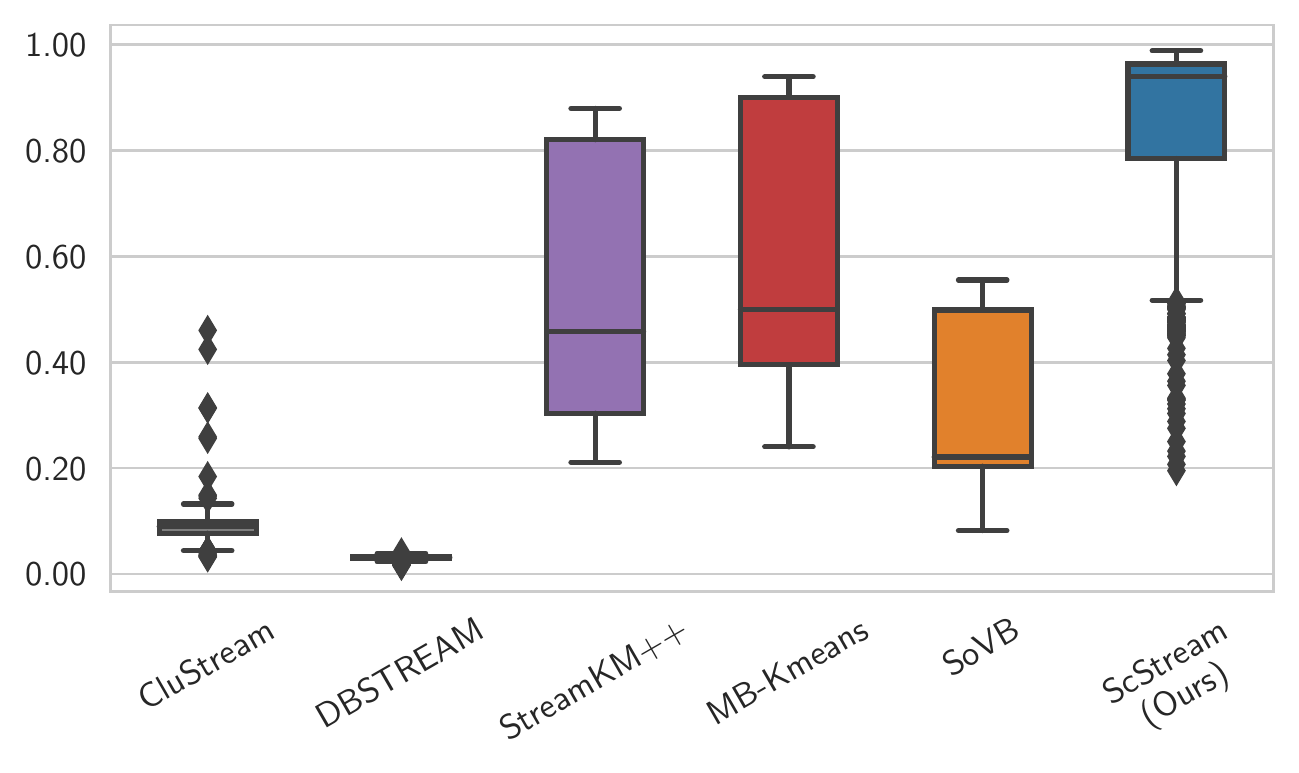}\MySpace}
    \subcaptionbox{20NewsGroups\label{Fig:Purity:20NewsGroups}}[0.32\linewidth]{\includegraphics[height=\MyHeight,trim=0.0cm 0.0cm 0.0cm 0.0cm]{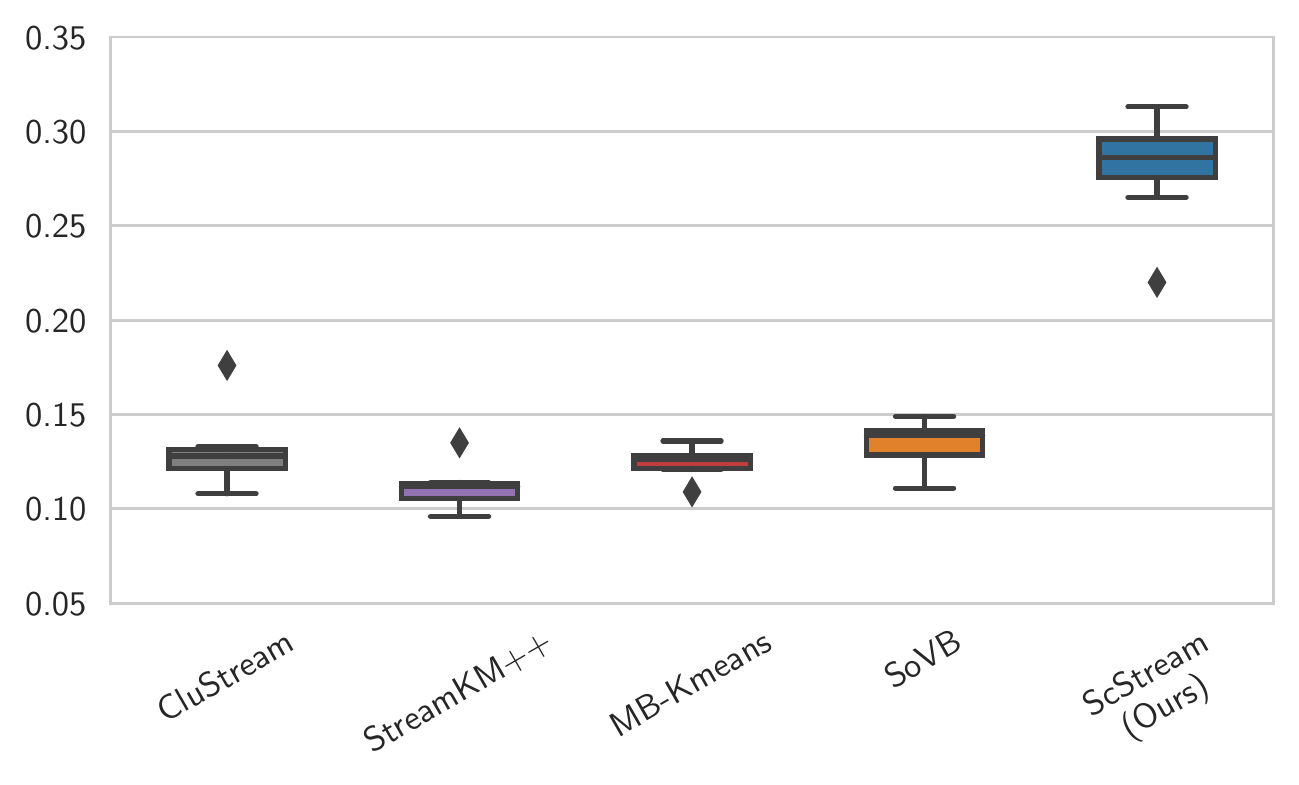}\MySpace}
\captionsetup{justification=centering, singlelinecheck=false}    
\caption{Box plots of the Purity metric for each of the experiments.
    }
     \label{Fig:purity}
\end{figure*}

\begin{figure*}[h]
    \centering
    \newcommand{\MySpace}{\vspace{-.15cm}}
    \newcommand{\MyHeight}{3.0cm}
    \subcaptionbox{Gaussian 2D\label{Fig:fmeasures:Gaussian2d}}[0.32\linewidth]{\includegraphics[height=\MyHeight,trim=0.0cm 0.0cm 0.0cm 0.0cm]{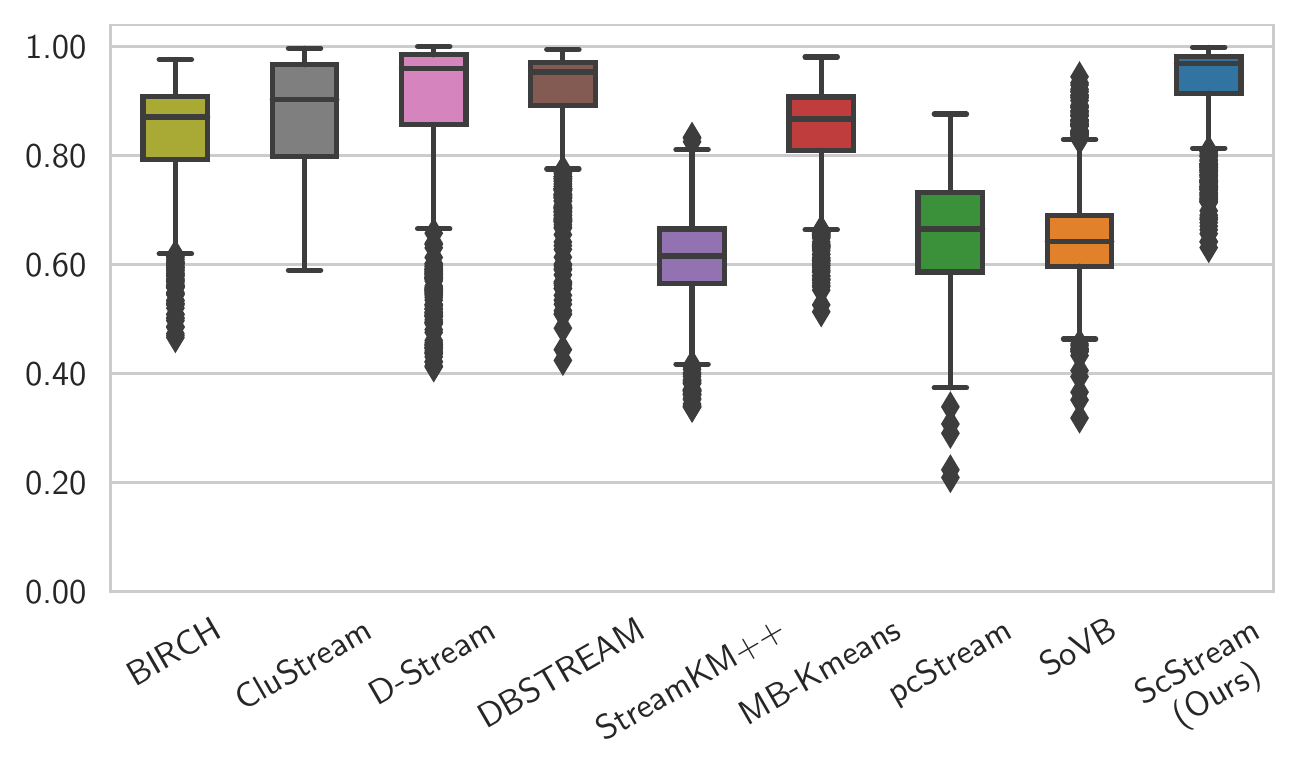}\MySpace}
    \subcaptionbox{CoverType\label{Fig:fmeasures:CoverType}}[0.32\linewidth]{\includegraphics[height=\MyHeight,trim=0.0cm 0.0cm 0.0cm 0.0cm]{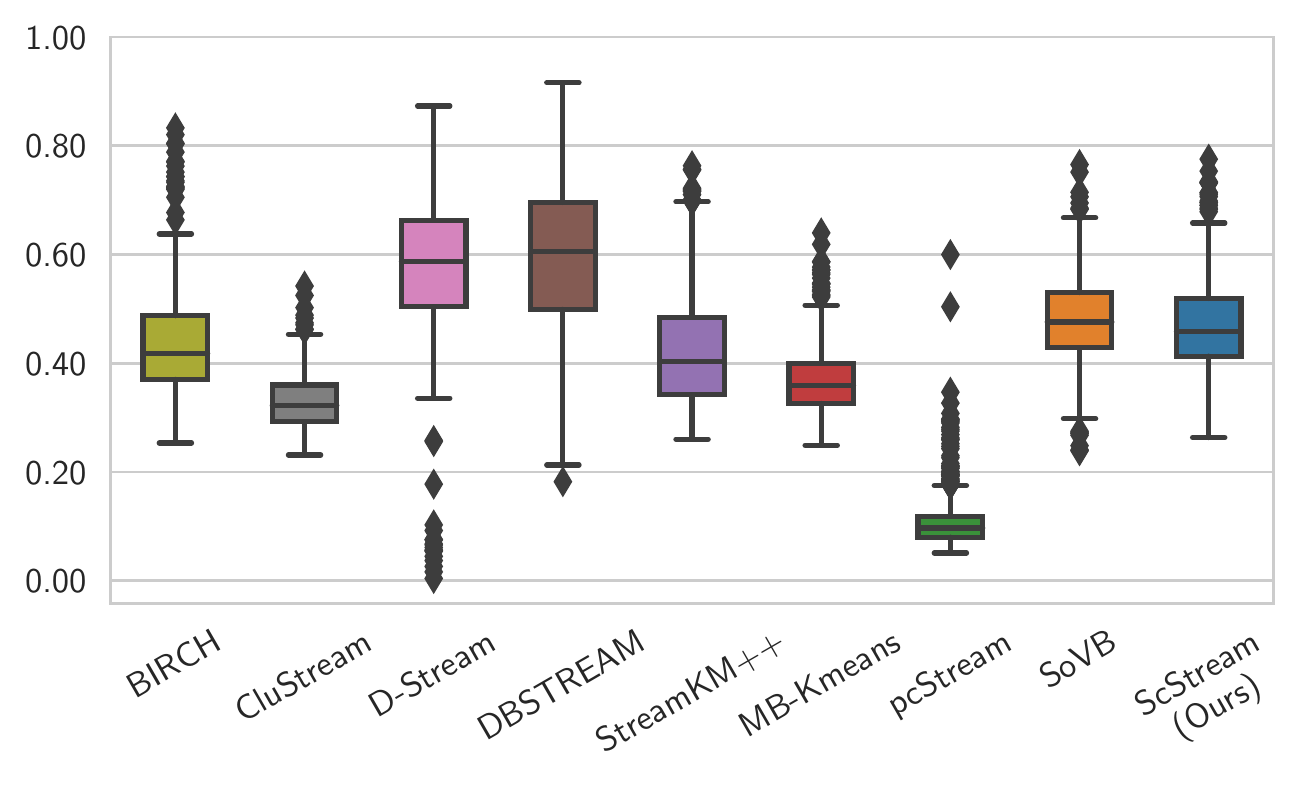}\MySpace}
    \subcaptionbox{ImageNet100\label{Fig:fmeasures:ImageNet100}}[0.32\linewidth]{\includegraphics[height=\MyHeight,trim=0.0cm 0.0cm 0.0cm 0.0cm]{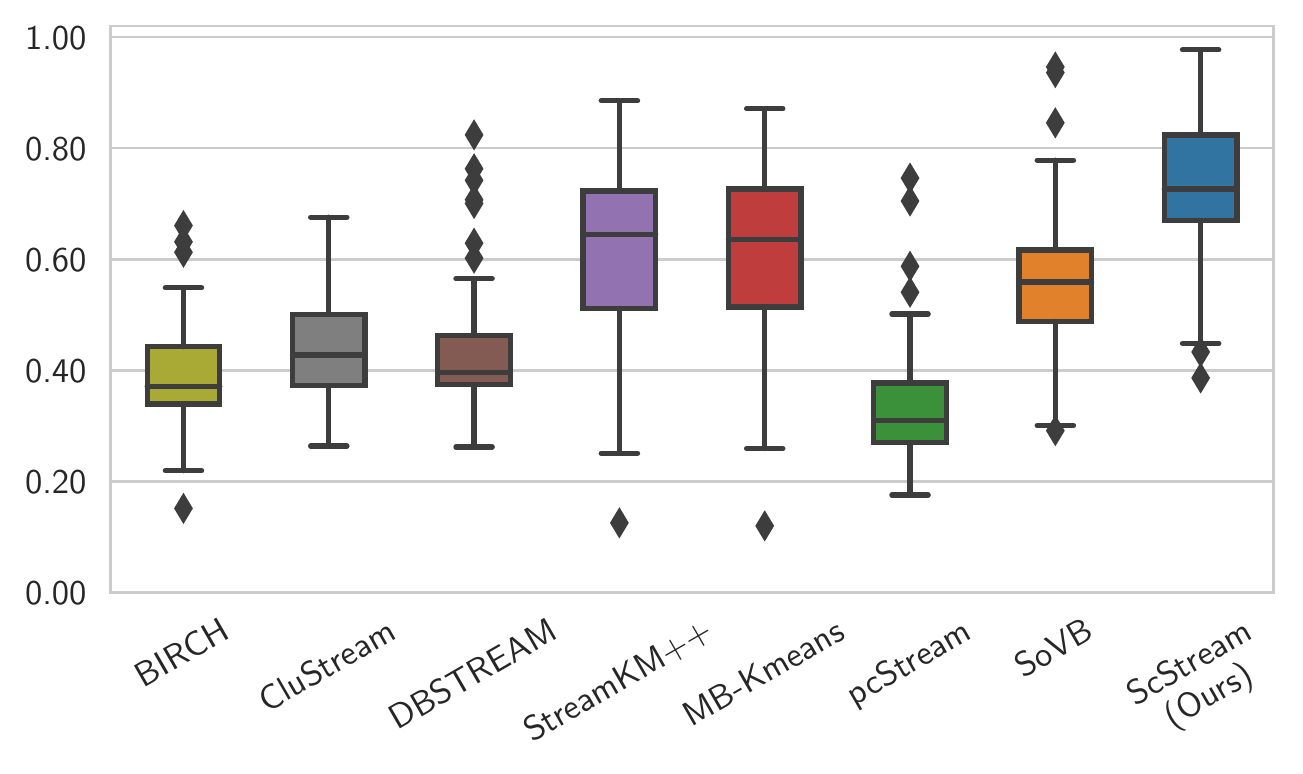}\MySpace}
    \subcaptionbox{ImageNet1K\label{Fig:fmeasures:ImageNet1K}}[0.32\linewidth]{\includegraphics[height=\MyHeight,trim=0.0cm 0.0cm 0.0cm 0.0cm]{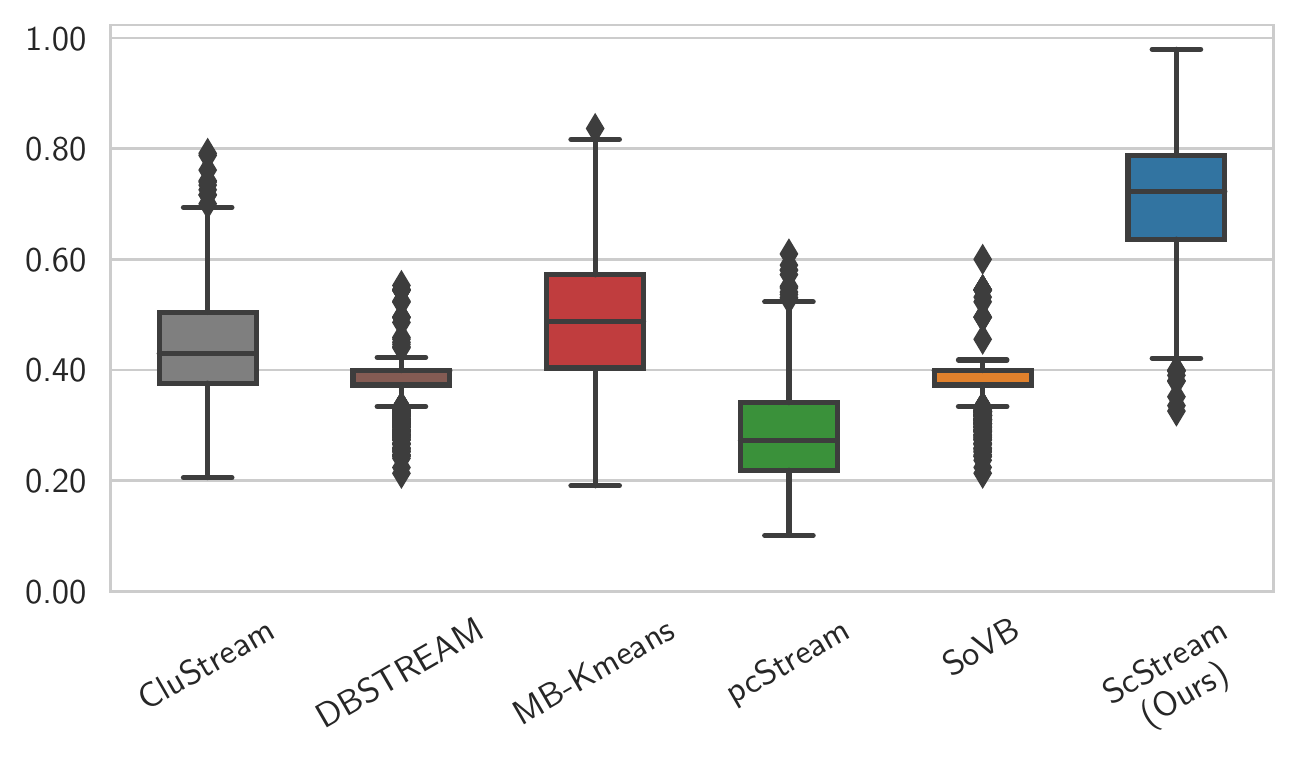}\MySpace}
    \subcaptionbox{Multinomial\label{Fig:fmeasures:Multinomial}}[0.32\linewidth]{\includegraphics[height=\MyHeight,trim=0.0cm 0.0cm 0.0cm 0.0cm]{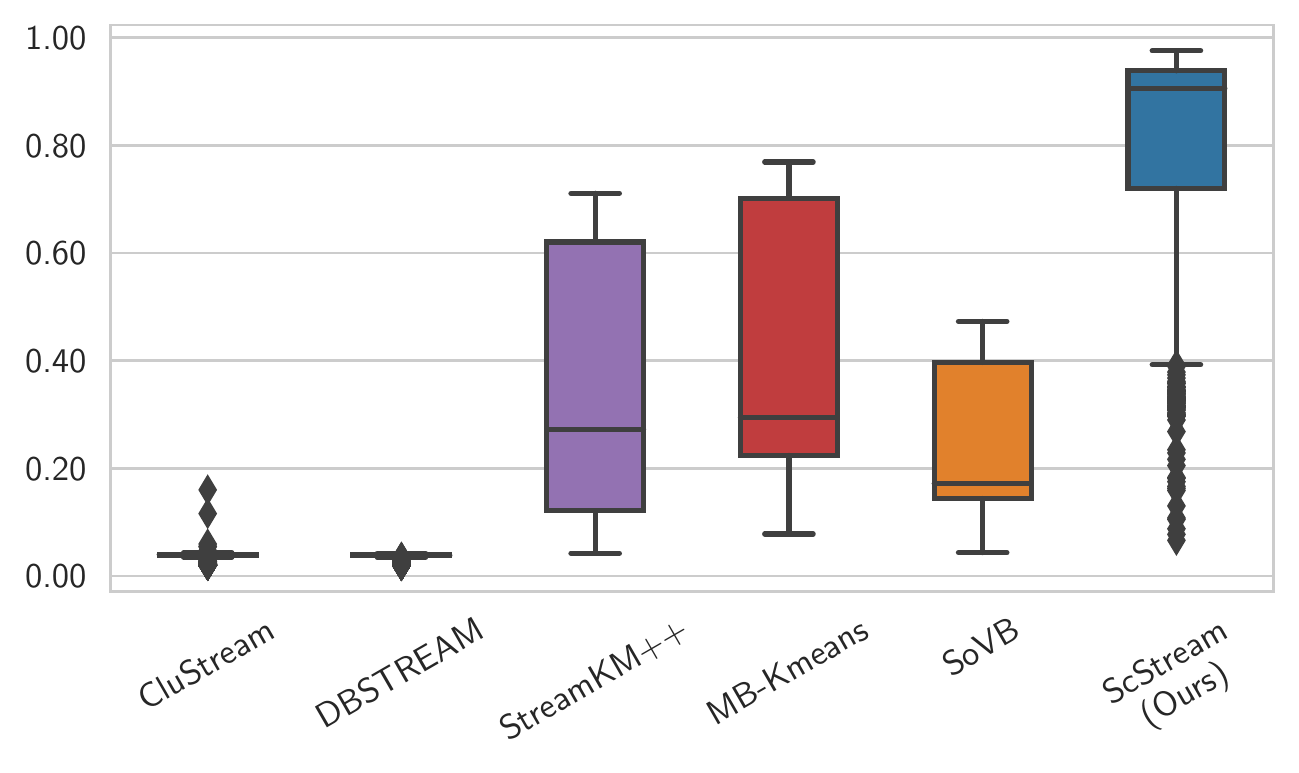}\MySpace}
    \subcaptionbox{20NewsGroups\label{Fig:fmeasures:20NewsGroups}}[0.32\linewidth]{\includegraphics[height=\MyHeight,trim=0.0cm 0.0cm 0.0cm 0.0cm]{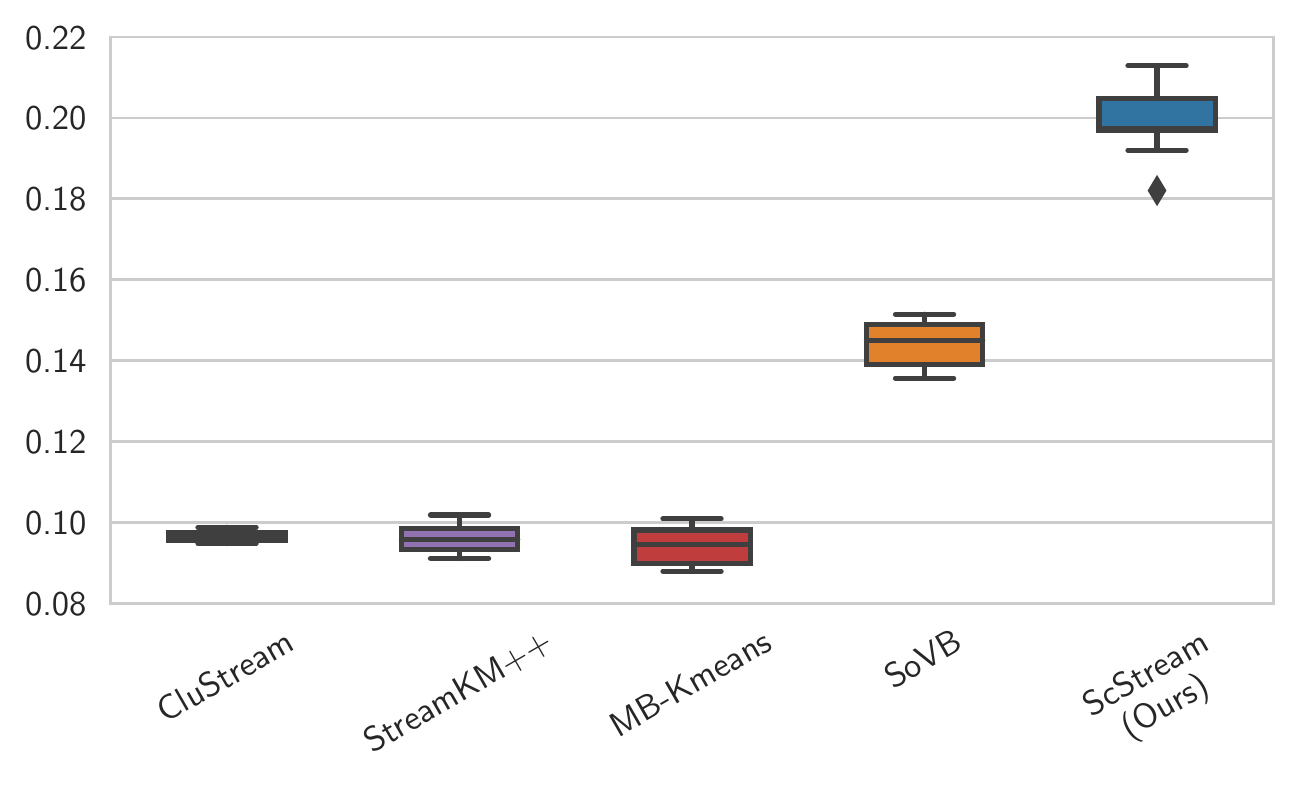}\MySpace}
\captionsetup{justification=centering, singlelinecheck=false}    
\caption{Box plots of the Pairwise F-measure metric for each of the experiments.
    }
     \label{Fig:fmeasures}
\end{figure*}
\clearpage

\section{An Empirical Verification that the Runtime Grows Linearly with $T$ (\ie, the Number of Iterations
 of the Restricted Gibbs Sampler) and that a Very Low $T$ Suffices for Good Results}
\begin{figure}[h]
    \centering
    \subcaptionbox{Running time as a function of $T$.\label{fig:it_count_metric}}{\includegraphics[height=5.5cm,trim=0 0 0 0, clip]{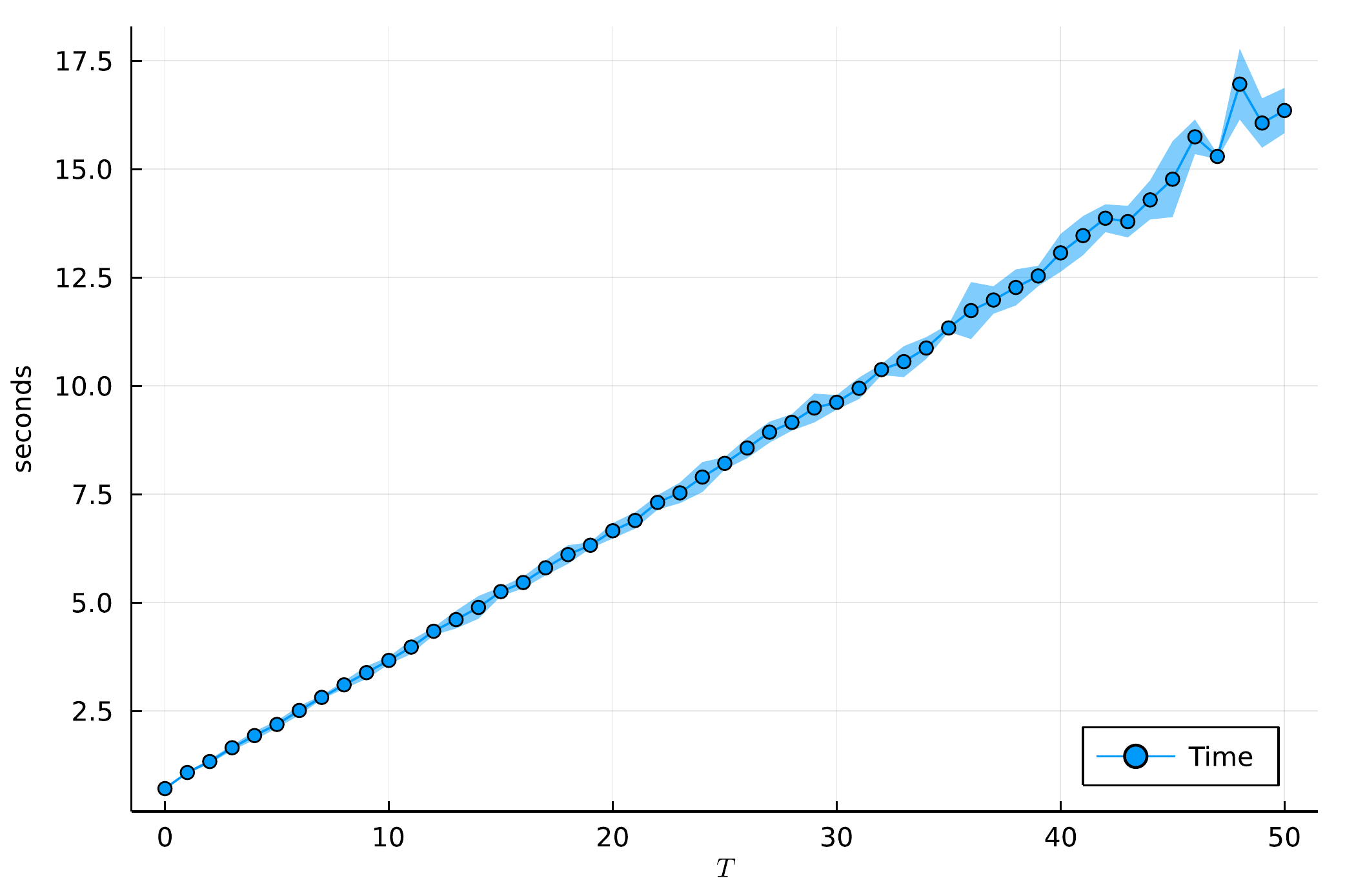}}
    \subcaptionbox{Performance as a function of $T$.\label{fig:it_count_metric}}{\includegraphics[height=5.5cm,trim=0 0 0 0, clip]{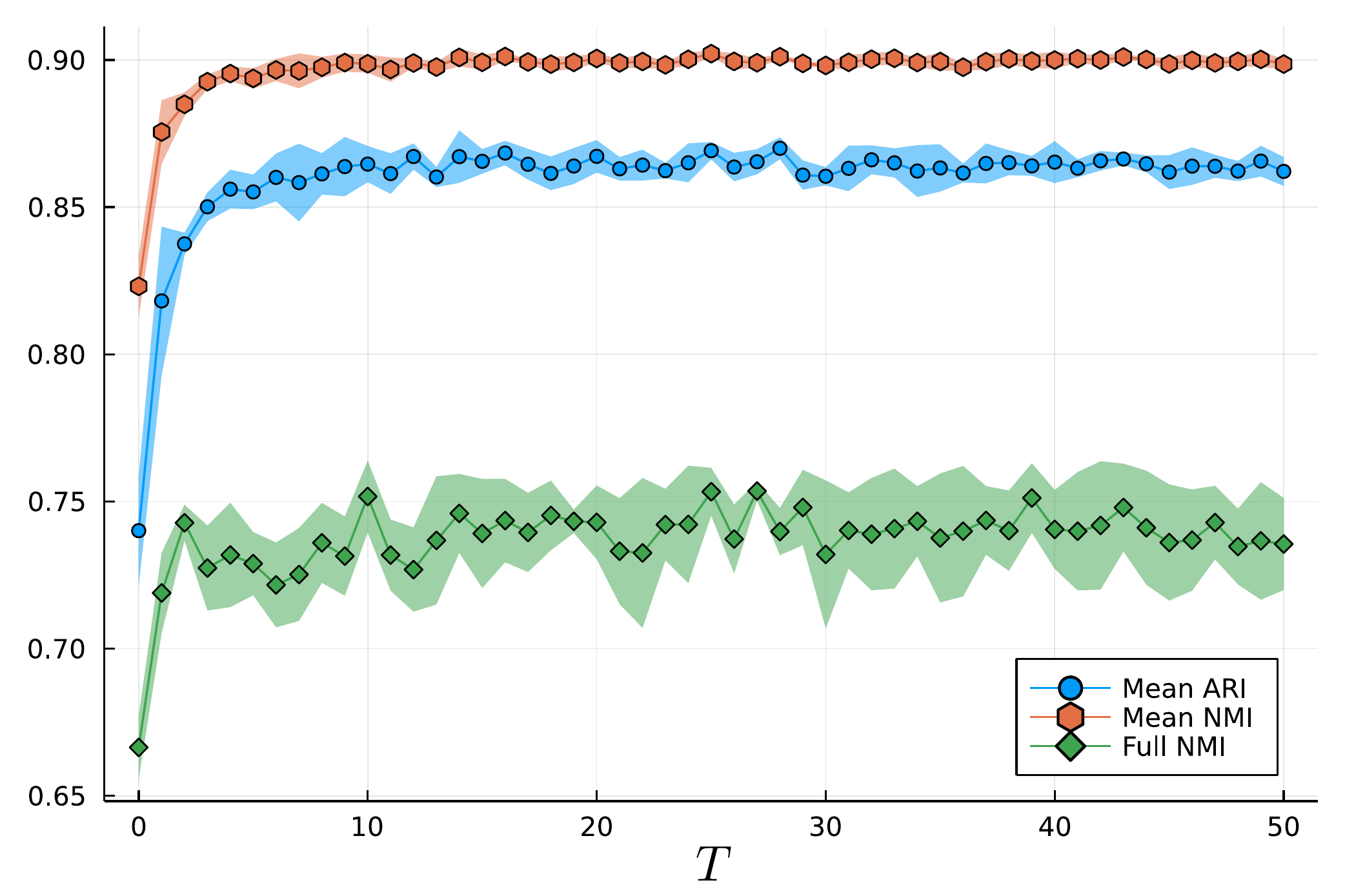}}    
         \caption{Running time and performance as functions to $T$}  
         \label{Fig:T}
\end{figure}

 Recall that $T$ is the number of iterations for which we run the restricted Gibbs sampler
 on each batch. Note that  $T$ should not be confused with the arrival time of the batch (we used $b$, not $T$, to denote the  the batch index). 
To empirically validate that runtime grows linearly with $T$, 
we ran our method on a synthetic Gaussian dataset with $10^5$ observations sampled from 20 overlapping components where, in addition, we have inserted both incremental and gradual concept drifts to the data.
The results, in terms of running time and performance  as functions of $T$, appear in~\autoref{Fig:T}. 
It is observable that while the performance gain initially increases with $T$, a plateau is quickly reached. 
Thus, further increasing $T$ will result in smaller and smaller gains. The runtime, however, is linear with $T$, thus using a low $T$ is usually preferred. 
Particularly, note that $T=1$ already achieves good results. 
Therefore, all the experiments in the paper were done with $T=1$.

\clearpage
\section{Iteration of the Restricted Gibbs Sampler}

{
    \SetKwComment{Comment}{}{}
    \begin{algorithm}[h]
    \KwIn{$H$, $\alpha$, $K$,$\bN$, $(S_k,\bar{S}_{k,1},\bar{S}_{k,2},N_k,\bar{N}_{k,1},\bar{N}_{k,2})_{k=1}^K$}
    \KwOut{$(z_i,\bar{z}_i)^N_{i=1}$}
    \KwData{$X$}
    \DontPrintSemicolon
    Draw weights by $\bpi \sim \mathrm{Dir}(N_1,\ldots,N_k,\alpha)$ \Comment*[f]{// a $K+1$ dimensional Dirichlet Distribution}\\
    \For{$k\in\set{1,\ldots,K}$}
    {
      Draw subcluster weights by $\bar{\bpi}_k \sim \mathrm{Dir}(\bar{N}_{k,1}+\frac{\alpha}{2},\bar{N}_{k,2}+\frac{\alpha}{2})$ \Comment*[f]{// a $2$-dimensional Dirichlet Distribution}
    }
    \For{$k\in\set{1,\ldots,K}$}
    {
      Draw cluster $\theta_k$ parameters by $\theta_k \sim f_\theta(\theta_k;S_k,N_k,H)$\\
      Draw subcluster $\bar{\theta}_{k,1}$ parameters by $\bar{\theta}_{k,1} \sim f_\theta(\bar{\theta}_{k,1};\bar{S}_{k,1},\bar{N}_{k,1},H)$\\
      Draw subcluster $\bar{\theta}_{k,2}$ parameters by $\bar{\theta}_{k,2} \sim f_\theta(\bar{\theta}_{k,2};\bar{S}_{k,2},\bar{N}_{k,2},H)$\\
    }
    \For{$i\in\set{1,\ldots,N}$}
    {
     Draw cluster label for $\bx_i$ with $p(z_i=k|x_i,\pi_k,\theta_k) \propto \pi_k f_\bx(\bx_i;\theta_k,z_i=k)$\\
     Draw subcluster label for $\bx_i$ with $p(\bar{z}_i=j|x_i,\bar{\pi}_{k,j},\bar{\theta}_{k,j}) \propto \bar{\pi}_{k,j} f_\bx(\bx_i;\bar{\theta}_{k,j},\bar{z}_i=j)$\\
    }
    \caption{Iteration of the Original Restricted Gibbs Sampler~\citep{Chang:NIPS:2013:ParallelSamplerDP}}
    \label{alg:originalrestricted_iter}
    \end{algorithm}
    }

\autoref{alg:originalrestricted_iter} (using a notation that is consistent with the one in our paper) is a single iteration
of the restricted Gibbs sampler from~\cite{Chang:NIPS:2013:ParallelSamplerDP}.
\section{Marginal Likelihoods}
For full derivation of the well-known results below, see~\cite{Chang:Thesis:2014:sampling}.
\subsection{Gaussian}
Let $X$ be the points in the cluster, let $N=|X|$ be the number of points, and let $D$ be the dimension of each point. Let $(\kappa,\bm,\nu,\bPsi)$ denote the $\mathrm{NIW}$ prior hyperparameters,
 and let $(\kappa^*,\bm^*,\nu^*,\bPsi^*)$ denote the posterior hyperparameters, as calculated in Example 1 in the main paper.
 The marginal likelihood of $X$ under the above parameters is:
 \begin{align}
     p(X) = \frac{\Gamma_D(\nu^*)|\nu\psi|^{\frac{\nu}{2}}}{\pi^{\frac{ND}{2}}\Gamma_D(\frac{\nu}{2})|\nu^*\psi^*|^{\frac{\nu^*}{2}}}\left(\frac{\kappa}{\kappa^*}\right)^{\frac{D}{2}}
 \end{align}
 where $\Gamma_D(\cdot)$ is the multivariate Gamma function of dimension $D$ and $|\cdot|$ is the determinant.

 \subsection{Multinomial}
 \label{sec:marginal:multinomial}
 Let $X$ be the points in the cluster, and let $N=|X|$ be the number of points. Each point in 
 \begin{align}\bx_i=\MATRIX{x_{i1}& \ldots & x_{iD}}\in X\end{align} is a $D$-length histogram,
 where $x_{ij}$ entry in it corresponds to how many times outcome $j$ was observed in the $i$-ith experiment. 
 Let $\mathrm{Dir}(d_1,\ldots,d_D)$ be the Dirichlet-distribution prior
 and let $A=\sum_{j=1}^D d_j$. 
 The marginal likelihood of $X$ is: 
 \begin{align}
     p(X)=\frac{N!\Gamma(A)}{\prod_{j=1}^D (x_{ij}!)\Gamma(A+N)}\prod_{j=1}^D\frac{d_j+x_{ij}}{\Gamma(d_j)}
     \, . 
 \end{align}

\section{Multinomial Posterior Calculation}
Let $X_b=(\bx_1,\dots,\bx_{n_b})$ denote the data points (in $\mathbb{Z}_{\ge0}^D$) in batch $b$. 
Using a classical result~\citep{Gelman:Book:2013:Bayesian}, the sufficient statistics here are 
\begin{align}
s^b_k = \left(\sum\nolimits_{\bx_i \in X_b}\bx_i \indicator _{z_i=k}\right) .
\end{align}
Let $(d_1,\ldots,d_D)$ be the hyperparameters for the Dirichlet distribution prior and its hyperparameters.
By conjugacy~\citep{Gelman:Book:2013:Bayesian}, and when using our time-weighted sufficient statistics, the hyperparameters of the posterior are:
\begin{align}
    (d^*_1,\ldots,d^*_D) = (d_1,\ldots,d_D)+\sum\nolimits_{b={q}}^B 
    \left[\Kcal(B,b) \sum\nolimits_{\bx_i \in X_b}\bx\indicator_{z_i=k}\right] \, . 
    \label{eqn:multinomial_post}
\end{align}

\section{Predictive Posterior of the Multinomial Distribution}
Following the notation in~\autoref{sec:marginal:multinomial},
and letting $\mathrm{Dir}(d_1^*,\ldots,d_D^*)$ denote the Dirichlet-distribution \emph{posterior}
and $A^*=\sum_{j=1}^D d_j^*$ denote the sum of the posterior hyperparameters, 
the predictive posterior for the multionmial distribution is the following $\mathrm{DirMult}$ distribution:
\begin{align}
    \mathrm{DirMult}(\bx_i;d^*_1,\ldots,d^*_D)=\frac{(\sum_{j=1}^D x_{ij})!}{\prod_{j=1}^D x_{ij}!}\frac{\Gamma(A^*)}{\Gamma(A^*+\sum_{j=1}^D x_{ij})}
    \prod_{j=1}^D\frac{\Gamma(d^*_j+x_{ij})}{\Gamma(d^*_j)}  
\end{align}
where $\bx_i$ is a single sample $\bx_i=(x_{i_1},\ldots,x_{i_D})$. 
Conveniently,  
when conditioning on $k$ (and adding the component weight) we can drop multiplicative constants from the RHS of the equation above, simplifying the expression:
\begin{align}
    p(z_i\hspace{-.08cm} =\hspace{-.08cm}k|\bx_i,H,S^B_k,N^B_k)
    \propto 
      \pi_k
        p(\bx_i|\bd_k^*=(d^*_1,\ldots,d^*_D),z_i=k) 
        = \pi_k\frac{\Gamma(A^*)}{\Gamma(A^*+\sum_{j=1}^D x_{ij})}
    \prod_{j=1}^D\frac{\Gamma(d^*_j+x_{ij})}{\Gamma(d^*_j)} 
\end{align}

\section{Experiments Details}
\textbf{Machine Spec:}
All the experiments were done on an Ubuntu 20.4 machine with an Intel® Core™ i9-11900K Processor.
For Julia we used version 1.6, for Python version 3.8 and for R version 4.1.1.\\
\\
\subsection{Experiment Parameters}
All the hyperparameters, for each of the competing methods and each of the datasets,
were tuned by black-box optimizers; see the paper for details.
All the parametric models were given the true value of $K$.
Here we provide the values of optimized parameters (found by the black-box optimizers) which were used in each of the experiments.\\
\\
\textbf{Gausian 2D:}\\
DBSTREAM: $\lambda=0.01,r=0.5768,Cm = 0.39244, \alpha = 0.2709, gaptime = 2971$. \\
D-Stream: $\lambda=0.01,gridsize=0.9561, gaptime=4931, Cm = 2.92, Cl=2.6721$. \\
CluStream: $\lambda=0.01,t=4$. \\
BIRCH: $\lambda=0.01,threshold=1.9339,branching=2,maxLeaf=26$.\\
StreamKM: $\lambda=0.01,sizeCoreset = 1000$. \\
PcStream: $driftThreshold = 0.74308542,percentVarience = 4.66021295, maxDriftSize= 250.13468314\cdot 2$. \\
SoVB:$\kappa = 1,\bm = zeros(2), \nu = 4,\Psi = \mathbb{I}\cdot 1.02,rhoexp=0.55,\alpha=1.0, rhodelay=1$.\\
ScStream: $\lambda = 1, \kappa = 1,\bm = zeros(2), \nu = 4,\Psi = \mathbb{I}\cdot1.02,\alpha=1.0, \epsilon=1e-08 $.
\\\\
\textbf{CoverType:}\\
DBSTREAM: $\lambda=0.01,r=0.7715,Cm= 2.748,alpha= 0.173,gaptime=597$. \\
D-Stream: $\lambda=0.01,gridsize=0.6142, gaptime=4490, Cm = 2.143, Cl=1.533$. \\
CluStream: $\lambda=0.01,t=3$. \\
BIRCH: $\lambda=0.01,treshold=1.4011,branching=1,maxLeaf=32$.\\
StreamKM: $\lambda=0.01,sizeCoreset = 1000$. \\
PcStream: $driftThreshold = 0.80786776,percentVarience = 3.5408475, maxDriftSize= 15.0128437\cdot 10$. \\
SoVB:$\kappa = 1,\bm = zeros(10), \nu = 25,\Psi = \mathbb{I}\cdot 0.78,rhoexp=0.55,\alpha=1.0, rhodelay=1$.\\
ScStream: $\lambda = 0.2, \kappa = 1,\bm = zeros(10), \nu = 25,\Psi = \mathbb{I}\cdot 0.78,\alpha=1.0, \epsilon=1e-08 $.
\\\\
\textbf{ImageNet100:}\\
DBSTREAM: $\lambda=0.01,r=2.5092,Cm= 1.2568,alpha= 0.118,gaptime=3528$. \\
D-Stream: $\lambda=0.01,gridsize=0.6216, gaptime=2924, Cm = 2.7936, Cl=1.8295$. \\
CluStream: $\lambda=0.01,t=3$. \\
BIRCH: $\lambda=0.01,treshold=1.9445,branching=3,maxLeaf=5$.\\
StreamKM: $\lambda=0.01,sizeCoreset = 1000$. \\
PcStream: $driftThreshold = 1.23666252,percentVarience = 1.50457986, maxDriftSize= 9.50255966\cdot 64$. \\
SoVB:$\kappa = 1,\bm = zeros(64), \nu = 562,\Psi = \mathbb{I}\cdot 0.51,rhoexp=0.55,\alpha=1.0, rhodelay=1$.\\
ScStream: $\lambda = 0.2, \kappa = 1,\bm = zeros(64), \nu = 562,\Psi = \mathbb{I}\cdot 0.51,\alpha=1.0, \epsilon=1e-08 $.
\\\\
\textbf{ImageNet1000:}\\
DBSTREAM: $\lambda=0.01,r=1.8247,Cm= 1.675,alpha= 0.1798,gaptime=454$. \\
CluStream: $\lambda=0.01,t=5$. \\
PcStream: $driftThreshold = 1.39396418,percentVarience = 1.91681159, maxDriftSize= 4.78852301\cdot 128$. \\
SoVB:$\kappa = 1,\bm = zeros(128), \nu = 834,\Psi = \mathbb{I}\cdot 0.177,rhoexp=0.55,\alpha=1.0, rhodelay=1$.\\
ScStream: $\lambda = 0.2, \kappa = 1,\bm = zeros(128), \nu = 834,\Psi = \mathbb{I}\cdot 0.177,\alpha=1.0, \epsilon=1e-08 $.

\end{document}